\documentclass[final,authoryear,3p,times,twocolumn]{elsarticle}
\pdfoutput = 0

\usepackage{amssymb}

\usepackage{here,wrapfig}

\usepackage{amsmath,amssymb,bm} 
\usepackage{url}

\usepackage{tabularx}
\usepackage{multirow}
\usepackage{color}
\usepackage{colortbl}
\usepackage{algorithm}
\usepackage{algorithmic}
\usepackage[table]{xcolor}
\usepackage[caption=false]{subfig}
\usepackage{hhline}
\usepackage{booktabs}
\usepackage{amsthm}
\usepackage{refcount}
\usepackage[toc,page]{appendix}

\usepackage{color}
\usepackage[normalem]{ulem}
\definecolor{cyan}{cmyk}{1,0,0,0}

\newcommand\indep{\protect\mathpalette{\protect\independenT}{\perp}}
\def\independenT#1#2{\mathrel{\rlap{$#1#2$}\mkern2mu{#1#2}}}

\newtheorem{thm}{Theorem}
\newtheorem*{thm*}{Theorem}
\newtheorem{lem}[thm]{Lemma}
\newtheorem*{lem*}{Lemma}
\newtheorem{defn}{Definition}
\newproof{pf}{Proof}
\newproof{pot}{Proof of Theorem \ref{thm2}}

\definecolor{n1}{gray}{0.9} 
\definecolor{n2}{gray}{0.75} 
\definecolor{n3}{gray}{0.6} 

\newlength{\tabcapsep}
\journal{Expert Systems With Applications}

\begin{document}

\begin{frontmatter}



\title{A hybrid algorithm for Bayesian network structure learning\\
with application to multi-label learning}


\author{Maxime Gasse}
\author{Alex Aussem\corref{cor1}}
\ead{aaussem@univ-lyon1.fr}
\author{Haytham Elghazel}

\cortext[cor1]{Corresponding author}

\address{Universit\'e de Lyon, CNRS\\
Universit\'e Lyon 1, LIRIS, UMR5205, F-69622, France}

\begin{abstract}
We present a novel hybrid algorithm for Bayesian network structure learning, called H2PC. It first reconstructs the skeleton of a Bayesian network and then performs a Bayesian-scoring greedy hill-climbing search to orient the edges. The algorithm is based on divide-and-conquer constraint-based subroutines to learn the local structure around a target variable.  We conduct two series of experimental comparisons of H2PC against Max-Min Hill-Climbing (MMHC), which is currently the most powerful state-of-the-art algorithm for Bayesian network structure learning. First, we use eight well-known Bayesian network benchmarks with various data sizes to assess the quality of the learned structure returned by the algorithms. Our extensive experiments show that H2PC outperforms MMHC in terms of goodness of fit to new data and quality of the network structure with respect to the true dependence structure of the data. Second, we investigate H2PC's ability to solve the multi-label learning problem. We provide theoretical results to characterize and identify graphically the so-called minimal label powersets that appear as irreducible factors in the joint distribution under the faithfulness condition. The multi-label learning problem is then decomposed into a series of multi-class classification problems, where each multi-class variable encodes a label powerset. H2PC is shown to compare favorably to MMHC in terms of global classification accuracy over ten multi-label data sets covering different application domains. Overall, our experiments support the conclusions that local structural learning with H2PC in the form of local neighborhood induction is a theoretically well-motivated and empirically effective learning framework that is well suited to multi-label learning. The source code (in \textit{R}) of H2PC as well as all data sets used for the empirical tests are publicly available.
\end{abstract}


\end{frontmatter}

\section{Introduction}

A Bayesian network (BN) is a probabilistic model formed by a structure and parameters. The structure of a BN is a directed acyclic graph (DAG), whilst its parameters are conditional probability distributions associated with the variables in the model. The problem of finding the DAG that encodes the conditional independencies present in the data attracted a great deal of interest over the last years \citep{Morais10b,Scutari10,Scutari12,Kojima10,Perrier08,Villanueva12,Pen12,Gasse12}. The inferred DAG is very useful for many applications, including feature selection \citep{Aliferis10a,Pen07,Morais10a}, causal relationships inference from observational data \citep{Byron08,Aliferis10a,Aussem12a,Aussem10c,Prestat13,Cawley08,Brown08} and more recently multi-label learning \citep{Dembczynski2012,Zhang10,Guo11}.

Ideally the DAG should coincide with the dependence structure of the global distribution, or it should at least identify a distribution as close as possible to the correct one in the probability space. This step, called structure learning, is similar in approaches and terminology to model selection procedures for classical statistical models. Basically, constraint-based (CB) learning methods systematically check the data for conditional independence relationships and use them as constraints to construct a partially oriented graph representative of a BN equivalence class, whilst search-and-score (SS) methods make use of a goodness-of-fit score function for evaluating graphical structures with regard to the data set. Hybrid methods attempt to get the best of both worlds: they learn a skeleton with a CB approach and constrain on the DAGs considered during the SS phase.

In this study, we present a novel hybrid algorithm for Bayesian network structure learning, called H2PC\footnote{A first version of HP2C without FDR control has been discussed in a paper that appeared in the Proceedings of ECML-PKDD, pages 58-73, 2012.}. It first reconstructs the skeleton of a Bayesian network and then performs a Bayesian-scoring greedy hill-climbing search to orient the edges. The algorithm is based on divide-and-conquer constraint-based subroutines to learn the local structure around a target variable. HPC may be thought of as a way to compensate for the large number of false negatives at the output of the weak PC learner, by performing extra computations. As this may arise at the expense of the number of false positives, we control the expected proportion of false discoveries (i.e. false positive nodes) among all the discoveries made in $\textbf{PC}_T$. We use a modification of the Incremental  association Markov boundary algorithm (IAMB), initially developed by Tsamardinos et al. in \citep{Tsamardinos03} and later modified by Jose Pe\~na in \citep{Pena08} to control the FDR of edges when learning Bayesian network models. HPC scales to thousands of variables and can deal with many fewer samples ($n<q$). To illustrate its performance by means of empirical evidence, we conduct two series of experimental comparisons of H2PC against Max-Min Hill-Climbing (MMHC), which is currently the most powerful state-of-the-art algorithm for BN structure learning \citep{Tsamardinos06}, using well-known BN benchmarks with various data sizes, to assess the goodness of fit to new data as well as the quality of the network structure with respect to the true dependence structure of the data.

We then address a real application of H2PC where the true dependence structure is unknown. More specifically, we investigate H2PC's ability to encode the joint distribution of the label set conditioned on the input features in the multi-label classification (MLC) problem. Many challenging applications, such as photo and video annotation and web page categorization, can benefit from being formulated as MLC tasks with large number of categories \citep{Dembczynski2012,Read2009,Madjarov2012,Kocev2007,Tsoumakas2010}. Recent research in MLC focuses on the exploitation of the label conditional dependency in order to better predict the label combination for each example.  We show that local BN structure discovery methods offer an elegant and powerful approach to solve this problem. We establish two theorems (Theorem 6 and 7) linking the concepts of marginal Markov boundaries, joint Markov boundaries and so-called label powersets under the faithfulness assumption. These Theorems offer a simple guideline to characterize graphically : i) the minimal label powerset decomposition, (i.e. into minimal subsets $\mathbf{Y}_{LP} \subseteq \mathbf{Y}$ such that  $\mathbf{Y}_{LP} \indep \mathbf{Y} \setminus \mathbf{Y}_{LP} \mid \mathbf{X}$), and ii) the minimal subset of features, w.r.t an Information Theory criterion, needed to predict each label powerset, thereby reducing the input space and the computational burden of the multi-label classification.  To solve the MLC problem with BNs, the DAG obtained from the data plays a pivotal role.  So in this second series of experiments, we assess the comparative ability of H2PC and MMHC to encode the label dependency structure by means of an indirect goodness of fit indicator, namely the $0/1$ loss function, which makes sense in the MLC context.

The rest of the paper is organized as follows: In the Section 2, we review the theory of BN and discuss the main BN structure learning strategies. We then present the H2PC algorithm in details in Section 3. Section 4 evaluates our proposed method and shows results for several tasks involving artificial data sampled from known BNs. Then we report, in Section 5, on our experiments on real-world data sets in a multi-label learning context so as to provide empirical support for the proposed methodology. The main theoretical results appear formally  as two theorems (Theorem 8 and 9) in Section 5. Their proofs are established in the Appendix. Finally, Section 6 raises several issues for future work and we conclude in Section 7 with a summary of our contribution.

\section{Preliminaries}

We define next some key concepts used along the paper and state some results that will support our analysis. In this paper, upper-case letters in italics denote random variables (e.g., $X,Y$) and lower-case letters in italics denote their values (e.g., $x,y$). Upper-case bold letters denote random variable sets (e.g., $\mathbf{X,Y,Z}$) and lower-case bold letters denote their values (e.g., $\mathbf{x,y}$). We denote by $\mathbf{X} \indep \mathbf{Y} \mid \mathbf{Z}$ the conditional independence between $\mathbf{X}$ and $\mathbf{Y}$ given the set of variables $\mathbf{Z}$. To keep the notation uncluttered, we use $p(\mathbf{y}\mid\mathbf{x})$  to denote $p(\mathbf{Y}=\mathbf{y}\mid\mathbf{X}=\mathbf{x})$.

\subsection{Bayesian networks}

Formally, a BN is a tuple $<{\cal G}, P>$, where ${\cal G} =
<\mathbf U, \mathbf E>$ is a directed acyclic graph (DAG) with
nodes representing the random variables $\mathbf U$ and $P$ a
joint probability distribution in ${\cal U}$. In addition, ${\cal
G}$ and $P$ must satisfy the Markov condition: every variable, $X
\in \mathbf U$, is independent of any subset of its non-descendant
variables conditioned on the set of its parents, denoted by
$\mathbf{Pa}_i^{\cal G}$. From the Markov condition, it is easy to
prove \citep{Neapolitan04} that the joint probability distribution
$P$ on the variables in $\mathbf U$ can be factored as follows :

\begin{equation}
P({\cal V}) = P(X_1,\ldots,X_n) = \prod_{i=1}^n
P(X_i|\mathbf{Pa}_i^{\cal G})
\end{equation}

Equation 1 allows a parsimonious decomposition of the joint
distribution $P$. It enables us to reduce the problem of
determining a huge number of probability values to that of
determining relatively few.

A BN structure  ${\cal G}$ entails a set of conditional
independence assumptions. They can all be identified by the
\emph{d-separation criterion} \citep{Pea88}. We use $X \perp_{\cal
G} Y|\mathbf Z$ to denote the assertion that $X$ is d-separated
from $Y$ given $\mathbf Z$ in ${\cal G}$. Formally, $X \perp_{\cal
G} Y|\mathbf Z$ is true when for every undirected path in ${\cal
G}$ between $X$ and $Y$, there exists a node $W$ in the path such
that either (1) $W$ does not have two parents in the path and $W
\in \mathbf Z$, or (2) $W$ has two parents in the path and
neither $W$ nor its descendants is in $\mathbf Z$. If $<{\cal G}, P>$ is a BN,  $X \perp_P Y|\mathbf
Z$ if $X \perp_{\cal G} Y|\mathbf Z$. The converse does not
necessarily hold. We say that $<{\cal G}, P>$ satisfies the
\emph{faithfulness condition} if the d-separations in ${\cal G}$
identify \emph{all and only} the conditional independencies in
$P$, i.e., $X \perp_P Y|\mathbf Z$ iff $X \perp_{\cal G} Y|\mathbf
Z$.

A Markov blanket $\textbf{M}_T$ of $T$ is any set of variables such that $T$ is conditionally independent of all the remaining variables given $\textbf{M}_T$. By extension, a Markov blanket of $T$ in $\mathbf{V}$ guarantees that $\textbf{M}_T \subseteq \mathbf{V}$, and that $T$ is conditionally independent of the remaining variables in $\mathbf{V}$, given $\textbf{M}_T$. A Markov boundary, $\textbf{MB}_T$, of $T$ is any Markov blanket such that none of its proper subsets is a Markov blanket of $T$.

We denote by $\textbf{PC}_T^{\cal G}$, the set of parents and
children of $T$ in ${\cal G}$, and by $\textbf{SP}_T^{\cal G}$,
the set of \emph{spouses} of $T$ in ${\cal G}$. The \emph{spouses}
of $T$ are the variables that have common children with $T$. These
sets are unique for all ${\cal G}$, such that $<{\cal G},P>$
satisfies the faithfulness condition and so we will drop the
superscript ${\cal G}$. We denote by $\textbf{dSep}(X)$, the set
that d-separates $X$ from the (implicit) target $T$.

\begin{thm}
  Suppose $<{\cal G}, P>$ satisfies the faithfulness condition. Then $X$ and $Y$ are not adjacent in ${\cal G}$ iff
  $\exists \mathbf{Z} \in \mathbf{U} \setminus \{X,Y\}$
  such that
  $X \indep Y \mid \mathbf{Z}$
  . Moreover,
  $\mathbf{MB}_X=\mathbf{PC}_X \cup \mathbf{SP}_X$
  .
\end{thm}

A proof can be found for instance in \citep{Neapolitan04}.

Two graphs are said \emph{equivalent} iff they encode the same set
of conditional independencies via the d-separation criterion. The
equivalence class of a DAG ${\cal G}$ is a set of DAGs that are
equivalent to ${\cal G}$. The next result showed by \citep{Pea88},
establishes that equivalent graphs have the same undirected graph
but might disagree on the direction of some of the arcs.

\begin{thm}
Two DAGs are equivalent iff they have the same underlying
undirected graph and the same set of v-structures (i.e. converging
edges into the same node, such as $X \rightarrow Y \leftarrow Z$).
\end{thm}

Moreover, an equivalence class of network structures can be
uniquely represented by a partially directed DAG (PDAG), also
called a DAG pattern. The DAG pattern is defined as the graph that
has the same links as the DAGs in the equivalence class and has
oriented all and only the  edges common to the DAGs in the
equivalence class. A structure learning algorithm from data is
said to be correct (or sound) if it returns the correct DAG
pattern (or a DAG in the correct equivalence class) under the
assumptions that the independence tests are reliable and that the
learning database is a sample from a distribution $P$ faithful to
a DAG ${\cal G}$, The (ideal) assumption that the independence
tests are reliable means that they decide (in)dependence iff the
(in)dependence holds in $P$.

\subsection{Conditional independence properties}

The following three theorems, borrowed from~\cite{Pen07}, are proven in~\cite{Pea88}:

\begin{thm}
  \label{th:4-axioms}
Let $\mathbf{X,Y,Z}$ and $\mathbf{W}$ denote four mutually disjoint subsets of $\mathbf{U}$. Any probability distribution $p$ satisfies the following four properties: Symmetry $\mathbf X \indep\mathbf Y \mid \mathbf Z \Rightarrow \mathbf Y  \indep \mathbf X \mid \mathbf Z$, Decomposition, $\mathbf X \indep (\mathbf Y \cup \mathbf W) \mid \mathbf Z \Rightarrow \mathbf X \indep\mathbf Y \mid \mathbf Z$, Weak Union $\mathbf X \indep (\mathbf Y \cup \mathbf W) \mid \mathbf Z \Rightarrow \mathbf X \indep\mathbf Y \mid (\mathbf Z \cup  \mathbf W)$ and Contraction, $\mathbf X \indep \mathbf Y  \mid (\mathbf Z \cup \mathbf W) \wedge \mathbf X \indep \mathbf W  \mid \mathbf Z  \Rightarrow \mathbf X \indep (\mathbf Y \cup \mathbf W) \mid \mathbf Z $ .
\end{thm}

\begin{thm}
  \label{th:intersection}
  If $p$ is strictly positive, then $p$ satisfies the previous four properties plus the Intersection property $\mathbf X \indep \mathbf Y  \mid (\mathbf Z \cup \mathbf W) \wedge \mathbf X \indep \mathbf W  \mid (\mathbf Z \cup \mathbf Y)  \Rightarrow \mathbf X \indep (\mathbf Y \cup \mathbf W) \mid \mathbf Z$. Additionally, each $\mathbf X \in \mathbf{U}$ has a unique Markov boundary, $\mathbf{MB}_\mathbf{X}$
\end{thm}

\begin{thm}
  \label{th:faithfulness}
  If $p$ is faithful to a DAG $\cal{G}$, then $p$ satisfies the previous five properties plus the Composition property $\mathbf X \indep \mathbf Y  \mid \mathbf Z  \wedge \mathbf X \indep \mathbf W  \mid \mathbf Z   \Rightarrow \mathbf X \indep (\mathbf Y \cup \mathbf W) \mid \mathbf Z$ and the local Markov property  $X \indep (\mathbf{ND}_X \setminus \mathbf{Pa}_X)\mid \mathbf{Pa}_X$ for each $X \in \mathbf{U}$, where $\mathbf{ND}_X$ denotes the non-descendants of $X$ in $\cal{G}$.
\end{thm}

\subsection{Structure learning strategies}

The number of DAGs, $\mathbb{G}$, is super-exponential in the number of random variables in the domain and the problem of learning the most probable \emph{a posteriori} BN from data is worst-case NP-hard \citep{ChickeringHM04}.  One needs to resort to heuristical methods in order to be able to solve very large problems effectively.

Both CB and SS heuristic approaches have advantages and disadvantages. CB
approaches are relatively quick, deterministic, and have a well defined
stopping criterion; however, they rely on an arbitrary significance
level to test for independence, and they can be unstable in the sense
that an error early on in the search can have a cascading effect that
causes many errors to be present in the final graph. SS approaches
have the advantage of being able to flexibly incorporate users'
background knowledge in the form of prior probabilities over the
structures and are also capable of dealing with incomplete records in
the database (e.g. EM technique). Although SS methods are favored in
practice when dealing with small dimensional data sets, they are slow
to converge and the computational complexity often prevents us from
finding optimal BN structures \citep{Perrier08,Kojima10}. With currently
available exact algorithms \citep{Koivisto04,Silander06,Cussens2013,Studeny2014} and a
decomposable score like BDeu, the computational complexity remains
exponential, and therefore, such algorithms are intractable for BNs
with more than around 30 vertices on current workstations. For larger sets of variables
the computational burden becomes prohibitive and restrictions
about the structure have to be imposed, such as a limit on the size of the parent sets.
With this in mind, the ability to restrict
the search locally around the target variable is a key advantage of CB
methods over SS methods. They are able to construct a local graph
around the target node without having to construct the whole BN first,
hence their scalability \citep{Pen07,Morais10a,Morais10b,Tsamardinos06,Pena08}.

With a view to balancing the computation cost with the desired accuracy
of the estimates, several hybrid methods have been proposed recently.
Tsamardinos et al. proposed in \citep{Tsamardinos06} the Min-Max Hill Climbing (MMHC)
algorithm and conducted one of the most extensive empirical comparison
performed in recent years showing that MMHC was the fastest and the
most accurate method in terms of structural error based on the
structural hamming distance. More specifically, MMHC outperformed both
in terms of time efficiency and quality of reconstruction the PC
\citep{Spi00}, the Sparse Candidate \citep{Friedman99}, the Three Phase
Dependency Analysis \citep{Cheng02}, the Optimal Reinsertion
\citep{Moore03}, the Greedy Equivalence Search \citep{Chickering02},
and the Greedy Hill-Climbing Search on a variety of networks, sample
sizes, and parameter values. Although MMHC is rather heuristic by
nature (it returns a local optimum of the score function), it is
currently considered as the most powerful state-of-the-art algorithm
for BN structure learning capable of dealing with thousands of nodes
in reasonable time.

In order to enhance its performance on small dimensional data sets, Perrier et al. proposed in \citep{Perrier08} a hybrid
algorithm that can learn an \textit{optimal} BN (i.e., it converges
to the true model in the sample limit) when an undirected graph is
given as a structural constraint. They defined this undirected graph
as a super-structure (i.e., every DAG considered in the SS phase is
compelled to be a subgraph of the super-structure). This algorithm can
learn optimal BNs containing up to 50 vertices when the average degree
of the super-structure is around two, that is, a sparse structural
constraint is assumed. To extend its feasibility to BN with a few
hundred of vertices and an average degree up to four, Kojima et al. proposed in
\citep{Kojima10} to divide the super-structure into several clusters and
perform an optimal search on each of them in order to scale up to
larger networks. Despite interesting improvements in terms of score
and structural hamming distance on several benchmark BNs, they report
running times about $10^3$ times longer than MMHC on average, which
is still prohibitive.

Therefore, there is great deal of interest in hybrid methods capable of
improving the structural accuracy of both CB and SS methods on graphs
containing up to thousands of vertices. However, they make the strong
assumption that the skeleton (also called super-structure) contains at
least the edges of the true network and as small as possible extra
edges. While controlling the false discovery rate (i.e., false extra
edges) in BN learning has attracted some attention recently
\citep{ArmenT11,Pena08,Tsamardinos08}, to our knowledge, there is no
work on controlling actively the rate of false-negative errors (i.e.,
false missing edges).

\subsection{Constraint-based structure learning}

Before introducing the H2PC algorithm, we discuss the general idea behind CB methods.
The induction of local or global BN structures is handled by CB methods
through the identification of local neighborhoods (i.e.,
$\textbf{PC}_X$), hence their scalability to very high dimensional
data sets. CB methods systematically check the data for conditional
independence relationships in order to infer a target's neighborhood.
Typically, the algorithms run either a $G^2$ or a $\chi^2$ independence
test when the data set is discrete and a Fisher's Z test when it is
continuous in order to decide on dependence or independence, that is,
upon the rejection or acceptance of the null hypothesis of conditional
independence. Since we are limiting ourselves to discrete data, both
the global and the local distributions are assumed to be multinomial,
and the latter are represented as conditional probability tables.
Conditional independence tests and network scores for discrete data are
functions of these conditional probability tables through the observed
frequencies $\{n_{ijk};i = 1,\ldots ,R;j = 1,\ldots, C;k = 1,\ldots, L\}$
for the random variables $X$ and $Y$ and all the configurations of the
levels of the conditioning variables $\mathbf{Z}$. We use $n_{i+k}$
as shorthand for the marginal $\sum_j n_{ijk}$ and similarly for $n_{i+k}, n_{++k}$ and $n_{+++}=n$.
We use a classic conditional independence test based on the mutual information. The
mutual information is an information-theoretic distance measure defined
as

$$
MI(X, Y |\mathbf{Z}) = \sum_{i=1}^R \sum_{j=1}^C \sum_{k=1}^L
\frac{n_{ijk}}{n} \log{\frac{n_{ijk}n_{++k}}{n_{i+k}n_{+jk}}}
$$

It is proportional to the log-likelihood ratio test $G^2$ (they differ
by a 2n factor, where n is the sample size). The asymptotic null
distribution is $\chi^2$ with $(R-1)(C -1)L$ degrees of freedom. For a
detailed analysis of their properties we refer the reader to
\citep{Agresti02}. The main limitation of this test is the rate of
convergence to its limiting distribution, which is particularly
problematic when dealing with small samples and sparse contingency
tables. The decision of accepting or rejecting the null hypothesis
depends implicitly upon the degree of freedom which increases
exponentially with the number of variables in the conditional set.
Several heuristic solutions have emerged in the literature
\citep{Spi00,Morais10b,Tsamardinos06,Tsamardinos10} to overcome some
shortcomings  of the asymptotic tests. In this study we use the two
following heuristics that are used in MMHC.  First, we do not perform
$MI(X, Y |\mathbf{Z}) $ and assume independence if there are not enough
samples to achieve large enough power. We require that the average
sample per count is above a user defined parameter, equal to 5, as in
\citep{Tsamardinos06}. This heuristic is called the power rule. Second,
we consider as structural zero either case $n_{+jk}$ or $n_{i+k} = 0$. For
example, if $n_{+jk} = 0$, we consider y as a structurally forbidden value
for Y when Z = z and we reduce R by 1 (as if we had one column less in
the contingency table where Z = z).  This is known as the degrees of
freedom adjustment heuristic.

\section{The H2PC algorithm}

In this section, we present our hybrid algorithm for Bayesian network structure learning, called Hybrid HPC (H2PC). It first reconstructs the skeleton
of a Bayesian network and then performs a Bayesian-scoring greedy hill-climbing search to filter and orient the edges. It is based on a CB subroutine
called HPC to learn the parents and children of a single variable. So, we shall discuss HPC first and then move to H2PC.

\subsection{Parents and Children Discovery}

HPC (Algorithm \ref{HPC}) can be viewed as an ensemble method for combining many weak PC learners in an attempt to produce a stronger PC learner. HPC is based on three subroutines: \emph{Data-Efficient Parents and Children Superset} (DE-PCS), \emph{Data-Efficient Spouses Superset} (DE-SPS), and \emph{Incremental Association Parents and Children} with false discovery rate control (FDR-IAPC), a weak PC learner based on FDR-IAMB \citep{Pena08} that requires little computation. HPC receives a target node \emph{T}, a data set ${\cal D}$ and a set of variables \textbf{U} as input and returns an estimation of $\textbf{PC}_T$. It is hybrid in that it combines the benefits of incremental and divide-and-conquer methods. The procedure starts by extracting a superset $\textbf{PCS}_T$ of $\textbf{PC}_T$ (line 1) and a superset $\textbf{SPS}_T$ of $\textbf{SP}_T$ (line 2) with a severe restriction on the maximum conditioning size ($\textbf{Z}<= 2$) in order to significantly increase the reliability of the tests. A first candidate PC set is then obtained by running the weak PC learner called FDR-IAPC on $\textbf{PCS}_T \cup \textbf{SPS}_T$ (line 3). The key idea is the decentralized search at lines 4-8 that includes, in the candidate PC set, all variables in the superset $\textbf{PCS}_T \setminus \textbf{PC}_T$ that have $T$ in their vicinity. So, HPC may be thought of as a way to compensate for the large number of false negatives at the output of the weak PC learner, by performing extra computations. Note that, in theory, $X$ is in the output of FDR-IAPC($Y$) if and only if $Y$ is in the output of FDR-IAPC($X$). However, in practice, this may not always be true, particularly when working in high-dimensional domains \citep{Pena08}. By loosening the criteria by which two nodes are said adjacent, the effective restrictions on the size of the neighborhood are now far less severe. The decentralized search has a significant impact on the accuracy of HPC as we shall see in in the experiments. We proved in \citep{Morais10b} that the original HPC(T) is consistent, i.e. its output converges in probability to $\textbf{PC}_T$ , if the hypothesis tests are consistent. The proof also applies to the modified version presented here.

We now discuss the subroutines in more detail. FDR-IAPC (Algorithm \ref{FDR-IAPC}) is a fast incremental method that receives a data set ${\cal D}$ and a target node $T$ as its input and promptly returns a rough estimation of $\textbf{PC}_T$, hence the term ``weak'' PC learner.  In this study, we use FDR-IAPC because it aims at controlling the expected proportion of false discoveries (i.e., false positive nodes in $\textbf{PC}_T$) among all the discoveries made. FDR-IAPC is a straightforward extension of the algorithm IAMBFDR developed by Jose Pe\~na in \citep{Pena08}, which is itself a modification of the incremental association Markov boundary algorithm (IAMB) \citep{Tsamardinos03}, to control the expected proportion of false discoveries (i.e., false positive nodes) in the estimated Markov boundary. FDR-IAPC simply removes, at lines 3-6, the spouses $\textbf{SP}_T$ from the estimated Markov boundary $\textbf{MB}_T$ output by IAMBFDR at line 1, and returns $\textbf{PC}_T$ assuming the faithfulness condition.

The subroutines DE-PCS (Algorithm \ref{DE-PCS}) and DE-SPS (Algorithm \ref{DE-SPS}) search a superset of $\textbf{PC}_T$ and $\textbf{SP}_T$ respectively with a severe restriction on the maximum conditioning size ($|\textbf{Z}|<= 1$ in DE-PCS and $|\textbf{Z}|<= 2$ in DE-SPS) in order to significantly increase the reliability of the tests. The variable filtering has two advantages : i) it allows HPC to scale to hundreds of thousands of variables by restricting the search to a subset of relevant variables, and ii) it eliminates many true or approximate deterministic relationships that produce many false negative errors in the output of the algorithm, as explained in \citep{Morais10a,Morais10b}. DE-SPS works in two steps. First, a growing phase (lines 4-8)  adds the variables that are d-separated from the target but still remain associated with the target when conditioned on another variable from $\textbf{PCS}_T$.  The shrinking phase (lines 9-16) discards irrelevant variables that are ancestors or descendants of a target's spouse.  Pruning such irrelevant variables speeds up HPC.


\begin{algorithm}
  \begin{small}
  \caption{\emph{HPC}}
  \label{HPC}
  \begin{algorithmic}[1]

    \REQUIRE
    $T$: target; ${\cal D}$: data set; \textbf{U}: set of variables\\
    \ENSURE
    $\textbf{PC}_T$: Parents and Children of $T$\\
    \emph{}\\

    \STATE $[\textbf{PCS}_T,\textbf{dSep}] \leftarrow \emph{DE-PCS}(T,{\cal D},\textbf{U})$
    \STATE $\textbf{SPS}_T \leftarrow \emph{DE-SPS}(T,{\cal D},\textbf{U},\textbf{PCS}_T,\textbf{dSep})$

    \STATE $\textbf{PC}_T \leftarrow \emph{FDR-IAPC}(T,{\cal D},(T \cup \textbf{PCS}_T \cup \textbf{SPS}_T))$\\
    \FORALL {$X \in \textbf{PCS}_T\setminus\textbf{PC}_T$}
    \IF {$T \in \emph{FDR-IAPC}(X,{\cal D},(T \cup \textbf{PCS}_T \cup \textbf{SPS}_T))$}
    \STATE $\textbf{PC}_T \leftarrow \textbf{PC}_T \cup{X}$
    \ENDIF
    \ENDFOR

  \end{algorithmic}
  \end{small}
\end{algorithm}

\begin{algorithm}
  \begin{small}
  \caption{\emph{FDR-IAPC}}
  \label{FDR-IAPC}
  \begin{algorithmic}[1]

    \REQUIRE
    $T$: target; \emph{D}: data set; $\textbf{U}$: set of variables \\
    \ENSURE
    $\textbf{PC}_T$: Parents and children of $T$;\\
    \emph{}\\

    \emph{\textbf{*} Learn the Markov boundary of} $T$ \\
    \STATE $\textbf{MB}_T \leftarrow \emph{IAMBFDR}(X,\cal{D},\textbf{U})$\\

    \emph{\textbf{*} Remove spouses of $T$ from} $\textbf{MB}_T$\\
    \STATE $\textbf{PC}_T \leftarrow \textbf{MB}_T$
    \FORALL {$X \in \textbf{MB}_T$}
      \IF {$\exists \textbf{Z} \subseteq (\textbf{MB}_T \setminus X)$ such that $T \perp X \mid \textbf{Z}$}
      \STATE $\textbf{PC}_T \leftarrow \textbf{PC}_T\setminus X$
      \ENDIF
    \ENDFOR

  \end{algorithmic}
  \end{small}
\end{algorithm}

\begin{algorithm}
  \begin{small}
  \caption{\emph{DE-PCS}}
  \label{DE-PCS}
  \begin{algorithmic}[1]

    \REQUIRE
    $T$: target; ${\cal D}$: data set; $\textbf{U}$: set of variables;\\
    \ENSURE
    $\textbf{PCS}_T$: parents and children superset of $T$;
    \textbf{dSep}: d-separating sets;\\
    \emph{}\\

    \textbf{Phase I:}\emph{ Remove X if $T\perp X$}\\

    \STATE $\textbf{PCS}_T \leftarrow \textbf{U}\setminus {T}$

    \FORALL {$X\in\textbf{PCS}_T$}
    \IF {$(T \perp X)$}
    \STATE $\textbf{PCS}_T \leftarrow \textbf{PCS}_T\setminus \emph{X}$
    \STATE $\textbf{dSep}(X) \leftarrow \emptyset$
    \ENDIF
    \ENDFOR \\
    \emph{}\\

    \textbf{Phase II:}\emph{ Remove X if $T\perp X|Y$}\\
    \FORALL {$X\in\textbf{PCS}_T$}
    \FORALL {$Y\in\textbf{PCS}_T\setminus X$}
    \IF {$(T \perp X \mid Y)$}
    \STATE $\textbf{PCS}_T \leftarrow \textbf{PCS}_T\setminus \emph{X}$
    \STATE $\textbf{dSep}(X) \leftarrow Y$
    \STATE \textbf{break loop FOR}
    \ENDIF
    \ENDFOR
    \ENDFOR
  \end{algorithmic}
  \end{small}
\end{algorithm}

\begin{algorithm}
  \begin{small}
  \caption{\emph{DE-SPS}}
  \label{DE-SPS}
  \begin{algorithmic}[1]

    \REQUIRE
    $T$: target; ${\cal D}$: data set; \textbf{U}: the set of variables;
    $\textbf{PCS}_T$: parents and children superset of $T$;
    \textbf{dSep}: d-separating sets;\\
    \ENSURE
    $\textbf{SPS}_T$: Superset of the spouses of $T$;

    \emph{}\\

    \STATE $\textbf{SPS}_T \leftarrow \emptyset$\\
    \FORALL {$X \in \textbf{PCS}_T$}
    \STATE $\textbf{SPS}_{T}^{X} \leftarrow \emptyset$\\
    \FORALL {$Y \in \textbf{U}\setminus \{T\cup \textbf{PCS}_T\}$}
    \IF {$(T \not\perp Y|\textbf{dSep}(Y)\cup{X})$}
    \STATE $\textbf{SPS}_{T}^{X} \leftarrow \textbf{SPS}_{T}^{X}\cup{Y}$
    \ENDIF
    \ENDFOR

    \FORALL {$Y \in \textbf{SPS}_{T}^{X}$}
    \FORALL {$Z \in \textbf{SPS}_{T}^{X} \setminus Y$}
    \IF {$(T \perp Y|X\cup Z)$}
    \STATE $\textbf{SPS}_{T}^{X} \leftarrow \textbf{SPS}_{T}^{X}\setminus Y$
    \STATE \textbf{break loop FOR}
    \ENDIF
    \ENDFOR
    \ENDFOR
    \STATE $\textbf{SPS}_T \leftarrow \textbf{SPS}_T\cup \textbf{SPS}_{T}^{X}$
    \ENDFOR

  \end{algorithmic}
  \end{small}
\end{algorithm}

\begin{algorithm}
  \begin{small}
  \caption{\emph{Hybrid HPC}}
  \label{H2PC}
  \begin{algorithmic}[1]

    \REQUIRE
    ${\cal D}$: data set; \textbf{U}: set of variables\\
    \ENSURE
    A DAG ${\cal G}$ on the variables \textbf{U} \\
    \emph{}\\

    \FORALL {pair of nodes $X,Y \in \textbf{U}$}
    \STATE Add X in $\textbf{PC}_Y$ and Add Y in $\textbf{PC}_X$ if $X \in HPC(Y)$ and $Y \in HPC(X)$
    \ENDFOR

    \STATE {Starting from an empty graph, perform greedy hill-climbing with operators \textit{add-edge, delete-edge, reverse-edge}.
    Only try operator \textit{add-edge} $X \rightarrow Y$ if $Y \in \textbf{PC}_X$}

  \end{algorithmic}
  \end{small}
\end{algorithm}

\subsection{Hybrid HPC (H2PC)}

In this section, we discuss the SS phase. The following discussion
draws strongly on \citep{Tsamardinos06} as the SS phase in Hybrid HPC
and MMHC are exactly the same. The idea of constraining the search to
improve time-efficiency first appeared in the Sparse Candidate
algorithm \citep{Friedman99}. It results in efficiency improvements over
the (unconstrained) greedy search. All recent hybrid algorithms build
on this idea, but employ a sound algorithm for identifying the
candidate parent sets. The Hybrid HPC first identifies the parents and
children set of each variable, then performs a greedy hill-climbing
search in the space of BN. The search begins with an empty graph. The
edge addition, deletion, or direction reversal that leads to the
largest increase in score (the BDeu score with uniform prior was used) is taken and the
search continues in a similar fashion recursively. The important
difference from standard greedy search is that the search is
constrained to only consider adding an edge if it was discovered by HPC
in the first phase. We extend the greedy search with a TABU list
\citep{Friedman99}. The list keeps the last 100 structures explored.
Instead of applying the best local change, the best local change that
results in a structure not on the list is performed in an attempt to
escape local maxima. When 15 changes occur without an increase in the
maximum score ever encountered during search, the algorithm terminates.
The overall best scoring structure is then returned. Clearly, the more
false positives the heuristic allows to enter candidate PC set, the
more computational burden is imposed in the SS phase.

\section{Experimental validation on synthetic data}
\label{sec:experiments}

Before we proceed to the experiments on real-world multi-label data with H2PC, we first conduct an experimental comparison of H2PC against MMHC on synthetic data sets sampled from eight well-known benchmarks with various data sizes in order to gauge the practical relevance of the H2PC. These BNs that have been previously used as benchmarks for BN learning algorithms (see Table \ref{tab:datasets} for details).  Results are reported in terms of various performance indicators to investigate how well the network generalizes to new data and how well the learned dependence structure matches the true structure of the benchmark network. We implemented H2PC in \textit{R} \citep{R} and integrated the code into the \textit{bnlearn} package from \citep{Scutari10}. MMHC was implemented by Marco Scutari in \textit{bnlearn}. The source code of H2PC as well as all data sets used for the empirical tests are publicly available \footnote{\url{https://github.com/madbix/bnlearn-clone-3.4}}. The threshold considered for the type I error of the test is $0.05$. Our experiments were carried out on PC with Intel(R) Core(TM) i5-3470M CPU @3.20 GHz 4Go RAM running under Linux 64 bits.

\begin{table}
  \caption{Description of the BN benchmarks used in the experiments.}
  \label{tab:datasets}
  \vspace{\tabcapsep}
  \footnotesize
  \centering
  \setlength{\tabcolsep}{3pt}
  \begin{tabular}{cccccc}
  \toprule
  \multicolumn{1}{c}{\multirow{2}{*}{network}} & \multicolumn{1}{c}{\multirow{2}{*}{\# var.}} & \multicolumn{1}{c}{\multirow{2}{*}{\# edges}} & \multicolumn{1}{c}{max. degree} & \multicolumn{1}{c}{domain} & \multicolumn{1}{c}{min/med/max} \\
                                              &                                                  &                                                  & \multicolumn{1}{c}{in/out}      &  \multicolumn{1}{c}{range}          & \multicolumn{1}{c}{PC set size} \\
  \midrule
  \multicolumn{1}{c}{child}      & \multicolumn{1}{c}{20}   & \multicolumn{1}{c}{25}   & \multicolumn{1}{c}{2/7}  & \multicolumn{1}{c}{2-6}   & \multicolumn{1}{c}{1/2/8} \\
  \multicolumn{1}{c}{insurance}  & \multicolumn{1}{c}{27}   & \multicolumn{1}{c}{52}   & \multicolumn{1}{c}{3/7}  & \multicolumn{1}{c}{2-5}   & \multicolumn{1}{c}{1/3/9} \\
  \multicolumn{1}{c}{mildew}     & \multicolumn{1}{c}{35}   & \multicolumn{1}{c}{46}   & \multicolumn{1}{c}{3/3}  & \multicolumn{1}{c}{3-100} & \multicolumn{1}{c}{1/2/5} \\
  \multicolumn{1}{c}{alarm}      & \multicolumn{1}{c}{37}   & \multicolumn{1}{c}{46}   & \multicolumn{1}{c}{4/5}  & \multicolumn{1}{c}{2-4}   & \multicolumn{1}{c}{1/2/6} \\
  \multicolumn{1}{c}{hailfinder} & \multicolumn{1}{c}{56}   & \multicolumn{1}{c}{66}   & \multicolumn{1}{c}{4/16} & \multicolumn{1}{c}{2-11}  & \multicolumn{1}{c}{1/1.5/17} \\
  \multicolumn{1}{c}{munin1}     & \multicolumn{1}{c}{186}  & \multicolumn{1}{c}{273} & \multicolumn{1}{c}{3/15} & \multicolumn{1}{c}{2-21}  & \multicolumn{1}{c}{1/3/15} \\
  \multicolumn{1}{c}{pigs}       & \multicolumn{1}{c}{441}  & \multicolumn{1}{c}{592}  & \multicolumn{1}{c}{2/39} & \multicolumn{1}{c}{3-3}   & \multicolumn{1}{c}{1/2/41} \\
  \multicolumn{1}{c}{link}       & \multicolumn{1}{c}{724}  & \multicolumn{1}{c}{1125} & \multicolumn{1}{c}{3/14} & \multicolumn{1}{c}{2-4}   & \multicolumn{1}{c}{0/2/17}  \\
  \bottomrule
  \end{tabular}
\end{table}

We do not claim that those data sets resemble real-world problems, however, they make
it possible to compare the outputs of the algorithms with the known
structure. All BN benchmarks (structure and probability tables) were
downloaded from the \textit{bnlearn} repository\footnote{
\emph{http://www.bnlearn.com/bnrepository}}. Ten sample sizes have
been considered: 50, 100, 200, 500, 1000, 2000, 5000, 10000, 20000 and 50000. All experiments are
repeated 10 times for each sample size and each BN. We investigate the
behavior of both algorithms using the same parametric tests as a
reference.

\begin{figure*}
\begin{center}$
  \begin{array}{cc} 
  \includegraphics[clip, trim = 0.05in 0.15in 0.1in 0.05in, scale=0.9]{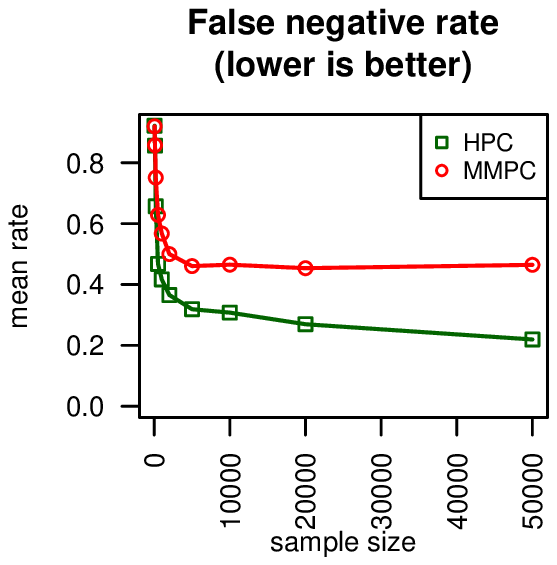} &
  \includegraphics[clip, trim = 0.05in 0.15in 0.1in 0.05in, scale=0.9]{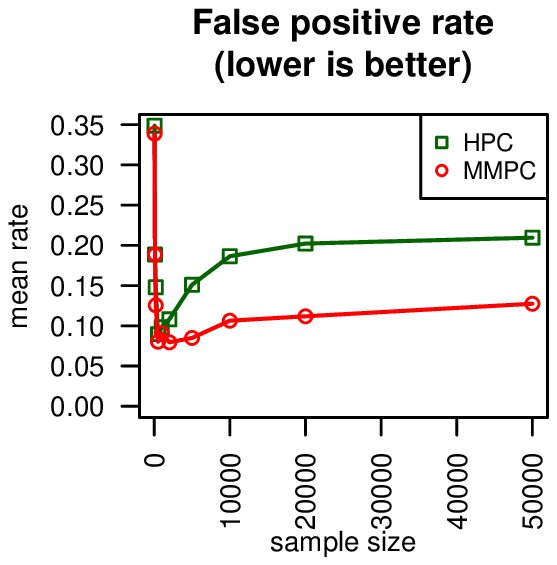} \\
  \includegraphics[clip, trim = 0.05in 0.15in 0.1in 0.05in, scale=0.9]{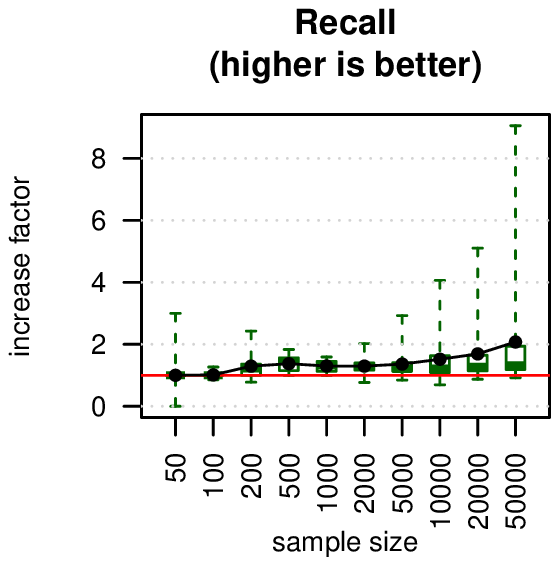} &
  \includegraphics[clip, trim = 0.05in 0.15in 0.1in 0.05in, scale=0.9]{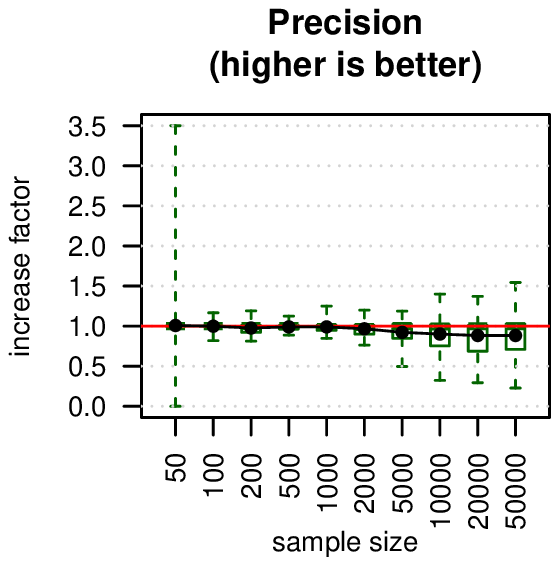} \\
  \includegraphics[clip, trim = 0.05in 0.15in 0.1in 0.05in, scale=0.9]{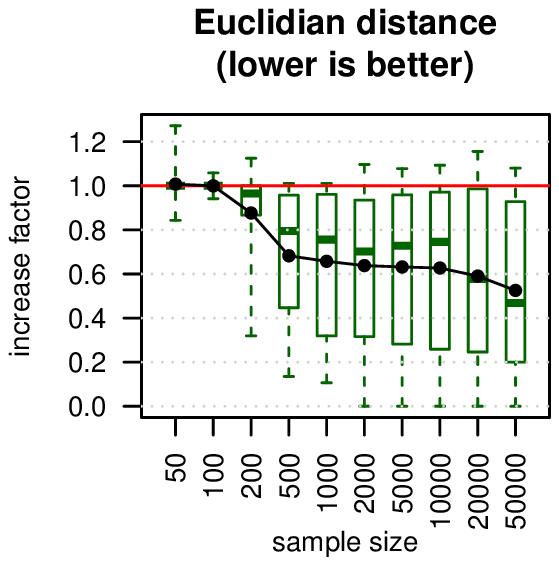} &
  \includegraphics[clip, trim = 0.05in 0.15in 0.1in 0.05in, scale=0.9]{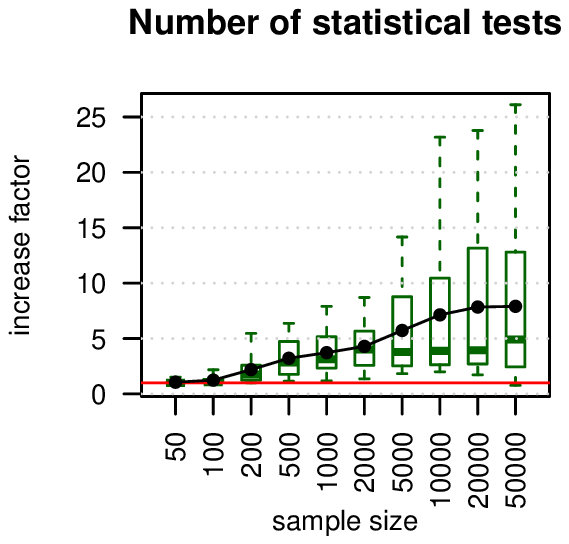} \\
  \end{array}$
  \end{center}
  \caption{Quality of the skeleton obtained with HPC over that obtained with MMPC (after the CB phase). The 2 first figures present mean values aggregated over the 8 benchmarks. The 4 last figures present increase factors of HPC / MMPC, with the median, quartile, and most extreme values (green boxplots), along with the mean value (black line).}
  \label{fig:h2pc-skeleton-scores}
\end{figure*}

\subsection{Performance indicators}

We first investigate the quality of the skeleton returned by H2PC
during the CB phase. To this end, we measure the false positive edge
ratio, the precision (i.e., the number of true positive edges in the
output divided by the number of edges in the output), the recall (i.e.,
the number of true positive edges divided the true number of edges) and
a combination of precision and recall defined as $\sqrt{(1-precision)^2
+ (1-recall)^2}$, to measure the Euclidean distance from perfect
precision and recall, as proposed in \citep{Pen07}. Second, to assess
the quality of the final DAG output at the end of the SS phase, we
report the five performance indicators \citep{Scutari12} described below:

\begin{itemize}
  \item the posterior density of the network for the data it was
  learned from, as a   measure of goodness of fit. It is known as the
  Bayesian Dirichlet equivalent score (BDeu) from \citep{Heckerman95,Bun91} and
  has a single parameter, the equivalent sample size, which can be
  thought of as the size of an imaginary sample supporting the prior
  distribution. The equivalent sample size was set to 10 as suggested
  in \citep{Koller09};
  \item the BIC score \citep{Schwarz78} of the network for the data it
  was learned from, again as a measure of goodness of fit;
  \item the posterior density of the network for a new data set, as a
  measure of how well the network generalizes to new data;
  \item  the BIC score of the network for a new data set, again as a
  measure of how well the network generalizes to new data;
  \item the Structural Hamming Distance (SHD) between the learned and
  the true structure of the network, as a measure of the quality of the
  learned dependence structure. The SHD between two PDAGs is defined as
  the number of the following operators required to make the PDAGs
  match: add or delete an undirected edge, and add, remove, or reverse
  the orientation of an edge.
\end{itemize}

For each data set sampled from the true probability distribution of the benchmark, we
first learn a network structure with the H2PC and MMHC and then we compute the
relevant performance indicators for each pair of network structures.
The data set used to assess how well the network generalizes to new data
is generated again from the true probability structure of the benchmark
networks and contains 50000 observations.

Notice that using the BDeu score as a metric of
reconstruction quality has the following two problems. First, the score
corresponds to the a posteriori probability of a network only under
certain conditions (e.g., a Dirichlet distribution of the hyper
parameters); it is unknown to what degree these assumptions hold in
distributions encountered in practice. Second, the score is highly sensitive to the
equivalent sample size (set to 10 in our experiments) and depends on the network priors used.
Since, typically, the same arbitrary value of this parameter is used both during learning and
for scoring the learned network, the metric favors algorithms that use
the BDeu score for learning. In fact, the BDeu score does not rely on
the structure of the original, gold standard network at all; instead it
employs several assumptions to score the networks. For those reasons,
in addition to the score we also report the BIC score and the SHD
metric.

\subsection{Results}

In Figure \ref{fig:h2pc-skeleton-scores}, we report the quality of the skeleton obtained with HPC over that obtained with MMPC (before the SS phase) as a function of the sample size. Results for each benchmark are not shown here in detail due to space restrictions. For sake of conciseness, the performance values are averaged over the 8 benchmarks depicted in Table~\ref{tab:datasets}. The increase factor for a given performance indicator is expressed as the ratio of the performance value obtained with HPC over that obtained with MMPC (the gold standard). Note that for some indicators, an increase is actually not an improvement but is worse (e.g., false positive rate, Euclidean distance). For clarity, we mention explicitly on the subplots whether an increase factor $>1$ should be interpreted as an improvement or not. Regarding the quality of the superstructure, the advantages of HPC against MMPC are noticeable. As observed, HPC consistently increases the recall and reduces the rate of false negative edges. As expected this benefit comes at a little expense in terms of false positive edges. HPC also improves the Euclidean distance from perfect precision and recall on all benchmarks, while increasing the number of independence tests and thus the running time in the CB phase (see number of statistical tests). It is worth noting that HPC is capable of reducing by 50\% the Euclidean distance with $50 000$ samples (lower left plot). These results are very much in line with other experiments presented in \citep{Morais10b,Villanueva12}.

\begin{figure*}
  \begin{center}$
  \begin{array}{cc} 
  \includegraphics[clip, trim = 0.05in 0.15in 0.1in 0.05in, scale=0.9]{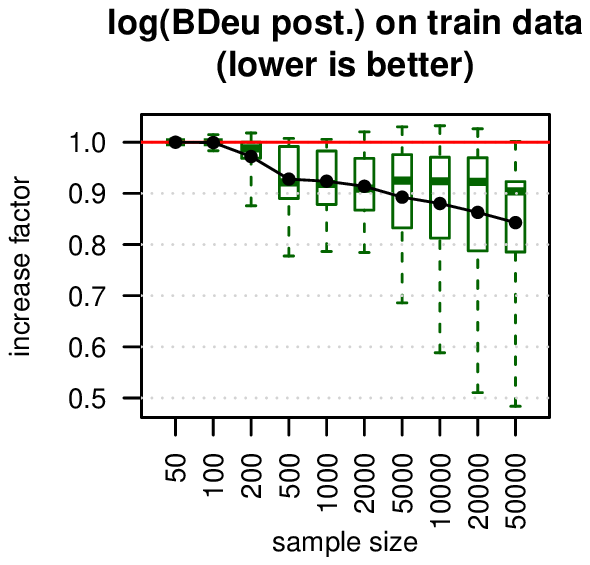} &
  \includegraphics[clip, trim = 0.05in 0.15in 0.1in 0.05in, scale=0.9]{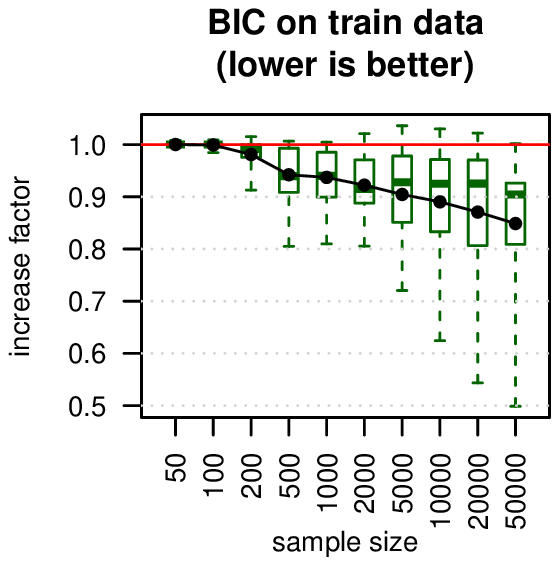}\\
  \includegraphics[clip, trim = 0.05in 0.15in 0.1in 0.05in, scale=0.9]{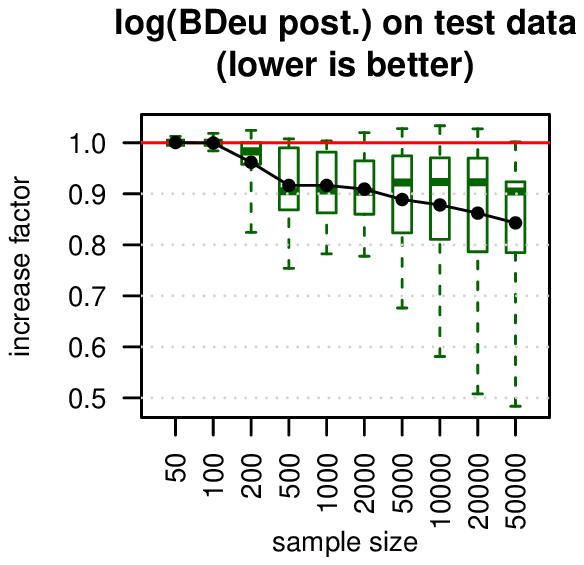} &
  \includegraphics[clip, trim = 0.05in 0.15in 0.1in 0.05in, scale=0.9]{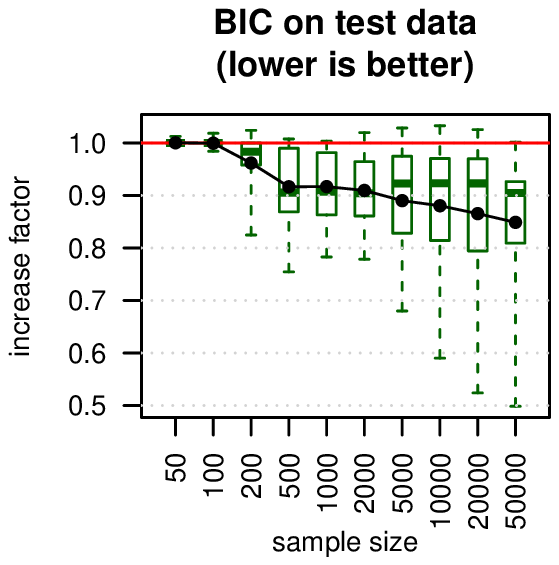}\\
  \includegraphics[clip, trim = 0.05in 0.15in 0.1in 0.05in, scale=0.9]{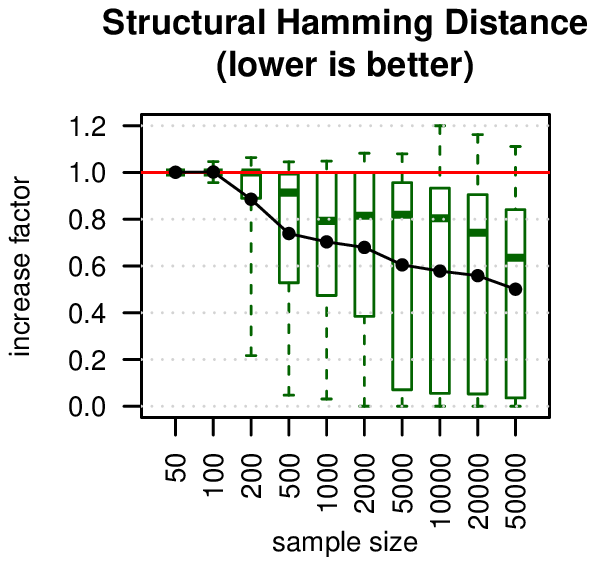} &
  \includegraphics[clip, trim = 0.05in 0.15in 0.1in 0.05in, scale=0.9]{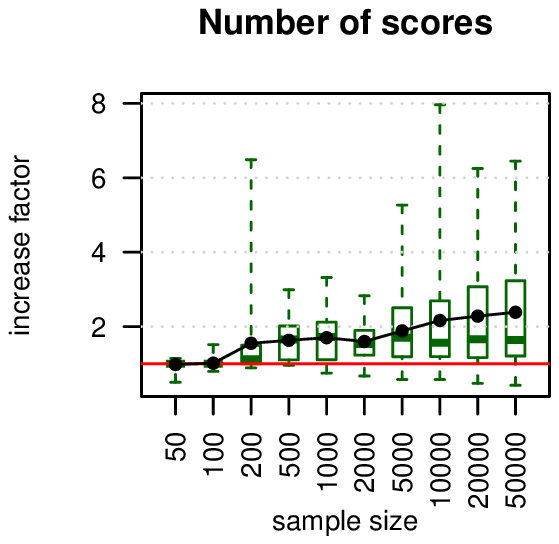} \\
  \end{array}$
  \end{center}
  \caption{Quality of the final DAG obtained with H2PC over that obtained with MMHC (after the SS phase). The 6 figures present increase factors of HPC / MMPC, with the median, quartile, and most extreme values (green boxplots), along with the mean value (black line).}
  \label{fig:h2pc-scores}
\end{figure*}

\begin{table*}
  \caption{Total running time ratio $R$ (H2PC/MMHC). White cells indicate a ratio $R<1$ (in favor of H2PC), while shaded cells indicate a ratio $R>1$ (in favor of MMHC). The darker, the larger the ratio.}
  \label{tab:total.time}
  \vspace{\tabcapsep}
  \footnotesize
  \centering
  \begin{tabular}{crrrrrrrrrr}
    \toprule
    \multicolumn{1}{c}{\multirow{2}{*}{Network}} & \multicolumn{10}{c}{Sample Size} \\
    \cmidrule(l){2-11}
    \multicolumn{1}{c}{} & \multicolumn{1}{c}{50} & \multicolumn{1}{c}{100} & \multicolumn{1}{c}{200} & \multicolumn{1}{c}{500} & \multicolumn{1}{c}{1000} & \multicolumn{1}{c}{2000} & \multicolumn{1}{c}{5000} & \multicolumn{1}{c}{10000} & \multicolumn{1}{c}{20000} & \multicolumn{1}{c}{50000} \\
    \midrule
    \multirow{2}{*}{child} & 0.94 & 0.87 & \cellcolor{n1} 1.14 & \cellcolor{n1} 1.99 & \cellcolor{n1} 2.26 & \cellcolor{n1} 2.12 & \cellcolor{n1} 2.36 & \cellcolor{n1} 2.58 & \cellcolor{n1} 1.75 & \cellcolor{n1} 1.78 \\
     & \(\pm\)0.1 & \(\pm\)0.1 & \(\pm\)0.1 & \(\pm\)0.2 & \(\pm\)0.1 & \(\pm\)0.2 & \(\pm\)0.4 & \(\pm\)0.3 & \(\pm\)0.6 & \(\pm\)0.5 \\[.1cm]
    \multirow{2}{*}{insurance} & 0.96 & \cellcolor{n1} 1.09 & \cellcolor{n1} 1.56 & \cellcolor{n1} 2.93 & \cellcolor{n1} 3.06 & \cellcolor{n1} 3.48 & \cellcolor{n1} 3.69 & \cellcolor{n1} 4.10 & \cellcolor{n1} 3.76 & \cellcolor{n1} 3.75 \\
     & \(\pm\)0.1 & \(\pm\)0.1 & \(\pm\)0.1 & \(\pm\)0.2 & \(\pm\)0.3 & \(\pm\)0.4 & \(\pm\)0.3 & \(\pm\)0.4 & \(\pm\)0.6 & \(\pm\)0.5 \\[.1cm]
    \multirow{2}{*}{mildew} & 0.77 & 0.80 & 0.79 & 0.94 & \cellcolor{n1} 1.01 & \cellcolor{n1} 1.23 & \cellcolor{n1} 1.74 & \cellcolor{n1} 2.14 & \cellcolor{n1} 3.26 & \cellcolor{n2} 6.20 \\
     & \(\pm\)0.1 & \(\pm\)0.1 & \(\pm\)0.1 & \(\pm\)0.1 & \(\pm\)0.1 & \(\pm\)0.1 & \(\pm\)0.2 & \(\pm\)0.2 & \(\pm\)0.6 & \(\pm\)1.0 \\[.1cm]
    \multirow{2}{*}{alarm} & 0.88 & \cellcolor{n1} 1.11 & \cellcolor{n1} 1.75 & \cellcolor{n1} 2.43 & \cellcolor{n1} 2.55 & \cellcolor{n1} 2.71 & \cellcolor{n1} 2.65 & \cellcolor{n1} 2.80 & \cellcolor{n1} 2.49 & \cellcolor{n1} 2.18 \\
     & \(\pm\)0.1 & \(\pm\)0.1 & \(\pm\)0.1 & \(\pm\)0.1 & \(\pm\)0.1 & \(\pm\)0.1 & \(\pm\)0.2 & \(\pm\)0.2 & \(\pm\)0.3 & \(\pm\)0.6 \\[.1cm]
    \multirow{2}{*}{hailfinder} & 0.85 & 0.85 & \cellcolor{n1} 1.40 & \cellcolor{n1} 1.69 & \cellcolor{n1} 1.83 & \cellcolor{n1} 2.06 & \cellcolor{n1} 2.13 & \cellcolor{n1} 2.12 & \cellcolor{n1} 1.95 & \cellcolor{n1} 1.96 \\
     & \(\pm\)0.1 & \(\pm\)0.1 & \(\pm\)0.1 & \(\pm\)0.1 & \(\pm\)0.1 & \(\pm\)0.1 & \(\pm\)0.1 & \(\pm\)0.2 & \(\pm\)0.2 & \(\pm\)0.6 \\[.1cm]
    \multirow{2}{*}{munin1} & 0.77 & 0.85 & 0.93 & \cellcolor{n1} 1.35 & \cellcolor{n1} 2.11 & \cellcolor{n1} 4.30 & \cellcolor{n3} 12.92 & \cellcolor{n3} 23.32 & \cellcolor{n3} 24.95 & \cellcolor{n3} 24.76 \\
     & \(\pm\)0.0 & \(\pm\)0.0 & \(\pm\)0.0 & \(\pm\)0.0 & \(\pm\)0.0 & \(\pm\)0.2 & \(\pm\)0.7 & \(\pm\)2.6 & \(\pm\)5.1 & \(\pm\)6.7 \\[.1cm]
    \multirow{2}{*}{pigs} & 0.80 & 0.80 & \cellcolor{n1} 4.55 & \cellcolor{n1} 4.71 & \cellcolor{n2} 5.00 & \cellcolor{n2} 5.62 & \cellcolor{n2} 7.63 & \cellcolor{n3} 11.10 & \cellcolor{n3} 14.02 & \cellcolor{n3} 11.74 \\
     & \(\pm\)0.0 & \(\pm\)0.0 & \(\pm\)0.1 & \(\pm\)0.1 & \(\pm\)0.2 & \(\pm\)0.2 & \(\pm\)0.3 & \(\pm\)0.6 & \(\pm\)1.7 & \(\pm\)3.2 \\[.1cm]
    \multirow{2}{*}{link} & \cellcolor{n1} 1.16 & \cellcolor{n1} 1.93 & \cellcolor{n1} 2.76 & \cellcolor{n2} 5.55 & \cellcolor{n2} 7.04 & \cellcolor{n2} 8.19 & \cellcolor{n3} 10.00 & \cellcolor{n3} 13.87 & \cellcolor{n3} 15.32 & \cellcolor{n3} 24.74 \\
     & \(\pm\)0.0 & \(\pm\)0.0 & \(\pm\)0.0 & \(\pm\)0.1 & \(\pm\)0.2 & \(\pm\)0.2 & \(\pm\)0.3 & \(\pm\)0.4 & \(\pm\)2.5 & \(\pm\)4.2 \\
    \midrule
    \multirow{2}{*}{all} & 0.89 & \cellcolor{n1} 1.04 & \cellcolor{n1} 1.86 & \cellcolor{n1} 2.70 & \cellcolor{n1} 3.11 & \cellcolor{n1} 3.71 & \cellcolor{n2} 5.39 & \cellcolor{n2} 7.75 & \cellcolor{n2} 8.44 & \cellcolor{n2} 9.64 \\
     & \(\pm\)0.1 & \(\pm\)0.1 & \(\pm\)1.2 & \(\pm\)1.5 & \(\pm\)1.9 & \(\pm\)2.2 & \(\pm\)4.0 & \(\pm\)7.3 & \(\pm\)8.4 & \(\pm\)9.7 \\
    \bottomrule
  \end{tabular}
\end{table*}

In Figure \ref{fig:h2pc-scores}, we report the quality of the final DAG obtained with H2PC over that obtained with MMHC (after the SS phase) as a function of the sample size. Regarding BDeu and BIC on both training and test data,  the improvements are noteworthy. The results in terms of goodness of fit to training data and new data using H2PC clearly dominate those obtained using MMHC, whatever the sample size considered, hence its ability to generalize better. Regarding the quality of the network structure itself (i.e., how close is the DAG to the true dependence structure of the data), this is pretty much a dead heat between the 2 algorithms on small sample sizes (i.e., 50 and 100), however we found H2PC to perform significantly better on larger sample sizes. The SHD increase factor decays rapidly (lower is better) as the sample size increases (lower left plot). For 50 000 samples, the SHD is on average only 50\% that of MMHC. Regarding the computational burden involved, we may observe from Table \ref{tab:total.time} that H2PC has a little computational overhead compared to MMHC. The running time increase factor grows somewhat linearly with the sample size. With $50 000$ samples, H2PC is approximately $10$ times slower on average than MMHC. This is mainly due to the computational expense incurred in obtaining larger PC sets with HPC, compared to MMPC.


%
%
%
%
%
%
%
%

\section{Application to multi-label learning}

In this section, we address the problem of multi-label learning with H2PC. MLC is a challenging problem in many real-world application domains, where each instance can be assigned simultaneously to multiple binary labels \citep{Dembczynski2012,Read2009,Madjarov2012,Kocev2007,Tsoumakas2010}. Formally, learning from multi-label examples amounts to  finding a mapping from a space of features to a space of labels. We shall assume throughout that $\mathbf{X}$ (a random vector in $\mathbb{R}^d$) is the feature set, $\mathbf{Y}$ (a random vector in $\{0,1\}^n$) is the label set, $\mathbf{U} = \mathbf{X} \cup \mathbf{Y}$ and $p$ a probability distribution defined over $\mathbf{U}$.  Given a multi-label training set ${\mathcal D}$, the goal of multi-label learning is to find a function which is able to map any unseen example to its proper set of labels. From the Bayesian point of view, this problem amounts to modeling the conditional joint distribution $p(\mathbf{Y} \mid \mathbf{X})$.

\subsection{Related work}

This MLC problem may be tackled in various ways \citep{Luaces2012,Alvares-Cherman2011,Read2009,Blockeel1998,Kocev2007}. Each of these approaches is supposed to capture - to some extent - the relationships between labels. The two most straightforward meta-learning methods \citep{Madjarov2012} are: Binary Relevance (BR)~\citep{Luaces2012}  and Label Powerset (LP)~\citep{Tsoumakas2007,Tsoumakas2010}. Both methods can be regarded as opposite in the sense that BR does consider each label independently, while LP considers the whole label set at once (one multi-class problem). An important question remains: what shall we capture from the statistical relationships between labels exactly to solve the multi-label classification problem? The problem attracted a great deal of interest \citep{Dembczynski2012,Zhang10}. It is well beyond the scope and purpose of this paper to delve deeper into these approaches, we point the reader to \citep{Dembczynski2012,Tsoumakas2010} for a review.  The second fundamental problem that we wish to address involves finding an optimal feature subset selection of a label set, w.r.t an Information Theory criterion \cite{Koller96}. As in the single-label case, multi-label feature selection has been studied recently and has encountered some success \citep{Gu2011,Spolaor2013}.

\subsection{Label powerset decomposition}

We shall first introduce the concept of label powerset that will play a pivotal role in the factorization of the conditional distribution $p(\mathbf{Y} \mid \mathbf{X})$.

\begin{defn} \rm
$\mathbf{Y}_{LP} \subseteq \mathbf{Y}$ is called a label powerset \textit{iff} $\mathbf{Y}_{LP} \indep \mathbf{Y} \setminus \mathbf{Y}_{LP} \mid \mathbf{X}$. Additionally, $\mathbf{Y}_{LP}$ is said minimal if it is non empty and has no label powerset as proper subset.
\end{defn}

Let $\mathbf{LP}$ denote the set of all powersets defined over $\mathbf{U}$, and $min\mathbf{LP}$ the set of all minimal label powersets. It is easily shown $\{\mathbf{Y},\emptyset\} \subseteq \mathbf{LP}$. The key idea behind label powersets is the decomposition of the conditional distribution of the labels into a product of factors
\begin{align*}
p(\mathbf{Y} \mid \mathbf{X}) = \prod_{j=1}^L p(\mathbf{Y}_{LP_j} \mid \mathbf{X}) = \prod_{j=1}^L p(\mathbf{Y}_{LP_j} \mid \mathbf{M}_{LP_j})
\end{align*}

where $\{\mathbf{Y}_{LP_1}, \dots, \mathbf{Y}_{LP_L}\}$ is a partition of label powersets and $\mathbf{M}_{LP_j}$ is a Markov blanket of $\mathbf{Y}_{LP_j}$ in $\mathbf{X}$. From the above definition, we have $\mathbf{Y}_{LP_i} \indep \mathbf{Y}_{LP_j} \mid \mathbf X \text{, } \forall i \neq j$.

In the framework of MLC, one can consider a multitude of loss functions. In this study, we focus on a non label-wise decomposable loss function called the subset $0/1$ loss which generalizes the well-known $0/1$ loss from the conventional to the multi-label setting. The risk-minimizing prediction for subset $0/1$ loss is simply given by the mode of the distribution \citep{Dembczynski2012}. Therefore, we seek a factorization into a product of minimal factors in order to facilitate the estimation of the mode of the conditional distribution
(also called the most probable explanation (MPE)):

\begin{align*}
  \max_\mathbf{y} p(\mathbf{y}\mid\mathbf{x}) = \prod_{j=1}^L \max_{\mathbf{y}_{LP_j}}  p(\mathbf{y}_{LP_j}\mid\mathbf{x})
	= \prod_{j=1}^L \max_{\mathbf{y}_{LP_j}}  p(\mathbf{y}_{LP_j}\mid\mathbf{m}_{LP_j})
\end{align*}

The next section aims to obtain theoretical results for the characterization of the minimal label powersets $\mathbf{Y}_{LP_j}$ and their Markov boundaries $\mathbf{M}_{LP_j}$ from a DAG, in order to be able to estimate the MPE more effectively.

\subsection{Label powerset characterization}

In this section, we show that minimal label powersets can be depicted graphically when $p$ is faithful to a DAG,

\begin{thm}
  \label{th:dSep_mlp}
  Suppose $p$ is faithful to a DAG $\cal{G}$. Then, $Y_i$ and $Y_j$ belong to the same minimal label powerset if and only if there exists an undirected path in $\cal{G}$ between nodes $Y_i$ and $Y_j$ in $\mathbf{Y}$ such that all intermediate nodes $Z$ are either (i) $Z \in \mathbf{Y}$, or (ii) $Z \in \mathbf{X}$ and $Z$ has two parents in $\mathbf{Y}$ (i.e. a collider  of the form $Y_p\rightarrow X \leftarrow Y_q$).
\end{thm}

The proof is given in the Appendix. We shall now address the following questions: What is the Markov boundary in $\mathbf{X}$ of a minimal label powerset? Answering this question for general distributions is not trivial. We establish a useful relation between a label powerset and its Markov boundary in $\mathbf{X}$ when $p$ is faithful to a DAG,

\begin{thm}
  \label{th:mb-lp-dsep}
  Suppose $p$ is faithful to a DAG $\cal{G}$. Let $\mathbf{Y}_{LP} = \{Y_1, Y_2, \dots, Y_n\}$ be a label powerset. Then, its Markov boundary $\mathbf{M}$ in $\mathbf{U}$ is also its Markov boundary in $\mathbf{X}$, and is given in $\cal{G}$ by $\mathbf{M}=\bigcup_{j = 1}^{n} \{\mathbf{PC}_{Y_j} \cup \mathbf{SP}_{Y_j}\} \setminus \mathbf{Y}$.
\end{thm}

The proof is given in the Appendix.

\subsection{Experimental setup}

The MLC problem is decomposed into a series of multi-class classification problems, where each multi-class variable encodes one label powerset, with as many classes as the number of possible combinations of labels, or those present in the training data. At this point, it should be noted that the LP method is a special case of this framework since the whole label set is a particular label powerset (not necessarily minimal though). The above procedure can been summarized as follows:

\begin{enumerate}
  \item Learn the BN local graph $\cal{G}$ around the label set;
  \item Read off $\cal{G}$ the minimal label powersets and their respective Markov boundaries;
  \item Train an independent multi-class classifier on each minimal LP, with the input space restricted to its Markov boundary in $\cal{G}$;
  \item Aggregate the prediction of each classifier to output the most probable explanation, i.e. $\operatorname*{arg\,max}_\mathbf{y} p(\mathbf{y} \mid \mathbf{x})$.
\end{enumerate}

To assess separately the quality of the minimal label powerset decomposition and the feature subset selection with Markov boundaries, we investigate 4 scenarios:

\begin{itemize}
  \item BR without feature selection (denoted  \textbf{BR}): a classifier is trained on each single label with all features as input. This is the simplest approach as it does not exploit any label dependency. It serves as a baseline learner for comparison purposes.
  \item BR with feature selection (denoted \textbf{BR+MB}): a classifier is trained on each single label with the input space restricted to its Markov boundary in $\cal{G}$. Compared to the previous strategy, we evaluate here the effectiveness of the feature selection task.
  \item Minimum label powerset method without feature selection (denoted \textbf{MLP}): the minimal label powersets are obtained from the DAG. All features are used as inputs.
  \item Minimum label powerset method with feature selection (denoted \textbf{MLP+MB}): the minimal label powersets are obtained from the DAG. the input space is restricted to the Markov boundary of the labels in that powerset.
\end{itemize}

We use the same base learner in each meta-learning method: the well-known Random Forest classifier \citep{Breiman01}. RF achieves good performance in standard classification as well as in multi-label problems \citep{Kocev2007, Madjarov2012}, and is able to handle both continuous and discrete data easily, which is much appreciated. The standard RF implementation in \textit{R} \citep{randomForest}\footnote{\url{http://cran.r-project.org/web/packages/randomForest}} was used. For practical purposes, we restricted the forest size of RF to 100 trees, and left the other parameters to their default values.

\subsubsection{Data and performance indicators}

A total of 10 multi-label data sets are collected for experiments in this section, whose characteristics are summarized in Table~\ref{tab:datasets-mll}. These data sets come from different problem domains including text, biology, and music. They can be found on the Mulan\footnote{\url{http://mulan.sourceforge.net/datasets.html}} repository, except for \emph{image} which comes from Zhou\footnote{\url{http://lamda.nju.edu.cn/data_MIMLimage.ashx}} \citep{Maron1998}. Let ${\mathcal D}$ be the multi-label data set, we use $|{\cal D}|, dim({\cal D}), L({\cal D}), F({\cal D})$ to represent the number of examples, number of features, number of possible labels, and feature type respectively. $DL({\cal D}) = |\{Y | \exists x : (x,Y ) \in {\cal D}\}|$ counts the number of distinct label combinations appearing in the data set. Continues values are binarized during the BN structure learning phase.

\begin{table}[ht]
  \caption{Data sets characteristics}
  \label{tab:datasets-mll}
  \centering
  \vspace{\tabcapsep}
  \footnotesize
  \setlength{\tabcolsep}{3pt}
  \centering
  \begin{tabular}{llrrlrr}
    \toprule
    data set & domain & $|{\cal D}|$ & $dim({\cal D})$ & $ F({\cal D})$ &   $L({\cal D})$ & $DL({\cal D})$ \\
    \midrule
    emotions & music & 593 & 72 & cont. & 6 & 27 \\
    yeast & biology & 2417 & 103 & cont. & 14 & 198 \\
    image & images & 2000 & 135 & cont. & 5 & 20 \\
    scene & images & 2407 & 294 & cont. & 6 & 15 \\
    slashdot & text & 3782 & 1079 & disc. & 22 & 156 \\
    genbase & biology & 662 & 1186 & disc. & 27 & 32 \\
    medical & text & 978 & 1449 & disc. & 45 & 94 \\
    enron & text & 1702 & 1001 & disc. & 53 & 753 \\
    bibtex & text & 7395 & 1836 & disc. & 159 & 2856 \\
    corel5k & images & 5000 & 499 & disc. & 374 & 3175 \\
    \bottomrule
  \end{tabular}
\end{table}

The performance of a multi-label classifier can be assessed by several evaluation measures \citep{Tsoumakas2010a}. We focus on maximizing a non-decomposable score function: the global accuracy (also termed subset accuracy, complement of the $0/1$-loss), which measures the correct classification rate of the whole label set (exact match of all labels required). Note that the global accuracy implicitly takes into account the label correlations. It is therefore a very strict evaluation measure as it requires an exact match of the predicted and the true set of labels. It was recently proved in \citep{Dembczynski2012} that BR is optimal for decomposable loss functions (e.g., the hamming loss), while non-decomposable loss functions (e.g. subset loss) inherently require the knowledge of the label conditional distribution. 10-fold cross-validation was performed for the evaluation of the MLC methods.

\subsubsection{Results}

\begin{figure*}
  \centering
  \subfloat[][Emotions]{
    \centering 
    \includegraphics[clip, trim = 0 0 0 0in, height=0.46\textwidth, width=0.46\textwidth]{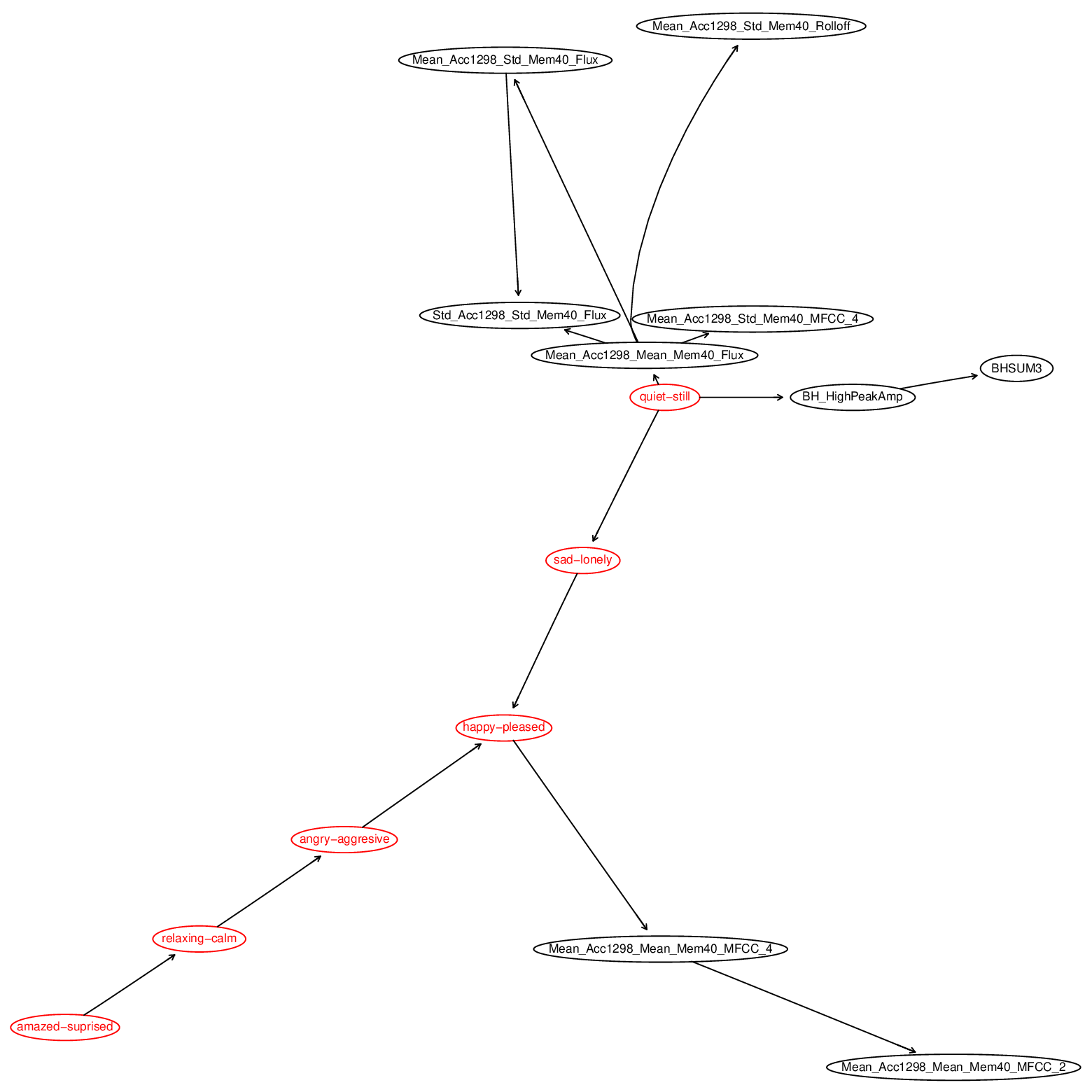}
    \includegraphics[clip, trim = 0 0 0 0in, height=0.46\textwidth, width=0.46\textwidth]{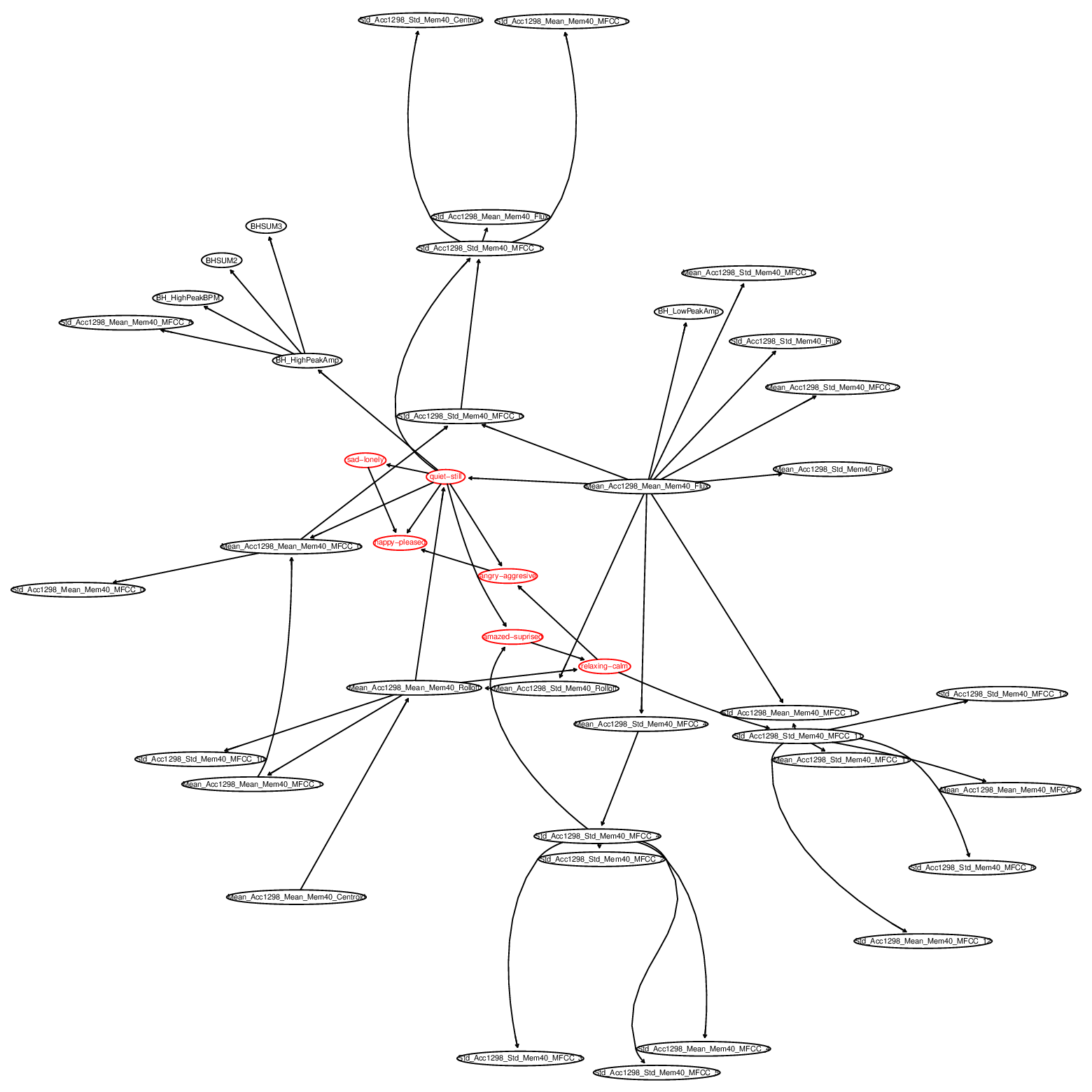}
  } \\
  \subfloat[][Yeast]{
    \centering 
    \includegraphics[clip, trim = 0 0 0 0in, height=0.46\textwidth, width=0.46\textwidth]{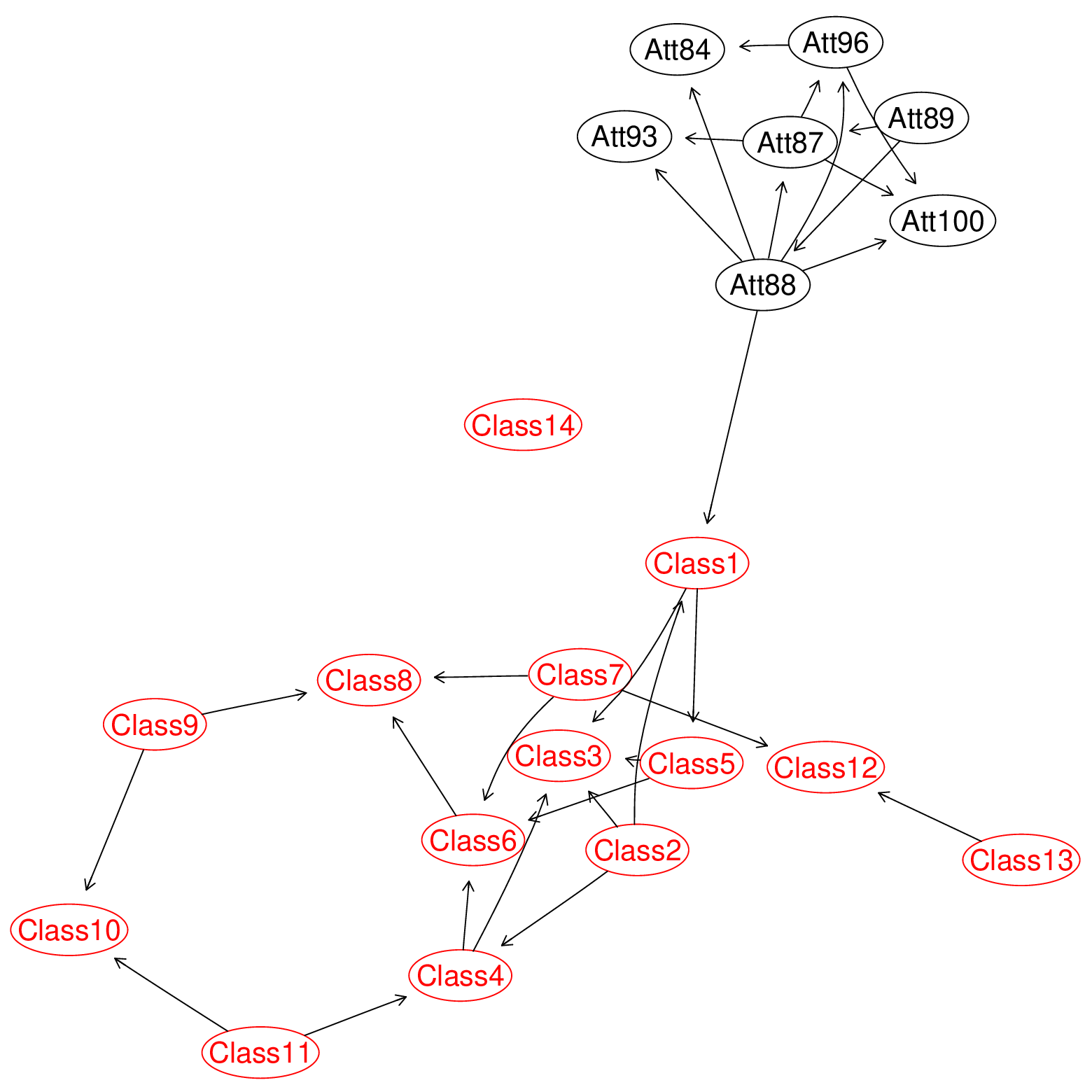}
    \includegraphics[clip, trim = 0 0 0 0in, height=0.46\textwidth, width=0.46\textwidth]{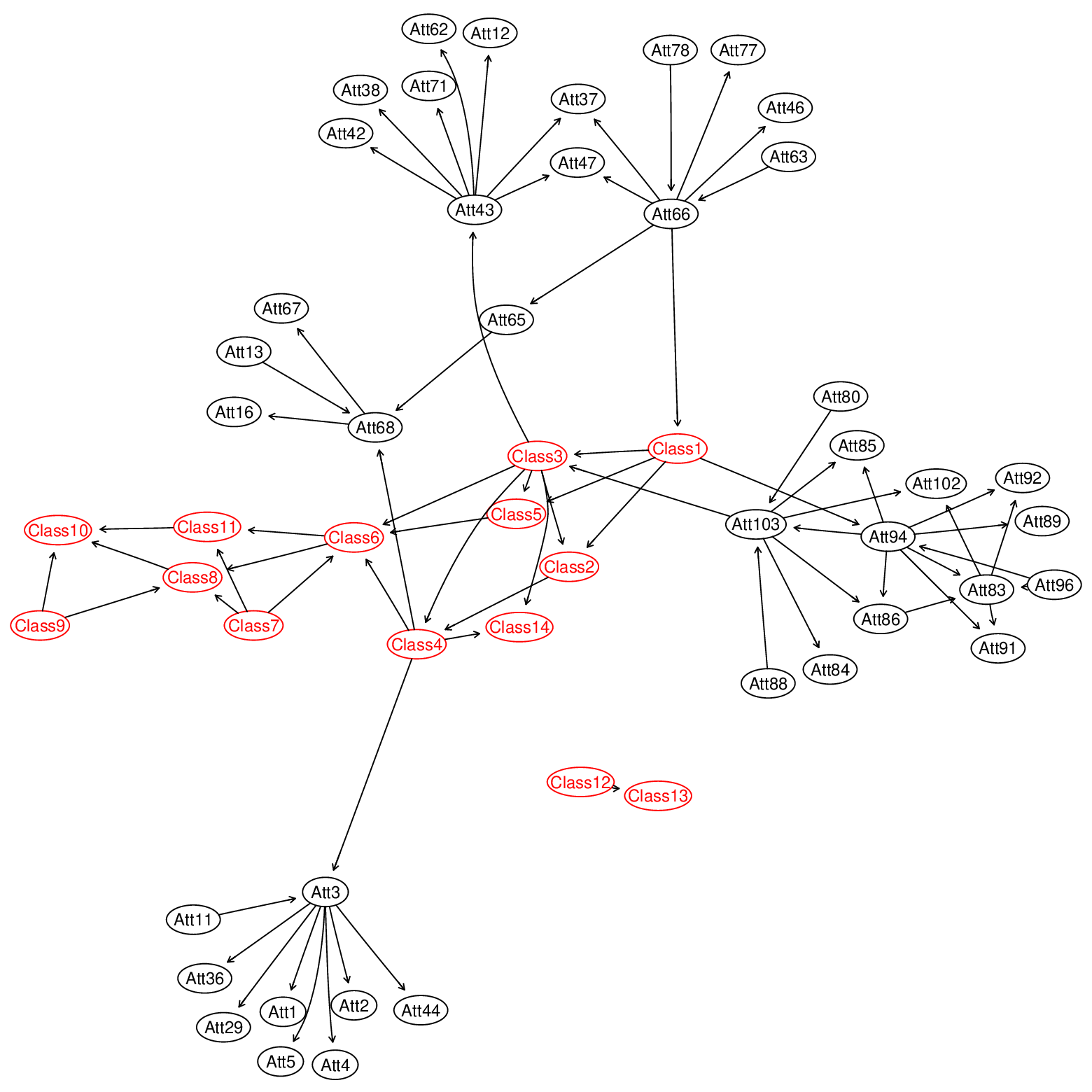}
  }
	  \caption{The local BN structures learned by MMHC (left plot) and H2PC (right plot) on a single cross-validation split, on Emotions and Yeast.}
  \label{fig:learned-dags-1}
\end{figure*}

\begin{figure*}
  \centering
  \subfloat[][Bibtex]{
    \centering 
    \includegraphics[clip, trim = 0 0 0 0in, height=0.46\textwidth, width=0.46\textwidth]{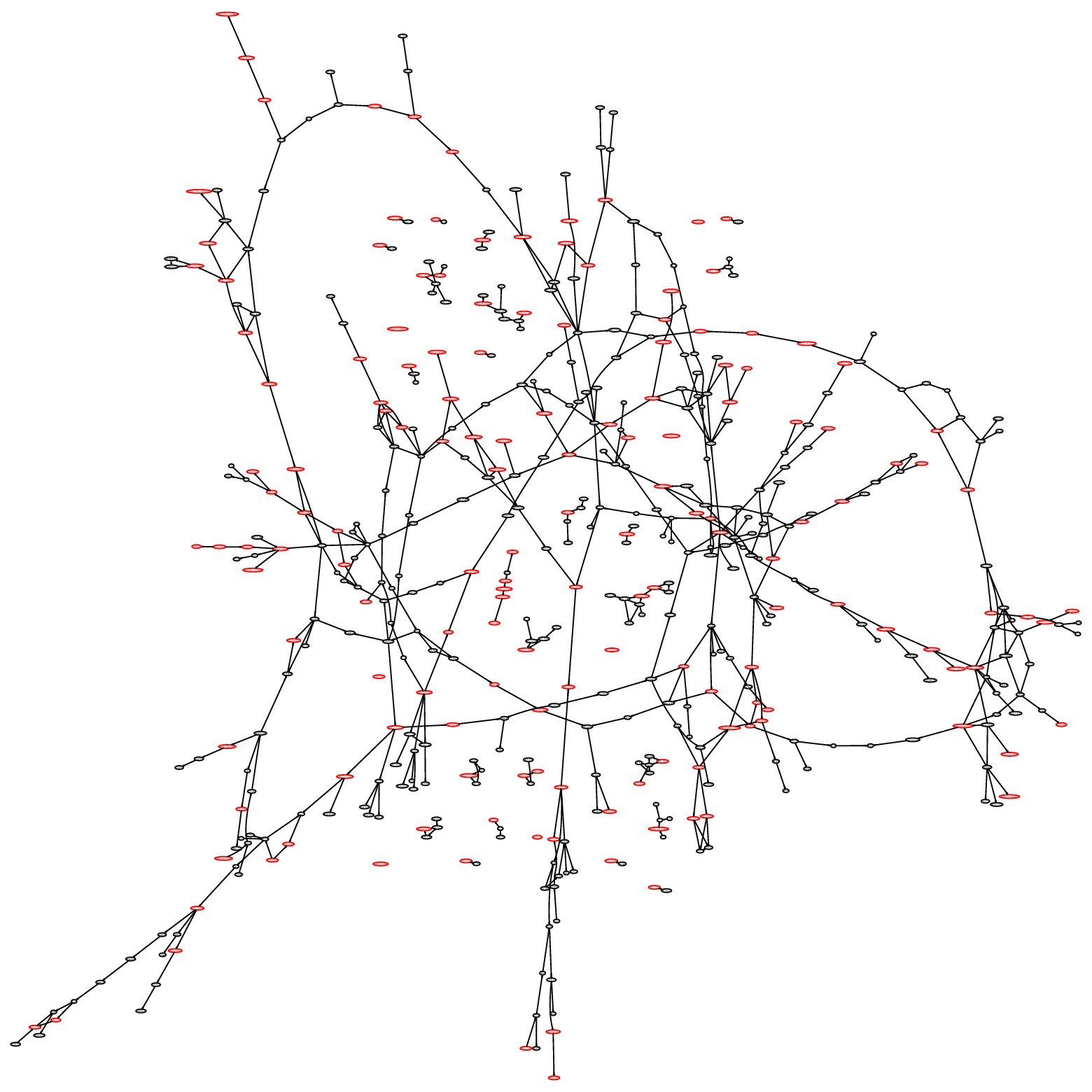}
    \includegraphics[clip, trim = 0 0 0 0in, height=0.46\textwidth, width=0.46\textwidth]{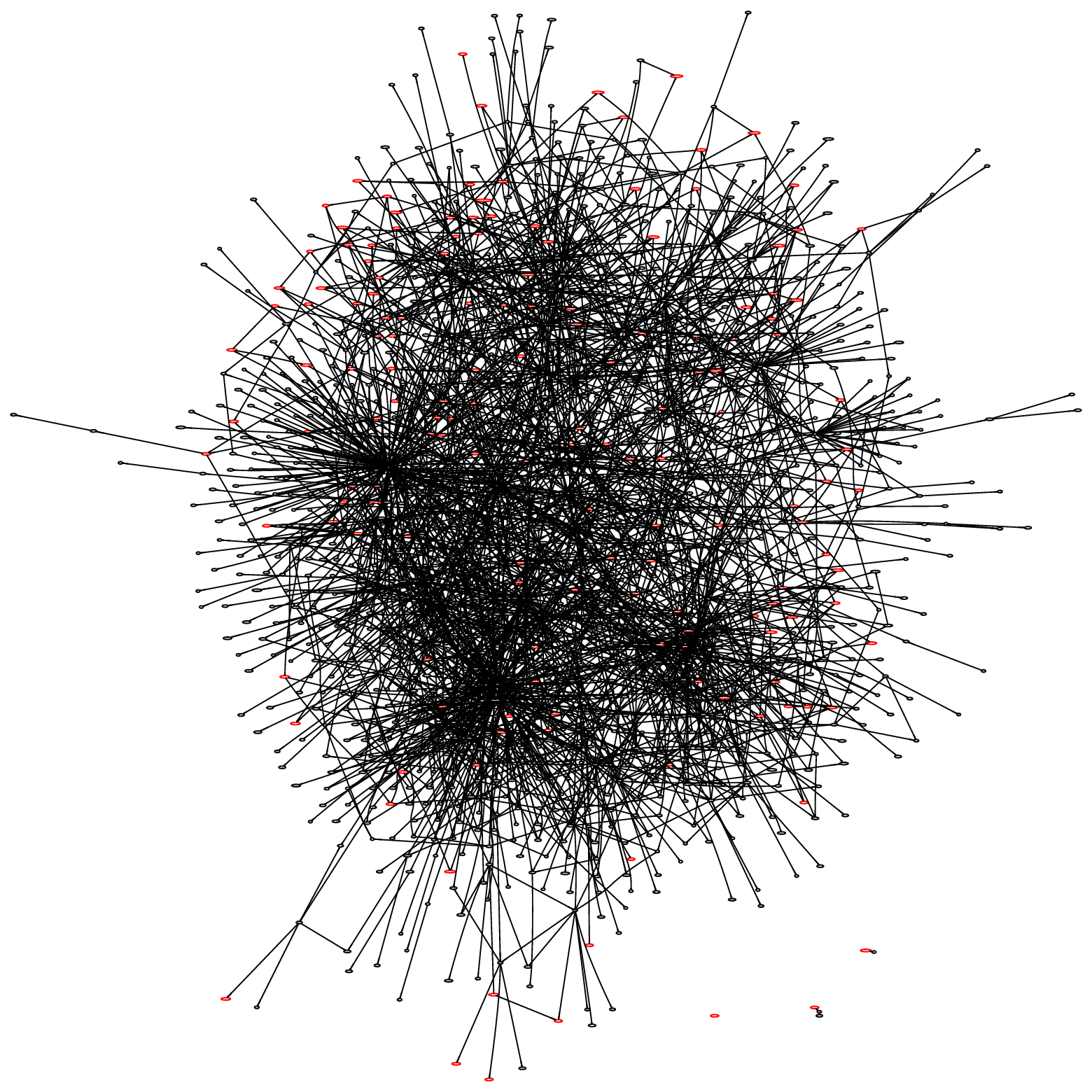}
  } \\
  \subfloat[][Corel5k]{
    \centering 
    \includegraphics[clip, trim = 0 0 0 0in, height=0.46\textwidth, width=0.46\textwidth]{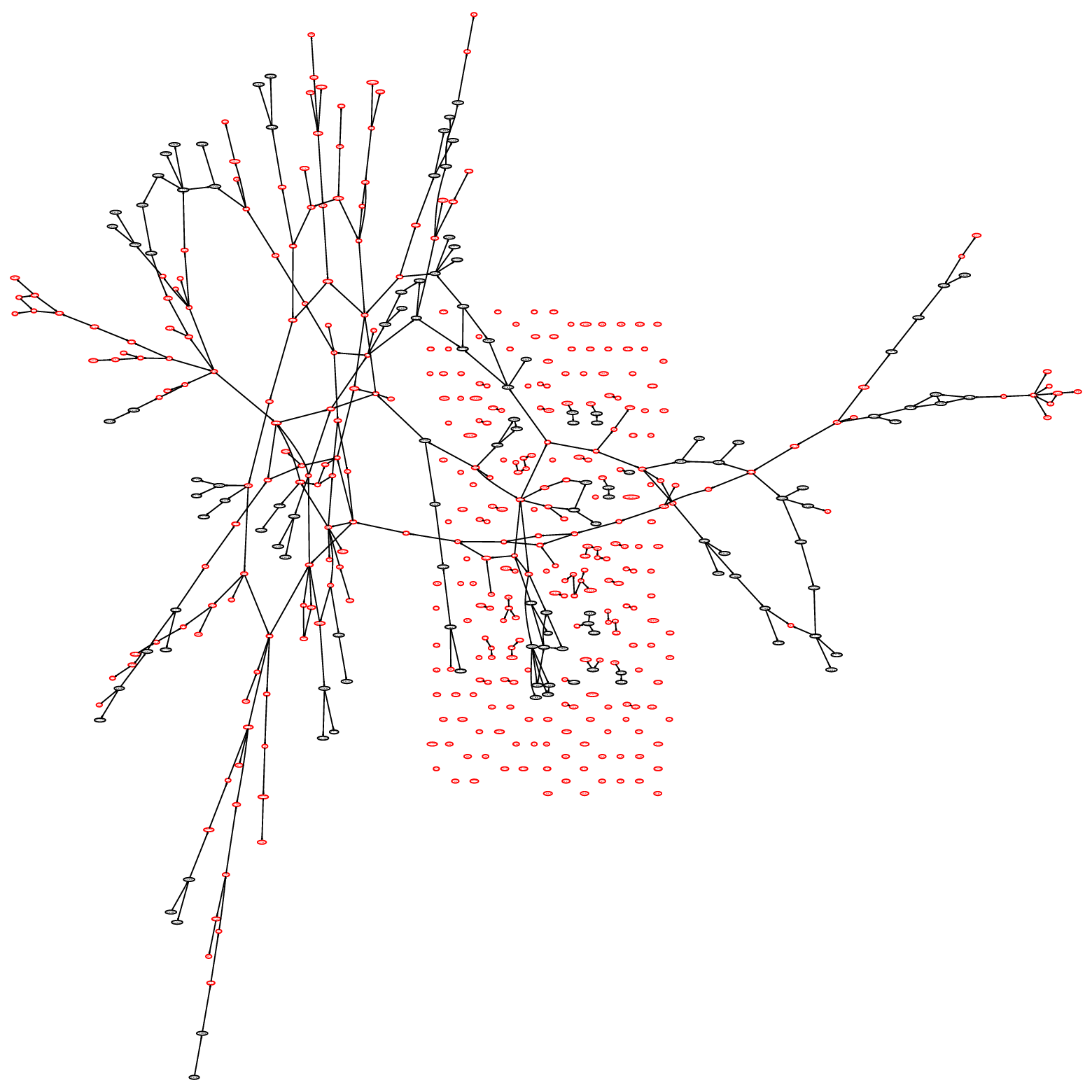}
    \includegraphics[clip, trim = 0 0 0 0in, height=0.46\textwidth, width=0.46\textwidth]{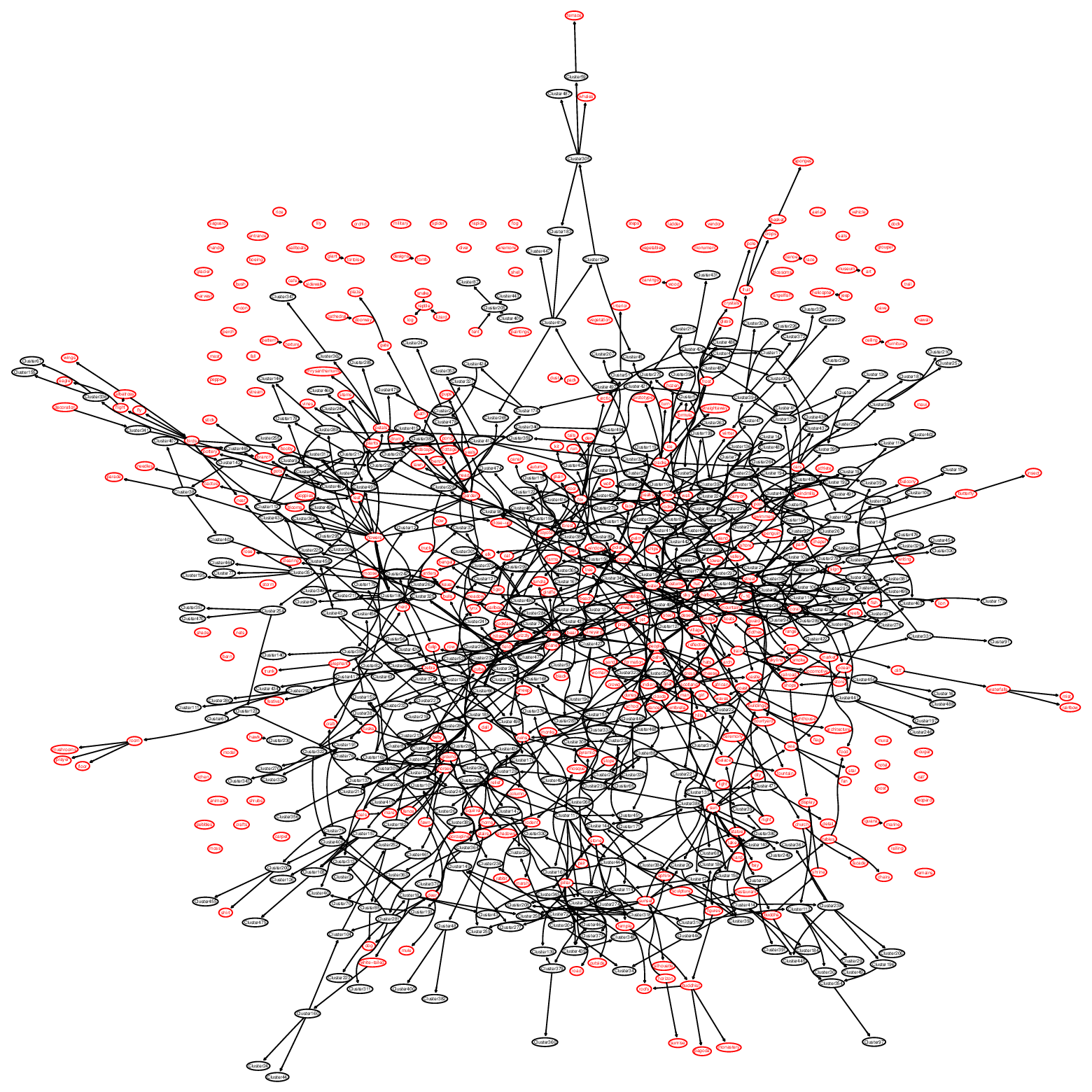}
  }
  \caption{The local BN structures learned by MMHC (left plot) and H2PC (right plot) on a single cross-validation split, on Bibtex and Corel5k.}
  \label{fig:learned-dags-5}
\end{figure*}

Table~\ref{tab:results-mlps} reports the outputs of H2PC and MMHC, Table~\ref{tab:results-gacc} shows the global accuracy of each method on the 10 data sets. Table~\ref{tab:mll-times} reports the running time and the average node degree of the labels in the DAGs obtained with both methods. Figures~\ref{fig:learned-dags-1} up to~\ref{fig:learned-dags-5} display graphically the local DAG structures around the labels, obtained with H2PC and MMHC, for illustration purposes.

Several conclusions may be drawn from these experiments. First, we may observe by inspection of the average degree of the label nodes in Table~\ref{tab:mll-times} that several DAGs are densely connected, like \textit{scene} or \textit{bibtex}, while others are rather sparse, like \textit{genbase, medical, corel5k}. The DAGs displayed in Figures~\ref{fig:learned-dags-1} up to~\ref{fig:learned-dags-5} lend themselves to interpretation. They can be used for encoding as well as portraying the conditional independencies, and the d-sep criterion can be used to read them off the graph. Many label powersets that are reported in Table~\ref{tab:results-mlps} can be identified by graphical inspection of the DAGs. For instance, the two label powersets in \textit{yeast} (Figure~\ref{fig:learned-dags-1}, bottom plot) are clearly noticeable in both DAGs.  Clearly, BNs have a number of advantages over alternative methods. They lay bare useful information about the label dependencies which is crucial if one is interested in gaining an understanding of the underlying domain. It is however well beyond the scope of this paper to delve deeper into the DAG interpretation. Overall, it appears that the structures recovered by H2PC are significantly denser, compared to MMHC. On \textit{bibtex} and \textit{enron}, the increase of the average label degree is the most spectacular. On \textit{bibtex} (resp. \textit{enron}), the average label degree has raised from 2.6 to 6.4 (resp. from 1.2 to 3.3). This result is in nice agreement with the experiments in the previous section, as H2PC was shown to consistently reduce the rate of false negative edges with respect to MMHC (at the cost of a slightly higher false discovery rate).

Second, Table~\ref{tab:results-mlps} is very instructive as it reveals that on \textit{emotions, image, and scene}, the MLP approach boils down to the LP scheme. This is easily seen as there is only one minimal label powerset extracted on average. In contrast, on \textit{genbase} the MLP approach boils down to the simple BR scheme.  This is confirmed by inspecting Table~\ref{tab:results-gacc}: the performances of MMHC and H2PC (without feature selection) are equal. An interesting observation upon looking at the distribution of the label powerset size shows that for the remaining data sets, the MLP mostly decomposes the label set in two parts: one on which it performs BR and the other one on which it performs LP.  Take for instance \textit{enron}, it can be seen from Table~\ref{tab:results-mlps} that there are approximately 23 label singletons and a powerset of 30 labels with H2PC for a total of 53 labels.  The gain in performance with MLP over BR, our baseline learner, can be ascribed to the quality of the label powerset decomposition as BR is ignoring label dependencies. As expected, the results using MLP clearly dominate those obtained using BR on all data sets except \textit{genbase}.  The DAGs display the minimum label powersets and their relevant features.

Third, H2PC compares favorably to MMHC. On \textit{scene} for instance, the accuracy of MLP+MB has raised from 20\% (with MMHC) to 56\% (with H2PC). On \textit{yeast}, it has raised from 7\% to 23\%. Without feature selection, the difference in global accuracy is less pronounced but still in favor of H2PC. A Wilcoxon signed rank paired test reveals statistically significant improvements of H2PC over MMHC in the MLP approach without feature selection ($p<0.02$). This trend is more pronounced when the feature selection is used ($p<0.001$), using MLP-MB. Note however that the LP decomposition for H2PC and MMHC on \textit{yeast} and \textit{image} are strictly identical, hence the same accuracy values in Table~\ref{tab:results-gacc} in the MLP column.

Fourth, regarding the utility of the feature selection, it is difficult to reach any conclusion. Whether BR or MLP is used, the use of the selected features as inputs to the classification model is not shown greatly beneficial in terms of global accuracy on average. The performance of BR and our MLP  with all the features outperforms that with the selected features in 6 data sets but the feature selection leads to actual improvements in 3 data sets. The difference in accuracy with and without the feature selection was not shown to be statistically significant ($p>0.20$ with a Wilcoxon signed rank paired test). Surprisingly, the feature selection did exceptionally well on \textit{genbase}. On this data set, the increase in accuracy is the most impressive: it raised from 7\% to 98\% which is atypical. The dramatic increase in accuracy on \textit{genbase} is due solely to the restricted feature set as input to the classification model. This is also observed on \textit{medical}, to a lesser extent though. Interestingly, on large and densely connected networks (e.g. \textit{bibtex}, \textit{slashdot} and \textit{corel5K}), the feature selection performed very well in terms of global accuracy and significantly reduced the input space which is noticeable. On \emph{emotions}, \emph{yeast}, \emph{image}, \emph{scene} and \emph{genbase}, the method reduced the feature set down to nearly 1/100 its original size. The feature selection should be evaluated in view of its effectiveness at balancing the increasing error and the decreasing computational burden by drastically reducing the feature space.  We should also keep in mind that the feature relevance cannot be defined independently of the learner and the model-performance metric (e.g., the loss function used). Admittedly, our feature selection based on the Markov boundaries is not necessarily optimal for the base MLC learner used here, namely the Random Forest model.

As far as the overall running time performance is concerned, we see from Table \ref{tab:mll-times} that for both methods, the running time grows somewhat exponentially with the size of the Markov boundary and the number of features, hence the considerable rise in total running time on \textit{bibtex}. H2PC takes almost 200 times longer on \textit{bibtex} (1826 variables) and \textit{enron} (1001 variables) which is quite considerable but still affordable (44 hours of single-CPU time on \textit{bibtex} and 13 hours on \textit{enron}). We also observe that the size of the parent set with H2PC is 2.5 (on \textit{bibtex}) and 3.6 (on \textit{enron}) times larger than that of MMHC (which may hurt interpretability). In fact, the running time is known to increase exponentially with the parent set size of the true underlying network. This is mainly due the computational overhead of greedy search-and-score procedure with larger parent sets which is the most promising part to optimize in terms of computational gains as we discuss in the Conclusion.

\begin{table}
  \caption{Distribution of the number and the size of the minimal label powersets output by H2PC (top) and MMHC (bottom). On each data set: mean number of powersets, minimum/median/maximum number of labels per powerset, and minimum/median/maximum number of distinct classes per powerset. The total number of labels and distinct labels combinations is recalled for convenience.}
  \label{tab:results-mlps}
  \vspace{\tabcapsep}
  \footnotesize
  \setlength{\tabcolsep}{3pt}
  \centering
  \centering
  \begin{tabular}{lrcccc}

	\toprule
    \multicolumn{6}{c}{H2PC} \\
    \midrule

    \multicolumn{1}{c}{\multirow{2}{*}{dataset}} &
    \multicolumn{1}{c}{\multirow{2}{*}{\# LPs}} &
    \multicolumn{1}{c}{\# labels / LP} &
    \multicolumn{1}{c}{\multirow{2}{*}{$L({\cal D})$}} &
    \multicolumn{1}{c}{\# classes / LP} &
    \multicolumn{1}{c}{\multirow{2}{*}{$DL({\cal D})$}} \\
    & & min/med/max & & min/med/max & \\

    \midrule
    emotions  & $1.0 \pm 0.0$    & 6 / 6 / 6   & (6)   & 26 / 27 / 27  & (27) \\
    yeast     & $2.0 \pm 0.0$    & 2 / 7 / 12  & (14)  & 3 / 64 / 135  & (198) \\
    image     & $1.0 \pm 0.0$    & 5 / 5 / 5   & (5)   & 19 / 20 / 20  & (20) \\
    scene     & $1.0 \pm 0.0$    & 6 / 6 / 6   & (6)   & 14 / 15 / 15  & (15) \\
    slashdot  & $9.5 \pm 0.8$    & 1 / 1 / 14  & (22)  & 1 / 2 / 111   & (156) \\
    genbase   & $26.9 \pm 0.3$   & 1 / 1 / 2   & (27)  & 1 / 2 / 2     & (32) \\
    medical   & $38.6 \pm 1.8$   & 1 / 1 / 6   & (45)  & 1 / 2 / 10    & (94) \\
    enron     & $24.5 \pm 1.9$   & 1 / 1 / 30  & (53)  & 1 / 2 / 334   & (753) \\
    bibtex    & $39.6 \pm 4.3$   & 1 / 1 / 112 & (159) & 2 / 2 / 1588  & (2856) \\
    corel5k   & $97.7 \pm 3.9$   & 1 / 1 / 263 & (374) & 1 / 2 / 2749  & (3175) \\
    \midrule

    \midrule
	\multicolumn{6}{c}{MMHC} \\
    \midrule

    \multicolumn{1}{c}{\multirow{2}{*}{dataset}} &
    \multicolumn{1}{c}{\multirow{2}{*}{\# LPs}} &
    \multicolumn{1}{c}{\# labels / LP} &
    \multicolumn{1}{c}{\multirow{2}{*}{$L({\cal D})$}} &
    \multicolumn{1}{c}{\# classes / LP} &
    \multicolumn{1}{c}{\multirow{2}{*}{$DL({\cal D})$}} \\
    & & min/med/max & & min/med/max & \\

    \midrule
    emotions  & $1.2 \pm 0.4$    & 1 / 6 / 6   & (6)   & 2 / 26 / 27   & (27) \\
    yeast     & $2.0 \pm 0.0$    & 1 / 7 / 13  & (14)  & 2 / 89 / 186  & (198) \\
    image     & $1.0 \pm 0.0$    & 5 / 5 / 5   & (5)   & 19 / 20 / 20  & (20) \\
    scene     & $1.0 \pm 0.0$    & 6 / 6 / 6   & (6)   & 14 / 15 / 15  & (15) \\
    slashdot  & $15.1 \pm 0.6$   & 1 / 1 / 9   & (22)  & 1 / 2 / 46    & (156) \\
    genbase   & $27.0 \pm 0.0$   & 1 / 1 / 1   & (27)  & 1 / 2 / 2     & (32) \\
    medical   & $41.7 \pm 1.4$   & 1 / 1 / 4   & (45)  & 1 / 2 / 7     & (94) \\
    enron     & $36.8 \pm 2.7$   & 1 / 1 / 12  & (53)  & 1 / 2 / 119   & (753) \\
    bibtex    & $77.2 \pm 3.1$   & 1 / 1 / 28  & (159) & 2 / 2 / 233   & (2856) \\
    corel5k   & $165.3 \pm 5.5$  & 1 / 1 / 177 & (374) & 1 / 2 / 2294  & (3175) \\
    \bottomrule

  \end{tabular}
\end{table}

\begin{table}
  \caption{Global classification accuracies using 4 learning methods (BR, MLP, BR+MB, MLP+MB) based on the DAG obtained with H2PC and MMHC. Best values between H2PC and MMHC are boldfaced.}
  \label{tab:results-gacc}
  \vspace{\tabcapsep}
  \footnotesize
  \setlength{\tabcolsep}{3pt}
  \centering
  \begin{tabular}{lccccccc}
    \toprule
    \multicolumn{1}{c}{\multirow{2}{*}{dataset}} &
    \multicolumn{1}{c}{\multirow{2}{*}{BR}} &
    \multicolumn{2}{c}{MLP} &
    \multicolumn{2}{c}{BR+MB} &
    \multicolumn{2}{c}{MLP+MB} \\
    & & MMHC & H2PC & MMHC & H2PC & MMHC & H2PC \\
    \midrule
    emotions & 0.327 & 0.371          & \textbf{0.391} & 0.147 & \textbf{0.266} & 0.189 & \textbf{0.285} \\
    yeast    & 0.169 & \textbf{0.273} & 0.271          & 0.062 & \textbf{0.155} & 0.073 & \textbf{0.233} \\
    image    & 0.342 & 0.504          & 0.504          & 0.227 & \textbf{0.252} & 0.308 & \textbf{0.360} \\
    scene    & 0.580 & 0.733          & 0.733          & 0.123 & \textbf{0.361} & 0.199 & \textbf{0.565} \\
    slashdot & 0.355 & 0.419          & \textbf{0.474} & 0.274 & \textbf{0.373} & 0.278 & \textbf{0.439} \\
    genbase  & 0.069 & 0.069          & 0.069          & 0.974 & \textbf{0.976} & 0.974 & \textbf{0.976} \\
    medical  & 0.454 & 0.499          & \textbf{0.502} & 0.630 & \textbf{0.651} & 0.636 & \textbf{0.669} \\
    enron    & 0.134 & 0.165          & \textbf{0.180} & 0.024 & \textbf{0.079} & 0.079 & \textbf{0.157} \\
    bibtex   & 0.108 & 0.115          & \textbf{0.167} & 0.120 & \textbf{0.133} & 0.121 & \textbf{0.177} \\
    corel5k  & 0.002 & 0.030          & \textbf{0.043} & 0.000 & \textbf{0.002} & 0.010 & \textbf{0.034} \\
    \bottomrule
  \end{tabular}
\end{table}

\begin{table}
  \caption{DAG learning time (in seconds) and average degree of the label nodes.}
  \label{tab:mll-times}
  \vspace{\tabcapsep}
  \footnotesize
  \centering
  \begin{tabular}{lrrrr}
    \toprule
    \multicolumn{1}{c}{\multirow{2}{*}{dataset}} & \multicolumn{2}{c}{\multirow{1}{*}{MMHC}} & \multicolumn{2}{c}{\multirow{1}{*}{H2PC}} \\
    \multicolumn{1}{c}{\multirow{2}{*}{}} & \multicolumn{1}{c}{\multirow{1}{*}{time}} & \multicolumn{1}{c}{\multirow{1}{*}{label degree}} & \multicolumn{1}{c}{\multirow{1}{*}{time}} & \multicolumn{1}{c}{\multirow{1}{*}{label degree}} \\
    \midrule
    emotions  & 1.5   & $2.6 \pm 1.1$ & 16.2     & $4.7 \pm  1.8$ \\
    yeast     & 9.0   & $3.0 \pm 1.6$ & 90.9     & $5.0 \pm  2.9$ \\
    image     & 3.1   & $4.0 \pm 0.8$ & 28.4     & $4.6 \pm  1.0$ \\
    scene     & 9.9   & $3.1 \pm 1.0$ & 662.9    & $6.6 \pm  1.8$ \\
    slashdot  & 44.8  & $2.7 \pm 1.4$ & 822.3    & $4.0 \pm  5.2$ \\
    genbase   & 12.5  & $1.0 \pm 0.3$ & 12.4     & $1.1 \pm  0.3$ \\
    medical   & 43.8  & $1.7 \pm 1.1$ & 334.8    & $2.9 \pm  2.4$ \\
    enron     & 199.8 & $1.2 \pm 1.5$ & 47863.0  & $4.3 \pm  5.2$ \\ 
    bibtex    & 960.8 & $2.6 \pm 1.3$ & 159469.6 & $6.4 \pm 10.3$ \\ 
    corel5k   & 169.6 & $1.5 \pm 1.4$ & 2513.0   & $2.9 \pm  2.8$ \\ 
    \bottomrule
  \end{tabular}
\end{table}

\section{Discussion \& practical applications}

Our prime conclusion is that H2PC is a promising approach to constructing BN global or local structures around specific nodes of interest, with potentially thousands of variables. Concentrating on higher recall values while keeping the false positive rate as low as possible pays off in terms of goodness of fit and structure accuracy. Historically, the main practical difficulty in the application of BN structure discovery approaches has been a lack of suitable computing resources and relevant accessible software. The number of variables which can be included in exact BN analysis is still limited. As a guide, this might be less than about 40 variables for exact structural search techniques \citep{Perrier08,Kojima10}.  In contrast, constraint-based heuristics like the one presented in this study is capable of processing many thousands of features within hours on a personal computer, while maintaining a very high structural accuracy. H2PC and MMHC could potentially handle up to 10,000 labels in a few days of single-CPU time and far less by parallelizing the skeleton identification algorithm as discussed in \citep{Tsamardinos06,VillanuevaM14}.

The advantages in terms of structure accuracy and its ability to scale to thousands of variables opens up many avenues of future possible applications of H2PC in various domains as we shall discuss next. For example, BNs have especially proven to be useful abstractions in computational biology \citep{Nagarajan13,Scutari13,Prestat13,Aussem12a,Aussem10c,Pena08,Pen05}. Identifying the gene network is crucial for understanding the behavior of the cell which, in turn, can lead to better diagnosis and treatment of diseases. This is also of great importance for characterizing the function of genes and the proteins they encode in determining traits, psychology, or development of an organism.  Genome sequencing uses high-throughput techniques like DNA microarrays, proteomics, metabolomics and mutation analysis to describe the function and interactions of thousands of genes \citep{Zhang2006}. Learning BN models of gene networks from these huge data is still a difficult task \citep{Badea2004,Bernard2005,Friedman1999,Ott2004,Peer2001}. In these studies, the authors had decide in advance which genes were included in the learning process, in all the cases less than 1000, and which genes were excluded from it \citep{Pena08}. H2PC overcome the problem by focusing the search around a targeted gene: the key step is the identification of the vicinity of a node $X$ \citep{Pen05}.

Our second objective in this study was to demonstrate the potential utility of hybrid BN structure discovery to multi-label learning. In multi-label data where many inter-dependencies between the labels may be present, explicitly modeling all relationships between the labels is intuitively far more reasonable (as demonstrated in our experiments). BNs explicitly account for such inter-dependencies and the DAG allows us to identify an optimal set of predictors for every label powerset. The experiments presented here support the conclusion that local structural learning in the form of local neighborhood induction and Markov blanket is a theoretically well-motivated approach that can serve as a powerful learning framework for label dependency analysis geared toward multi-label learning. Multi-label scenarios are found in many application domains, such as multimedia annotation \citep{Snoek06,Trohidis08}, tag recommendation, text categorization \citep{McCallum99,Zhang2006}, protein function classification \citep{Roth07}, and antiretroviral drug categorization \citep{Borchani13}.

\section{Conclusion \& avenues for future research}

We first discussed a hybrid algorithm for global or local (around target nodes) BN structure learning called Hybrid HPC (H2PC).  Our extensive experiments showed that H2PC outperforms the state-of-the-art MMHC by a significant margin in terms of edge recall without sacrificing the number of extra edges, which is crucial for the soundness of the super-structure used during the second stage of hybrid methods, like the ones proposed in \citep{Perrier08,Kojima10}. The code of H2PC is open-source and publicly available online at \url{https://github.com/madbix/bnlearn-clone-3.4}.  Second, we discussed an application of H2PC to the multi-label learning problem which is a challenging problem in many real-world application domains. We established theoretical results, under the faithfulness condition, in order to characterize graphically the so-called minimal label powersets that appear as irreducible factors in the joint distribution and their respective Markov boundaries. As far as we know, this is the first investigation of Markov boundary principles  to the optimal variable/feature selection problem in multi-label learning. These formal results offer a simple guideline to characterize graphically : i) the minimal label powerset decomposition, (i.e. into minimal subsets $\mathbf{Y}_{LP} \subseteq \mathbf{Y}$ such that  $\mathbf{Y}_{LP} \indep \mathbf{Y} \setminus \mathbf{Y}_{LP} \mid \mathbf{X}$), and ii) the minimal subset of features, w.r.t an Information Theory criterion, needed to predict each label powerset, thereby reducing the input space and the computational burden of the multi-label classification. The theoretical analysis laid the foundation for a practical multi-label classification procedure. Another set of experiments were carried out on a broad range of multi-label data sets from different problem domains to demonstrate its effectiveness. H2PC was shown to outperform MMHC by a significant margin in terms of global accuracy.

We suggest several avenues for future research. As far as BN structure learning is concerned, future work will aim at: 1) ascertaining which independence test (e.g. tests targeting specific distributions, employing parametric assumptions etc.) is most suited to the data at hand \citep{Tsamardinos10,Scutari11}; 2) controlling the false discovery rate of the edges in the graph output by H2PC \citep{Pena08} especially when dealing with more nodes than samples, e.g. learning gene networks from gene expression data.  In this study, H2PC was run independently on each node without keeping track of the dependencies found previously. This lead to some loss of efficiency due to redundant calculations. The optimization of the H2PC code is currently being undertaken to lower the computational cost while maintaining its performance. These optimizations will include the use of a cache to store the (in)dependencies and the use of a global structure. Other interesting research avenues to explore are extensions and modifications to the greedy search-and-score procedure which is the most promising part to optimize in terms of computational gains. Regarding the  multi-label learning problem, experiments with several thousands of labels are currently been conducted and will be reported in due course.  We also intend to work on relaxing the faithfulness assumption and derive a practical multi-label classification algorithm based on H2PC that is correct under milder assumptions underlying the joint distribution (e.g. Composition, Intersection).  This needs further substantiation through more analysis.

\section*{Acknowledgments}

The authors thank Marco Scutari for sharing his \textit{bnlearn} package in \textit{R}. The research was also supported by grants from the European ENIAC Joint Undertaking (INTEGRATE project) and from the French Rh{\^o}ne-Alpes Complex Systems Institute (IXXI).

\section*{Appendix}

\begin{lem}
  \label{lem:mlp-uniqueness}
  $\forall \mathbf{Y}_{LP_i}, \mathbf{Y}_{LP_j} \in \mathbf{LP}$,  then $\mathbf{Y}_{LP_i} \cap \mathbf{Y}_{LP_j} \in \mathbf{LP}$.
\end{lem}

\begin{pf}
  To keep the notation uncluttered and for the sake of simplicity, consider a partition $\{\mathbf{Y}_1, \mathbf{Y}_2,\mathbf{Y}_3, \mathbf{Y}_4\}$ of $\mathbf{Y}$ such that
  $\mathbf{Y}_{LP_i} = \mathbf{Y}_1 \cup \mathbf{Y}_2$
  ,
  $\mathbf{Y}_{LP_j} = \mathbf{Y}_2 \cup \mathbf{Y}_3$
  . From the label powerset assumption for $\mathbf{Y}_{LP_i}$ and $\mathbf{Y}_{LP_j}$, we have
  $(\mathbf{Y}_1 \cup \mathbf{Y}_2) \indep (\mathbf{Y}_3 \cup \mathbf{Y}_4) \mid \mathbf{X}$
  and
  $(\mathbf{Y}_2 \cup \mathbf{Y}_3) \indep (\mathbf{Y}_1 \cup \mathbf{Y}_4) \mid \mathbf{X}$
  . From the Decomposition, we have that
  $\mathbf{Y}_2 \indep (\mathbf{Y}_3 \cup \mathbf{Y}_4) \mid \mathbf{X}$
  . Using the Weak union, we obtain that
  $\mathbf{Y}_2 \indep \mathbf{Y}_1 \mid (\mathbf{Y}_3 \cup \mathbf{Y}_4 \cup \mathbf{X})$
  . From these two facts, we can use the Contraction property to show that
  $\mathbf{Y}_2 \indep (\mathbf{Y}_1 \cup \mathbf{Y}_3 \cup \mathbf{Y}_4) \mid \mathbf{X}$
  . Therefore,
  $\mathbf{Y}_2 = \mathbf{Y}_{LP_i} \cap \mathbf{Y}_{LP_j}$
  is a label powerset by definition.
\end{pf}

\begin{lem}
  \label{th:yiyj-mlp}
  Let  $Y_i$ and $Y_j$ denote two distinct labels in $\mathbf{Y}$ and define by $\mathbf{Y}_{LP_i}$ and $\mathbf{Y}_{LP_j}$ their respective minimal label powerset. Then we have,
$$
\exists \mathbf{Z} \subseteq \mathbf{Y} \setminus \{Y_i,Y_j\}, \{Y_i\} \not\indep \{Y_j\} \mid (\mathbf{X} \cup \mathbf{Z}) \Rightarrow \mathbf{Y}_{LP_i} = \mathbf{Y}_{LP_j}
$$
\end{lem}

\begin{pf}
  Let us suppose $\mathbf{Y}_{LP_i} \neq \mathbf{Y}_{LP_j}$. By the label powerset definition for $\mathbf{Y}_{LP_i}$, we have $\mathbf{Y}_{LP_i} \indep \mathbf{Y} \setminus \mathbf{Y}_{LP_i} \mid \mathbf{X}$. As $\mathbf{Y}_{LP_i} \cap \mathbf{Y}_{LP_j} = \emptyset$ owing to Lemma~\ref{lem:mlp-uniqueness}, we have that $Y_j \not\in \mathbf{Y}_{LP_i}$.
  $\forall \mathbf{Z} \subseteq \mathbf{Y} \setminus \{Y_i,Y_j\}$, $\mathbf{Z}$ can be decomposed as $\mathbf{Z}_i \cup \mathbf{Z}_j$ such that
  $\mathbf{Z}_i = \mathbf{Z} \cap (\mathbf{Y}_{LP_i} \setminus \{Y_i\})$
  and
  $\mathbf{Z}_j = \mathbf{Z} \setminus \mathbf{Z}_i$
  . Using the Decomposition property, we have that
  $(\{Y_i\} \cup \mathbf{Z}_i)  \indep (\{Y_j\} \cup \mathbf{Z}_j) \mid \mathbf{X}$
  . Using the Weak union property, we have that
  $\{Y_i\} \indep \{Y_j\} \mid (\mathbf{X} \cup \mathbf{Z})$
  . As this is true $\forall \mathbf{Z} \subseteq \mathbf{Y} \setminus \{Y_i,Y_j\}$, then $\nexists \mathbf{Z} \subseteq \mathbf{Y} \setminus \{Y_i,Y_j\}, \{Y_i\} \not\indep \{Y_j\} \mid (\mathbf{X} \cup \mathbf{Z})$
  , which completes the proof.
\end{pf}

\bigskip

\begin{lem}
  \label{lem:mlp-split}
  Consider $\mathbf{Y}_{LP}$ a minimal label powerset. Then, if $p$ satisfies the Composition property, $\mathbf{Z} \not\indep \mathbf{Y}_{LP} \setminus \mathbf{Z} \mid \mathbf{X}$.
\end{lem}

\begin{pf}
  By contradiction, suppose a nonempty $\mathbf{Z}$ exists, such that
  $\mathbf{Z} \indep \mathbf{Y}_{LP} \setminus \mathbf{Z} \mid \mathbf{X}$
  . From the label powerset assumption of $\mathbf{Y}_{LP}$, we have that
  $\mathbf{Y}_{LP} \indep \mathbf{Y} \setminus \mathbf{Y}_{LP} \mid \mathbf{X}$
  . From these two facts, we can use the Composition property to show that
  $\mathbf{Z} \indep \mathbf{Y} \setminus \mathbf{Z} \mid \mathbf{X}$
  which contradicts the minimal label powerset assumption of $\mathbf{Y}_{LP}$. This concludes the proof.
\end{pf}

\bigskip

\setcounterref{thm}{th:dSep_mlp}

\begin{thm}
  Suppose $p$ is faithful to a DAG $\cal{G}$. Then, $Y_i$ and $Y_j$ belong to the same minimal label powerset if and only if there exists an undirected path in $\cal{G}$ between nodes $Y_i$ and $Y_j$ in $\mathbf{Y}$ such that all intermediate nodes $Z$ are either (i) $Z \in \mathbf{Y}$, or (ii) $Z \in \mathbf{X}$ and $Z$ has two parents in $\mathbf{Y}$ (i.e. a collider  of the form $Y_p\rightarrow X \leftarrow Y_q$).
\end{thm}

\setcounterref{thm}{lem:mlp-split}

\begin{pf}
  Suppose such a path exists. By conditioning on all the intermediate colliders $Y_k$ in $\mathbf{Y}$ of the form $Y_p \rightarrow Y_k \leftarrow Y_q$ along an undirected path in $\cal{G}$ between nodes $Y_i$ and $Y_j$, we ensure that  $\exists \mathbf{Z} \subseteq \mathbf{Y} \setminus \{Y_i,Y_j\},  \overline{\textbf{dSep}}(Y_i,Y_j \mid\mathbf{X}\cup\mathbf{Z})$. Due to the  faithfulness, this is equivalent to $\{Y_i\} \not\indep \{Y_j\} \mid (\mathbf{X} \cup \mathbf{Z})$. From Lemma~\ref{th:yiyj-mlp}, we conclude that $Y_i$ and $Y_j$ belong to the same minimal label powerset.
  To show the inverse, note that owing to Lemma~\ref{lem:mlp-split}, we know that there exists no partition $\{\mathbf{Z}_i,\mathbf{Z}_j\}$ of $\mathbf{Z}$ such that $(\{Y_i\}\cup\mathbf{Z}_i) \indep (\{Y_j\}\cup\mathbf{Z}_j) \mid \mathbf{X}$. Due to the faithfulness, there exists at least a link between $(\{Y_i\} \cup \mathbf{Z}_i)$ and $(\{Y_j\} \cup \mathbf{Z}_j)$ in the DAG, hence by recursion, there exists a path between $Y_i$ and $Y_j$ such that all intermediate nodes $Z$ are either (i) $Z \in \mathbf{Y}$, or (ii) $Z \in \mathbf{X}$ and $Z$ has two parents in $\mathbf{Y}$ (i.e. a collider  of the form $Y_p\rightarrow X \leftarrow Y_q$).
\end{pf}

\bigskip

\setcounterref{thm}{th:mb-lp-dsep}

\begin{thm}
  Suppose $p$ is faithful to a DAG $\cal{G}$. Let $\mathbf{Y}_{LP} = \{Y_1, Y_2, \dots, Y_n\}$ be a label powerset. Then, its Markov boundary $\mathbf{M}$ in $\mathbf{U}$ is also its Markov boundary in $\mathbf{X}$, and is given in $\cal{G}$ by $\mathbf{M}=\bigcup_{j = 1}^{n} \{\mathbf{PC}_{Y_j} \cup \mathbf{SP}_{Y_j}\} \setminus \mathbf{Y}$.
\end{thm}

\setcounterref{thm}{lem:mlp-split}

\begin{pf}
  First, we prove that $\mathbf{M}$ is a Markov boundary of $\mathbf{Y}_{LP}$ in $\mathbf{U}$. Define $\mathbf{M}_i$ the Markov boundary of $Y_i$ in $\mathbf{U}$, and $\mathbf{M'}_i = \mathbf{M}_i \setminus \mathbf{Y}$. From Theorem~\ref{th:faithfulness}, $\mathbf{M}_i$ is given in $\cal{G}$ by $\mathbf{M}_i = \mathbf{PC}_{Y_j} \cup \mathbf{SP}_{Y_j}$. We may now prove that $\mathbf{M'}_1 \cup \dots \cup \mathbf{M'}_n$ is a Markov boundary of $\{Y_1, Y_2, \dots, Y_n\}$ in $\mathbf{U}$. We show first that the statement holds for $n=2$ and then conclude that it holds for all $n$ by induction. Let $\mathbf{W}$ denote $\mathbf{U} \setminus (\mathbf{Y}_1 \cup \mathbf{Y}_2 \cup \mathbf{M'}_1 \cup \mathbf{M'}_2)$, and define $\mathbf{Y}_1 = \{Y_1\}$ and $\mathbf{Y}_2 = \{Y_2\}$. From the Markov blanket assumption for $\mathbf{Y}_1$ we have
  $\mathbf{Y}_1 \indep \mathbf{U} \setminus (\mathbf{Y}_1 \cup \mathbf{M}_1) \mid \mathbf{M}_1$
  . Using the Weak Union property, we obtain that
  $\mathbf{Y}_1 \indep \mathbf{W} \mid \mathbf{M'}_1 \cup \mathbf{M'}_2 \cup \mathbf{Y}_2$
  . Similarly we can derive
  $\mathbf{Y}_2 \indep \mathbf{W} \mid \mathbf{M'}_2 \cup \mathbf{M'}_1 \cup \mathbf{Y}_1$
  . Combining these two statements yields
  $\mathbf{Y}_1 \cup \mathbf{Y}_2 \indep \mathbf{W} \mid \mathbf{M'}_1 \cup \mathbf{M'}_2$
  due to the Intersection property. Let
  $\mathbf{M} = \mathbf{M'}_1 \cup \mathbf{M'}_2$
  the last expression can be formulated as
  $\mathbf{Y}_1 \cup \mathbf{Y}_2 \indep \mathbf{U} \setminus (\mathbf{Y}_1 \cup \mathbf{Y}_2 \cup \mathbf{M}) \mid \mathbf{M}$
  which is the definition of a Markov blanket of $\mathbf{Y}_1 \cup \mathbf{Y}_2$ in $\mathbf{U}$. We shall now prove that $\mathbf{M}$ is minimal. Let us suppose that it is not the case, i.e., $\exists \mathbf{Z} \subset \mathbf{M}$ such that
  $\mathbf{Y}_1 \cup \mathbf{Y}_2 \indep \mathbf{Z} \cup (\mathbf{U} \setminus (\mathbf{Y}_1 \cup \mathbf{Y}_2 \cup \mathbf{M})) \mid \mathbf{M} \setminus \mathbf{Z}$
  . Define $\mathbf{Z}_1 = \mathbf{Z} \cap \mathbf{M}_1$ and $\mathbf{Z}_2 = \mathbf{Z} \cap \mathbf{M}_2$, we may apply the Weak Union property to get
  $\mathbf{Y}_1 \indep \mathbf{Z}_1 \mid (\mathbf{M} \setminus \mathbf{Z}) \cup \mathbf{Y}_2 \cup \mathbf{Z}_2 \cup (\mathbf{U} \setminus (\mathbf{Y} \cup \mathbf{M}))$
  which can be rewritten more compactly as
  $\mathbf{Y}_1 \indep \mathbf{Z}_1 \mid (\mathbf{M}_1 \setminus \mathbf{Z}_1) \cup (\mathbf{U} \setminus (\mathbf{Y}_1 \cup \mathbf{M}_1))$
  . From the Markov blanket assumption, we have
  $\mathbf{Y}_1 \indep \mathbf{U} \setminus (\mathbf{Y}_1 \cup \mathbf{M}_1) \mid \mathbf{M}_1$
  . We may now apply the Intersection property on these two statements to obtain
  $\mathbf{Y}_1 \indep \mathbf{Z}_1 \cup (\mathbf{U} \setminus (\mathbf{Y}_1 \cup \mathbf{M}_1)) \mid \mathbf{M}_1 \setminus \mathbf{Z}_1$
  . Similarly, we can derive
  $\mathbf{Y}_2 \indep \mathbf{Z}_2 \cup (\mathbf{U} \setminus (\mathbf{Y}_2 \cup \mathbf{M}_2)) \mid \mathbf{M}_2 \setminus \mathbf{Z}_2$
  . From the Markov boundary assumption of $\mathbf{M}_1$ and $\mathbf{M}_2$, we have necessarily $\mathbf{Z}_1 = \emptyset$ and $\mathbf{Z}_2 = \emptyset$, which in turn yields $\mathbf{Z} = \emptyset$. To conclude for any $n>2$, it suffices to set $\mathbf{Y}_1 = \bigcup_{j = 1}^{n-1} \{Y_j\}$ and $\mathbf{Y}_2 = \{Y_n\}$ to conclude by induction.

  Second, we prove that $\mathbf{M}$ is a Markov blanket of $\mathbf{Y}_{LP}$ in $\mathbf{X}$. Define $\mathbf{Z} = \mathbf{M} \cap \mathbf{Y}$. From the label powerset definition, we have
  $\mathbf{Y}_{LP} \indep \mathbf{Y} \setminus \mathbf{Y}_{LP} \mid \mathbf{X}$
  . Using the Weak Union property, we obtain
  $\mathbf{Y}_{LP} \indep \mathbf{Z} \mid \mathbf{U} \setminus (\mathbf{Y}_{LP} \cup \mathbf{Z})$
  which can be reformulated as
  $\mathbf{Y}_{LP} \indep \mathbf{Z} \mid (\mathbf{U} \setminus (\mathbf{Y}_{LP} \cup \mathbf{M})) \cup (\mathbf{M} \setminus \mathbf{Z})$
  . Now, the Markov blanket assumption for $\mathbf{Y}_{LP}$ in $\mathbf{U}$ yields
  $\mathbf{Y}_{LP} \indep \mathbf{U} \setminus (\mathbf{Y}_{LP} \cup \mathbf{M}) \mid \mathbf{M}$
  which can be rewritten as
  $\mathbf{Y}_{LP} \indep \mathbf{U} \setminus (\mathbf{Y}_{LP} \cup \mathbf{M}) \mid (\mathbf{M} \setminus \mathbf{Z}) \cup \mathbf{Z}$
  . From the Intersection property, we get
  $\mathbf{Y}_{LP} \indep \mathbf{Z} \cup (\mathbf{U} \setminus (\mathbf{Y}_{LP} \cup \mathbf{M})) \mid \mathbf{M} \setminus \mathbf{Z}$
  . From the Markov boundary assumption of $\mathbf{M}$ in $\mathbf{U}$, we know that there exists no proper subset of $\mathbf{M}$ which satisfies this statement, and therefore $\mathbf{Z} = \mathbf{M} \cap \mathbf{Y} = \emptyset$.
  From the Markov blanket assumption of $\mathbf{M}$ in $\mathbf{U}$, we have
  $\mathbf{Y}_{LP} \indep \mathbf{U} \setminus (\mathbf{Y}_{LP} \cup \mathbf{M}) \mid \mathbf{M}$
  . Using the Decomposition property, we obtain
  $\mathbf{Y}_{LP} \indep \mathbf{X} \setminus \mathbf{M} \mid \mathbf{M}$
  which, together with the assumption $\mathbf{M} \cap \mathbf{Y} = \emptyset$, is the definition of a Markov blanket of $\mathbf{Y}_{LP}$ in $\mathbf{X}$.

  Finally, we prove that $\mathbf{M}$ is a Markov boundary of $\mathbf{Y}_{LP}$ in $\mathbf{X}$. Let us suppose that it is not the case, i.e., $\exists \mathbf{Z} \subset \mathbf{M}$ such that
  $\mathbf{Y}_{LP} \indep \mathbf{Z} \cup (\mathbf{X} \setminus \mathbf{M}) \mid \mathbf{M} \setminus \mathbf{Z}$
  . From the label powerset assumption of $\mathbf{Y}_{LP}$, we have
  $\mathbf{Y}_{LP} \indep \mathbf{Y} \setminus \mathbf{Y}_{LP} \mid \mathbf{X}$
  which can be rewritten as
  $\mathbf{Y}_{LP} \indep \mathbf{Y} \setminus \mathbf{Y}_{LP} \mid (\mathbf{X} \setminus \mathbf{M}) \cup (\mathbf{M} \setminus \mathbf{Z}) \cup \mathbf{Z}$
  . Due to the Contraction property, combining these two statements yields
  $\mathbf{Y}_{LP} \indep \mathbf{Z} \cup (\mathbf{U} \setminus (\mathbf{M} \cup \mathbf{Y}_{LP}) \mid \mathbf{M} \setminus \mathbf{Z}$
  . From the Markov boundary assumption of $\mathbf{M}$ in $\mathbf{U}$, we have necessarily $\mathbf{Z} = \emptyset$, which suffices to prove that $\mathbf{M}$ is a Markov boundary of $\mathbf{Y}_{LP}$ in $\mathbf{X}$.
\end{pf}

\begin{figure*}
  \centering
  \subfloat[][Image]{
    \centering 
    \includegraphics[clip, trim = 0 0 0 0in, height=0.46\textwidth, width=0.46\textwidth]{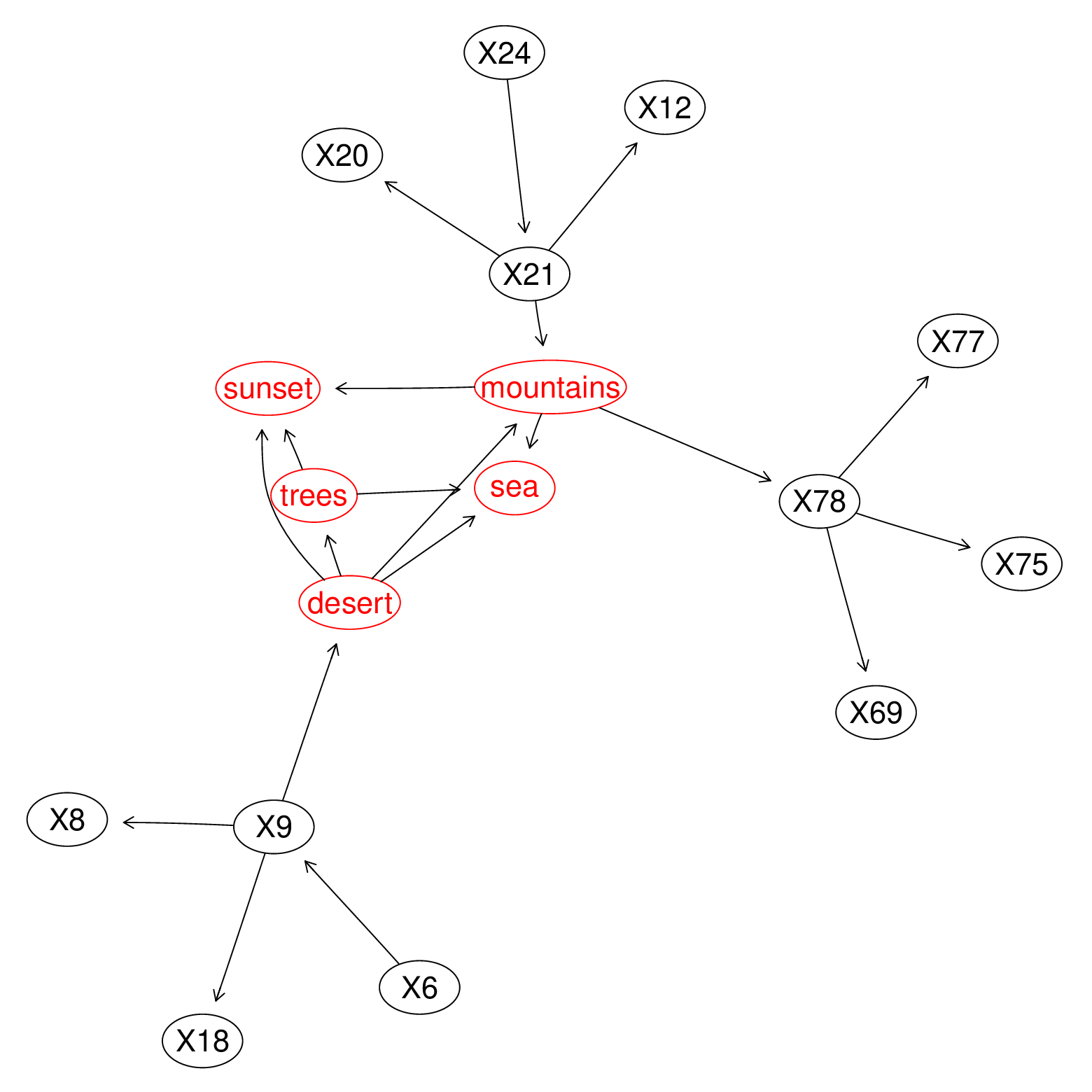}
    \includegraphics[clip, trim = 0 0 0 0in, height=0.46\textwidth, width=0.46\textwidth]{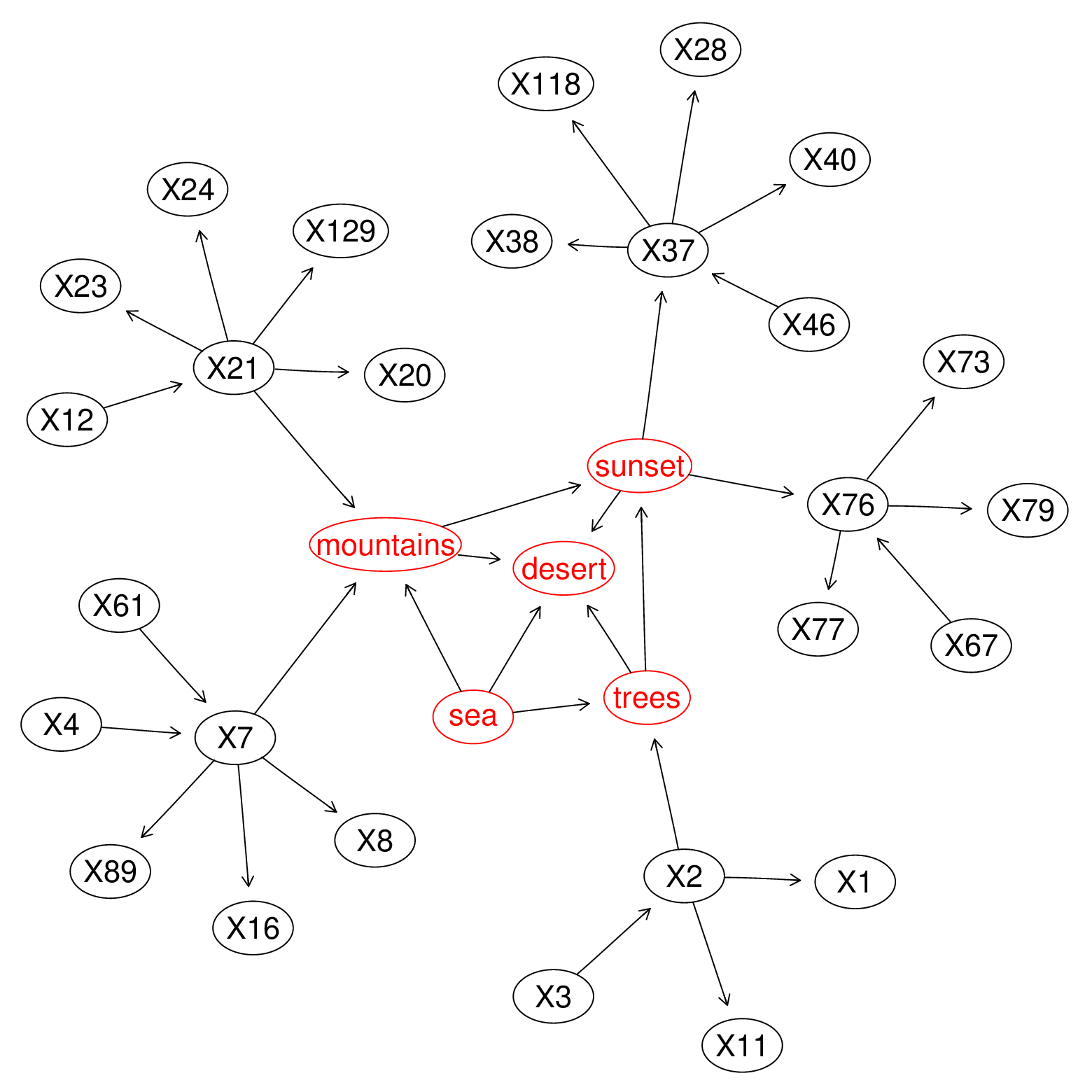}
  } \\
  \subfloat[][Scene]{
    \centering 
    \includegraphics[clip, trim = 0 0 0 0in, height=0.46\textwidth, width=0.46\textwidth]{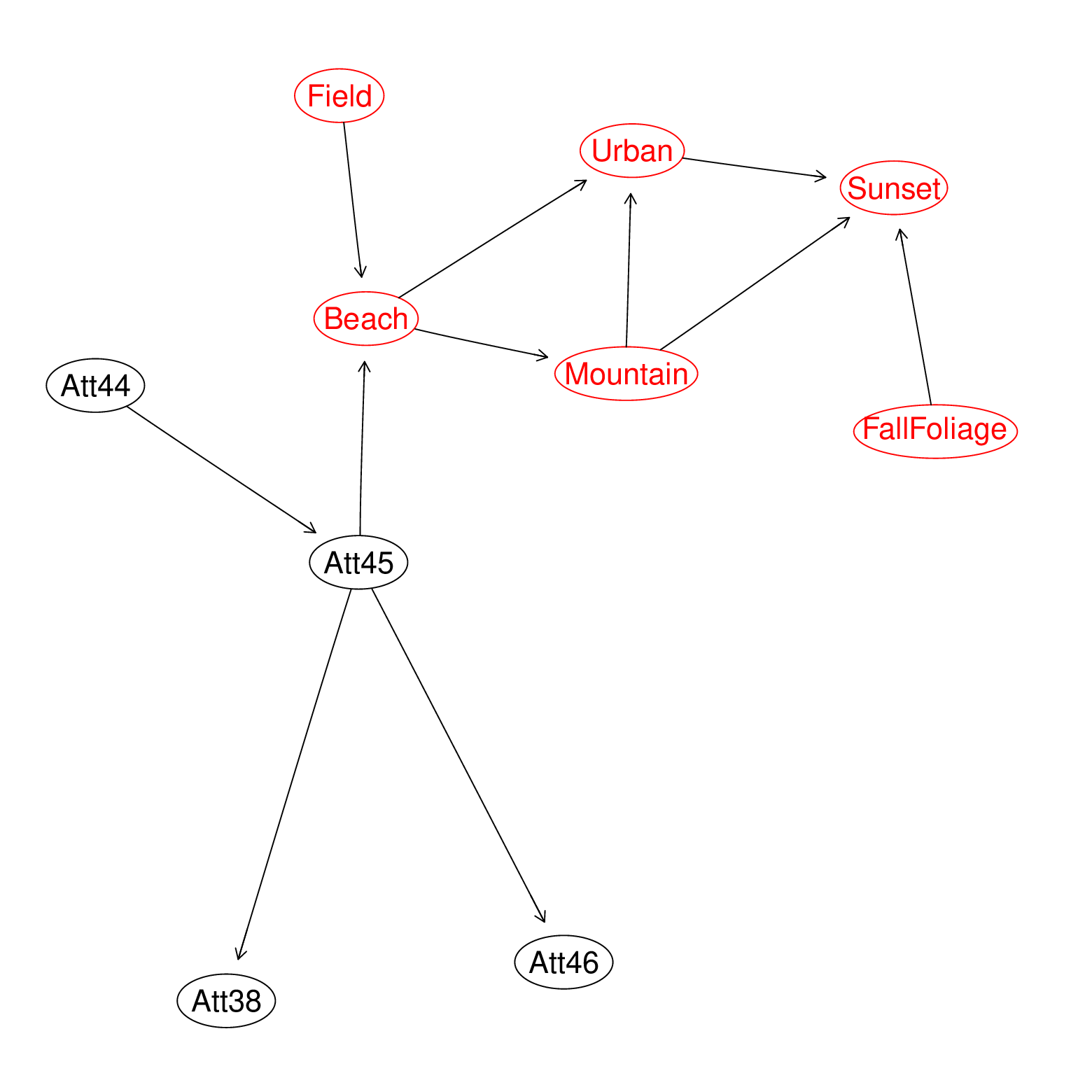}
    \includegraphics[clip, trim = 0 0 0 0in, height=0.46\textwidth, width=0.46\textwidth]{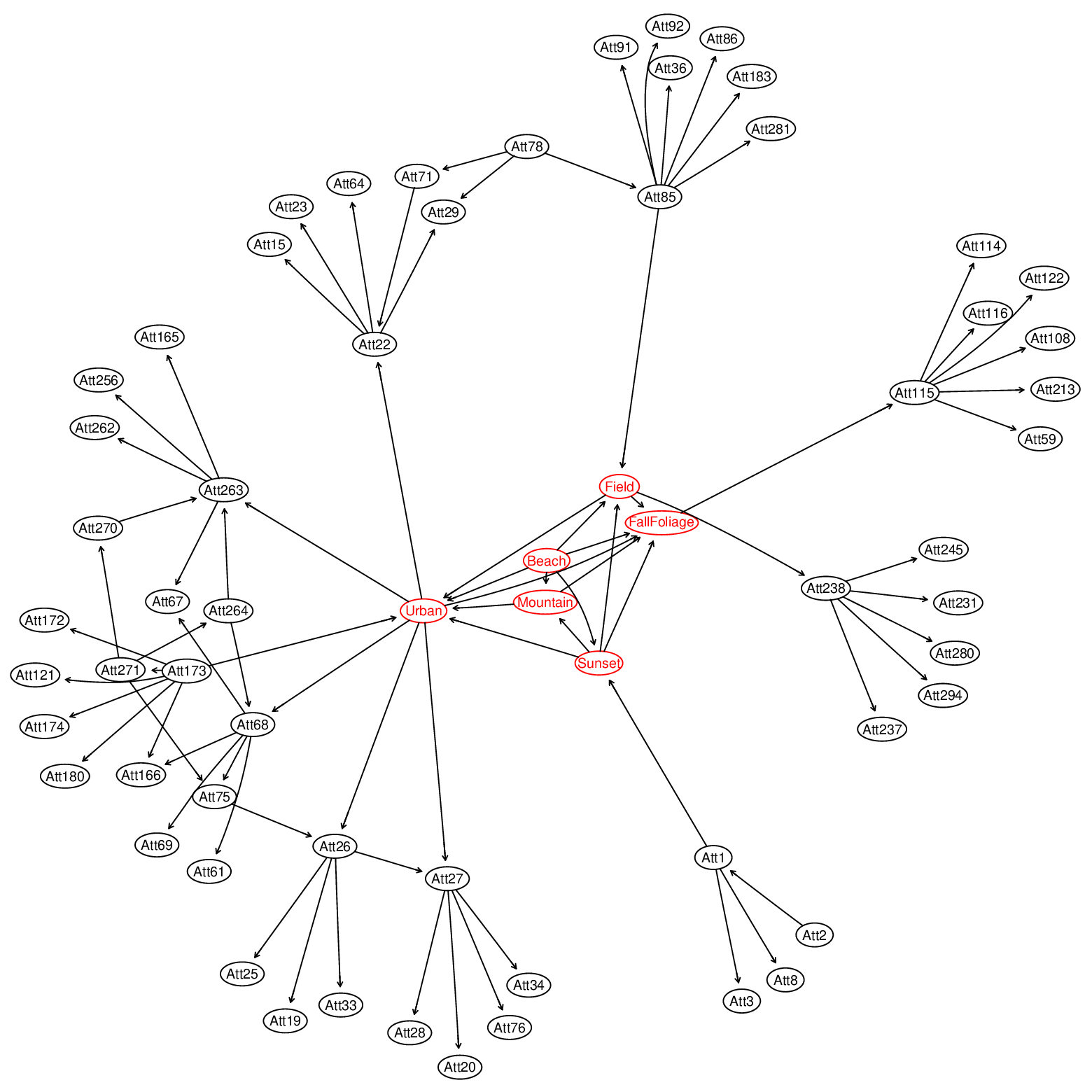}
  }
	  \caption{The local BN structures learned by MMHC (left plot) and H2PC (right plot) on a single cross-validation split,  on Image and Scene.}
  \label{fig:learned-dags-2}
\end{figure*}

\begin{figure*}
  \centering
  \subfloat[][Slashdot]{
    \centering 
    \includegraphics[clip, trim = 0 0 0 0in, height=0.46\textwidth, width=0.46\textwidth]{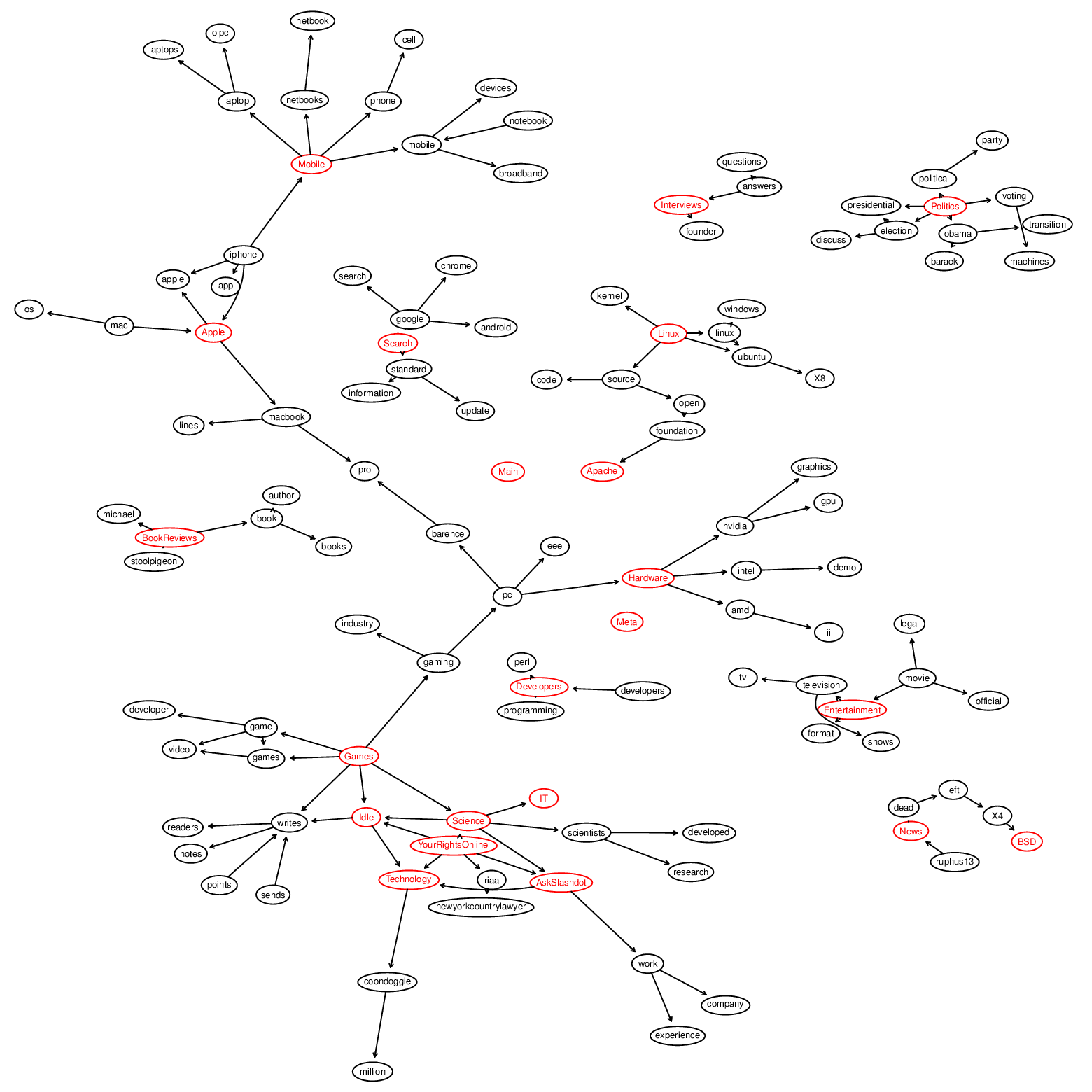}
    \includegraphics[clip, trim = 0 0 0 0in, height=0.46\textwidth, width=0.46\textwidth]{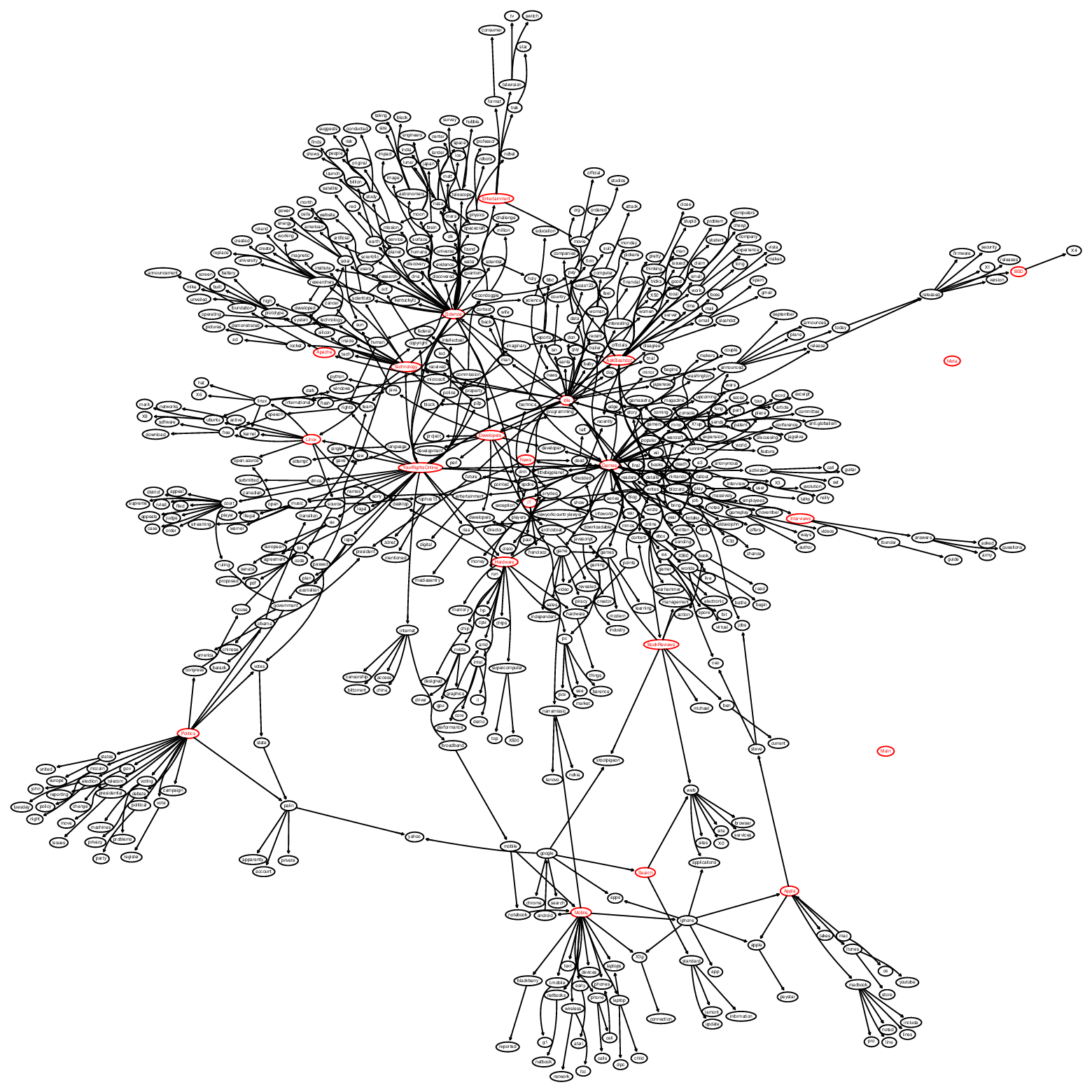}
  } \\
  \centering
  \subfloat[][Genbase]{
    \centering 
    \includegraphics[clip, trim = 0 0 0 0in, height=0.46\textwidth, width=0.46\textwidth]{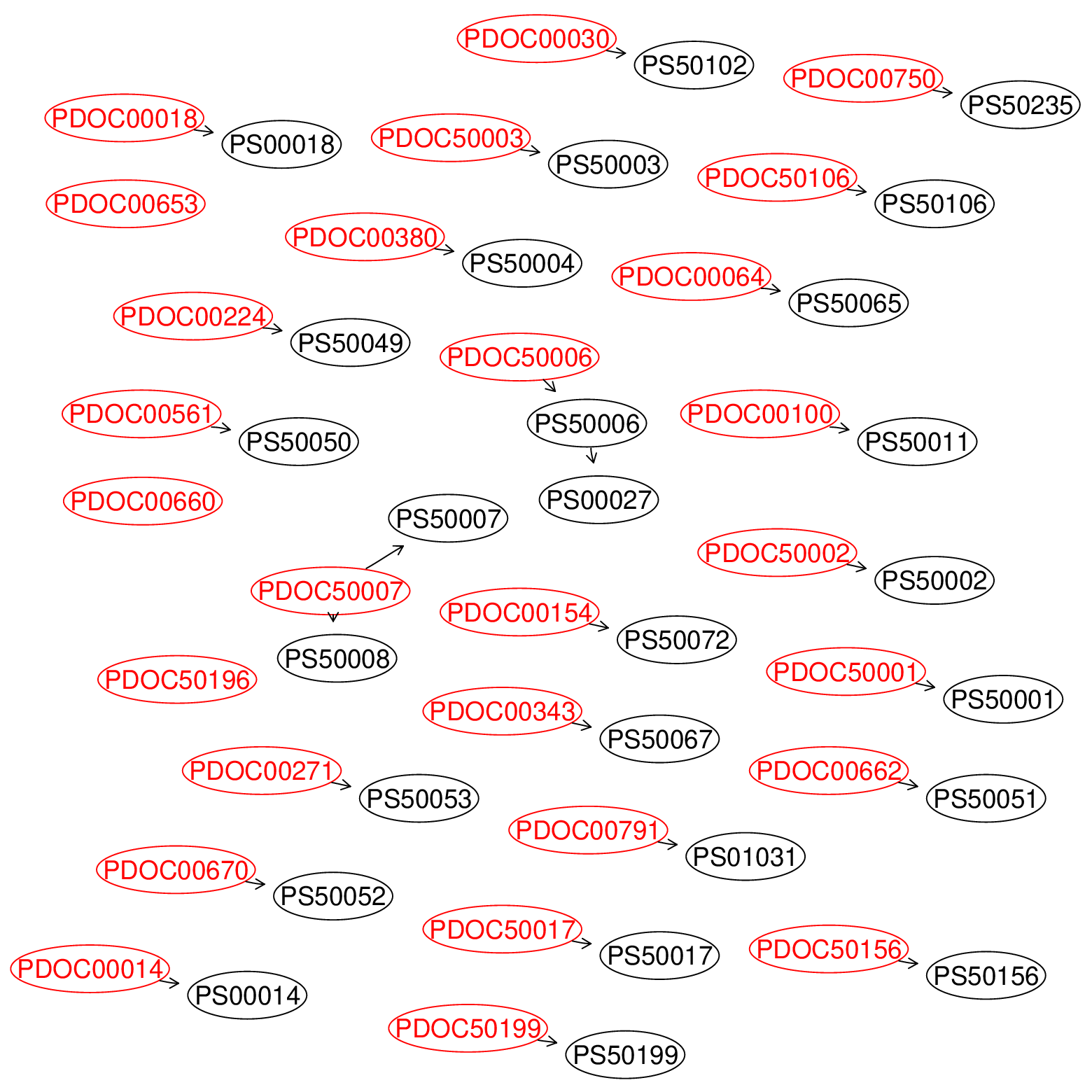}
    \includegraphics[clip, trim = 0 0 0 0in, height=0.46\textwidth, width=0.46\textwidth]{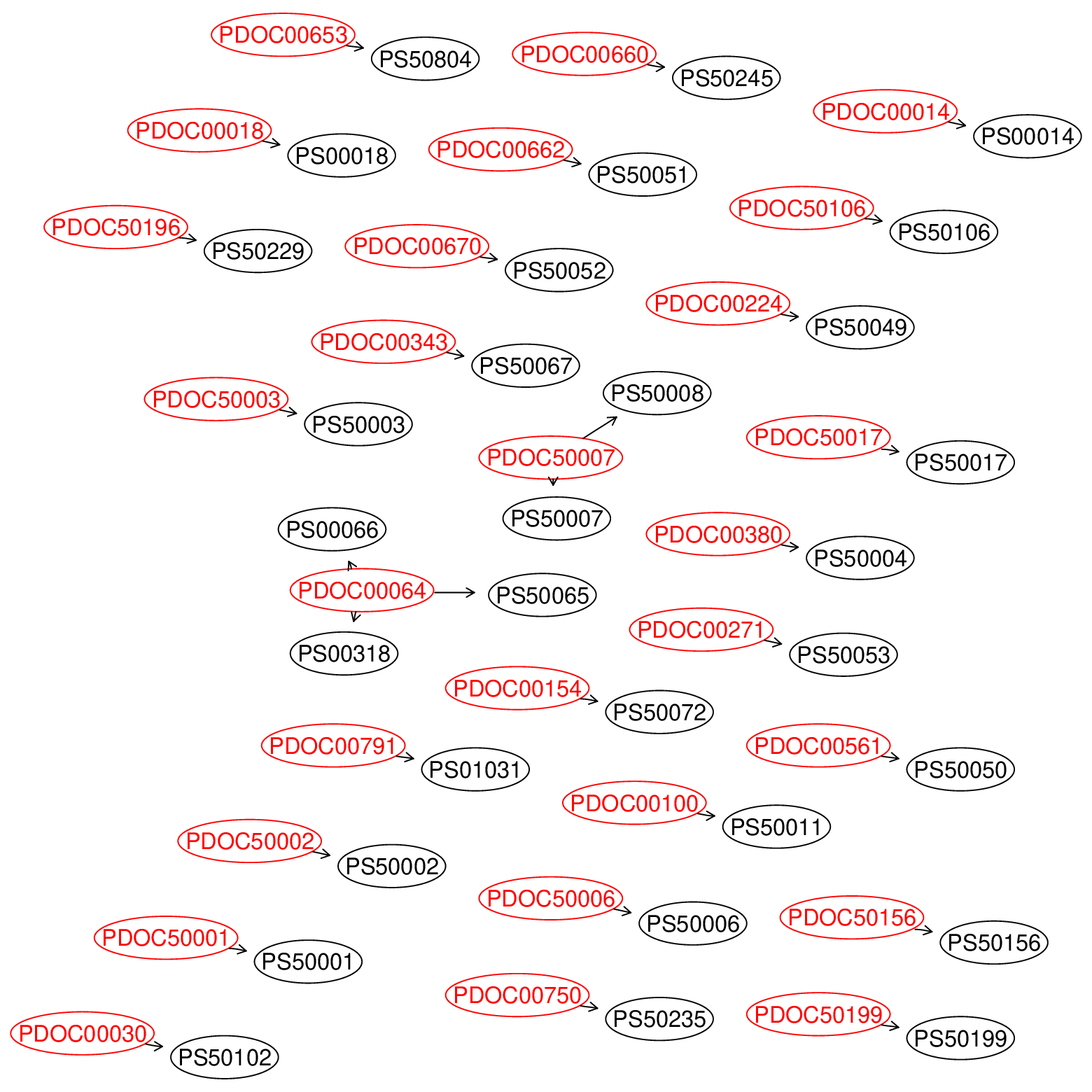}
  }
	  \caption{The local BN structures learned by MMHC (left plot) and H2PC (right plot) on a single cross-validation split, on Slashdot and Genbase.}
  \label{fig:learned-dags-3}
\end{figure*}

\begin{figure*}
  \centering
  \subfloat[][Medical]{
    \centering 
    \includegraphics[clip, trim = 0 0 0 0in, height=0.46\textwidth, width=0.46\textwidth]{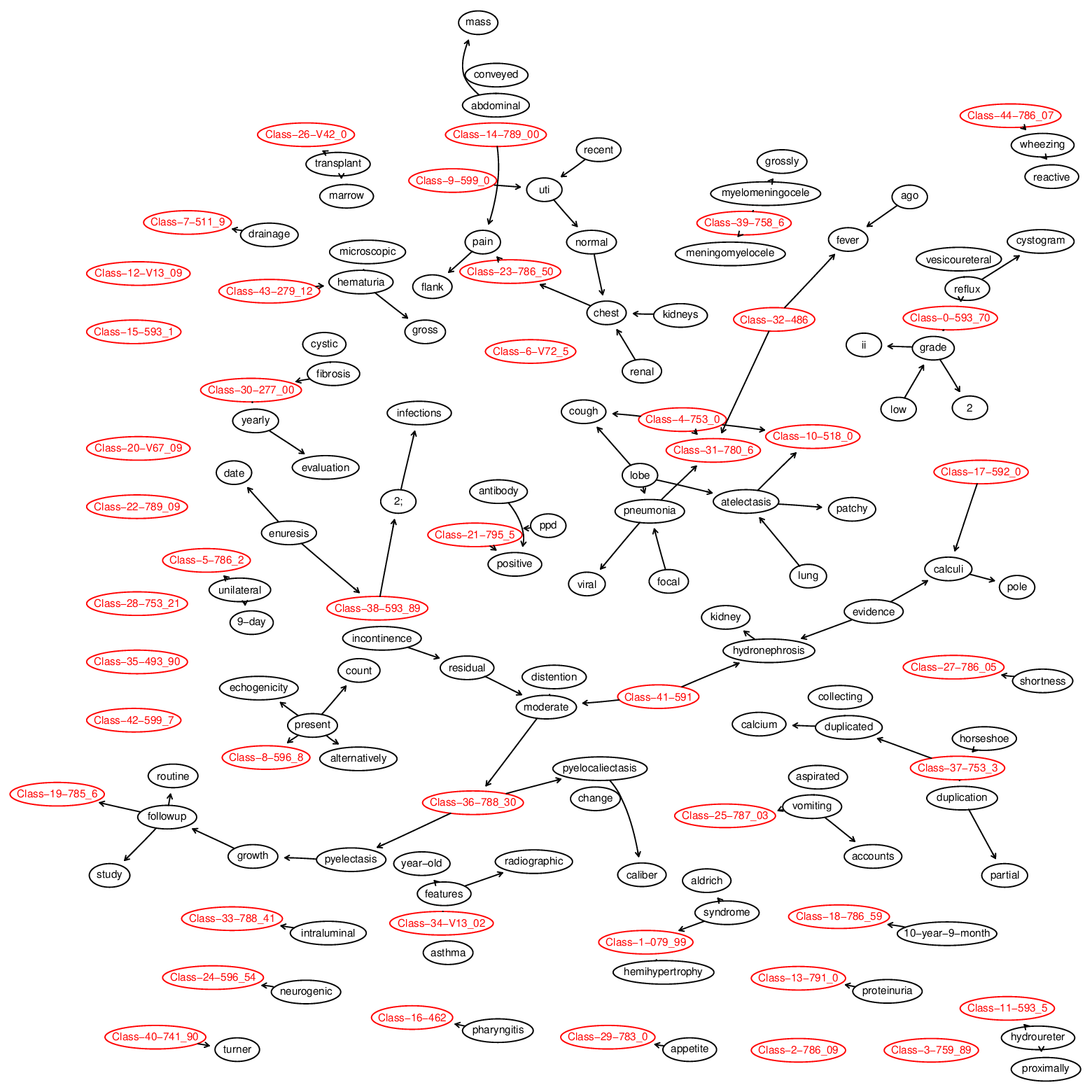}
    \includegraphics[clip, trim = 0 0 0 0in, height=0.46\textwidth, width=0.46\textwidth]{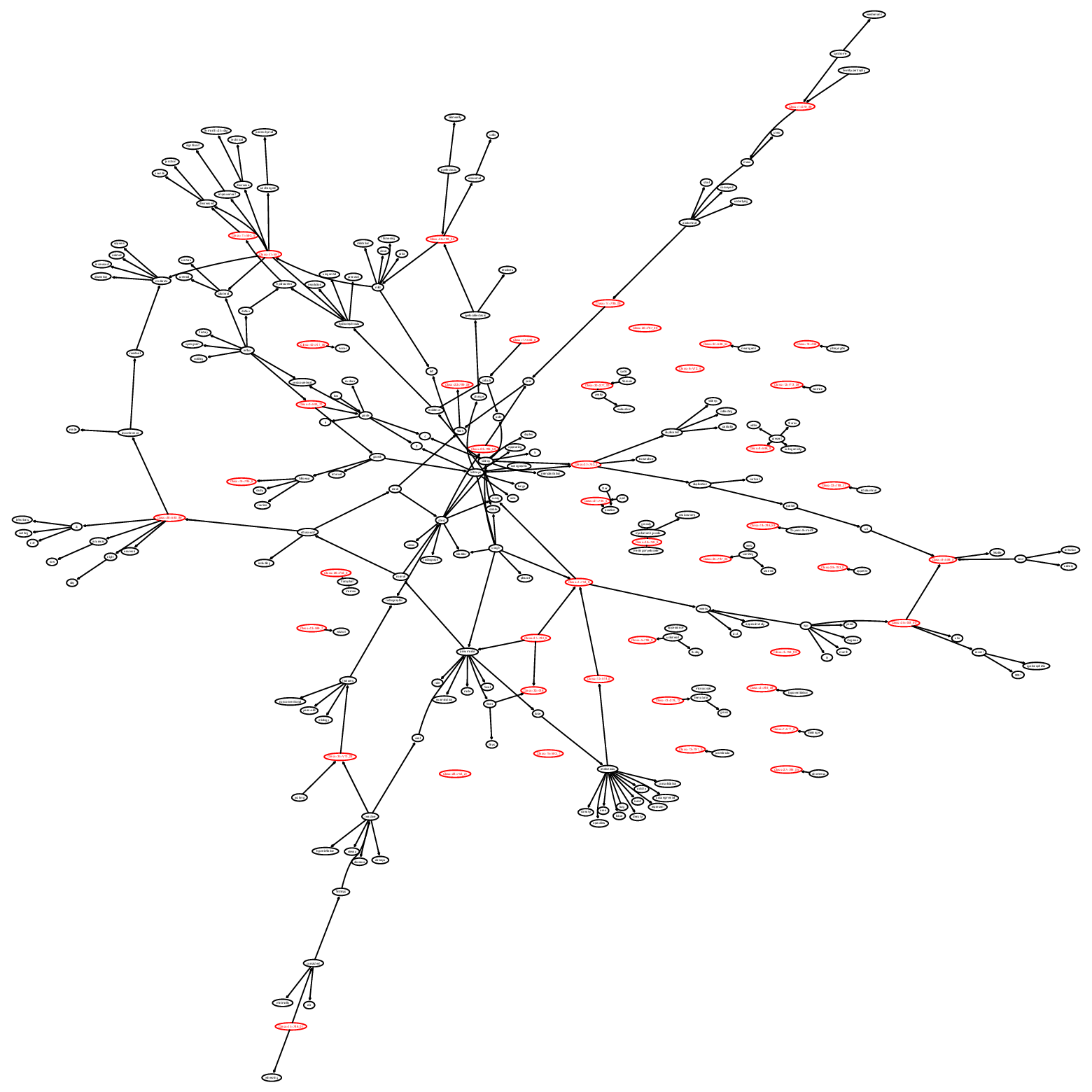}
  }\\
  \subfloat[][Enron]{
    \centering 
    \includegraphics[clip, trim = 0 0 0 0in, height=0.46\textwidth, width=0.46\textwidth]{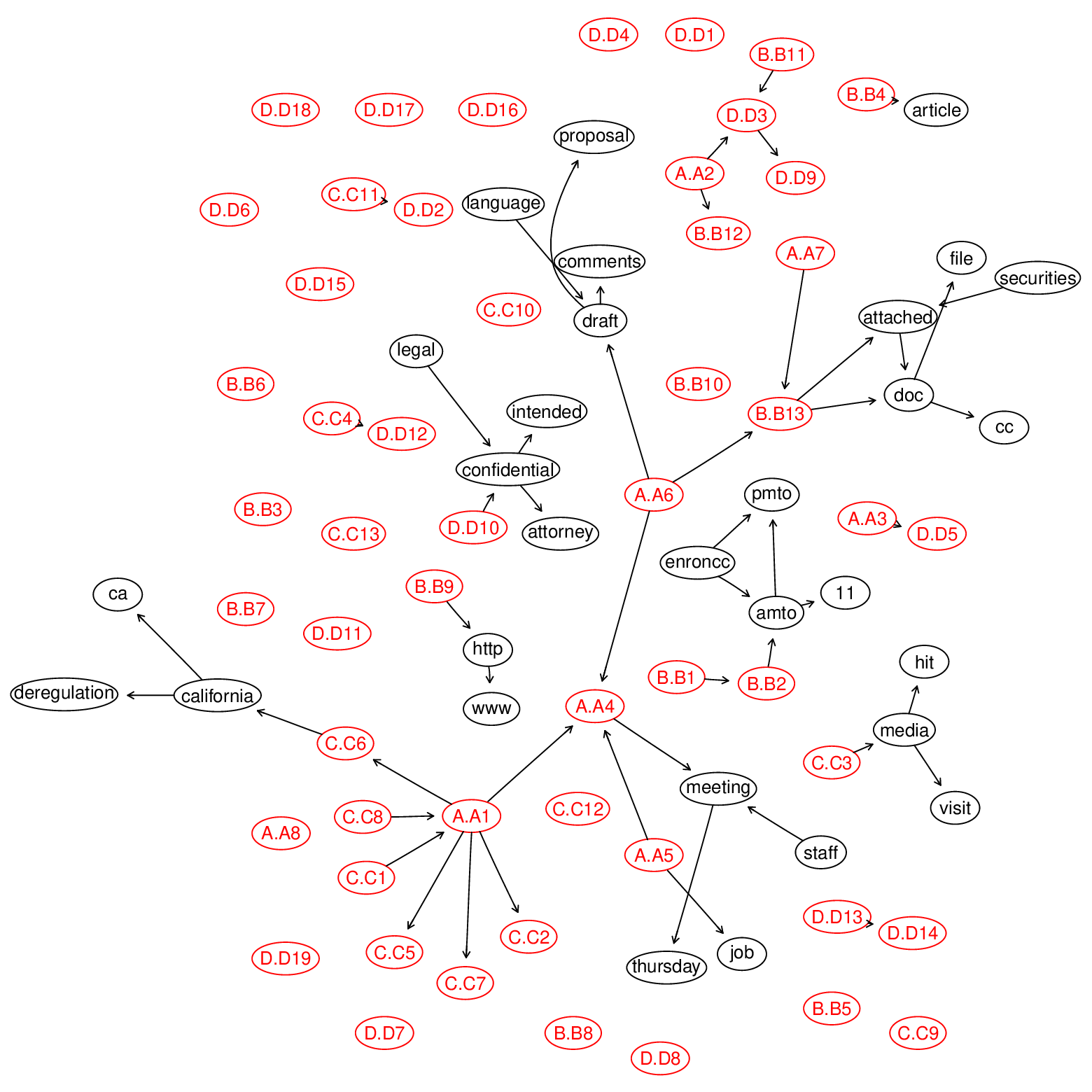}
    \includegraphics[clip, trim = 0 0 0 0in, height=0.46\textwidth, width=0.46\textwidth]{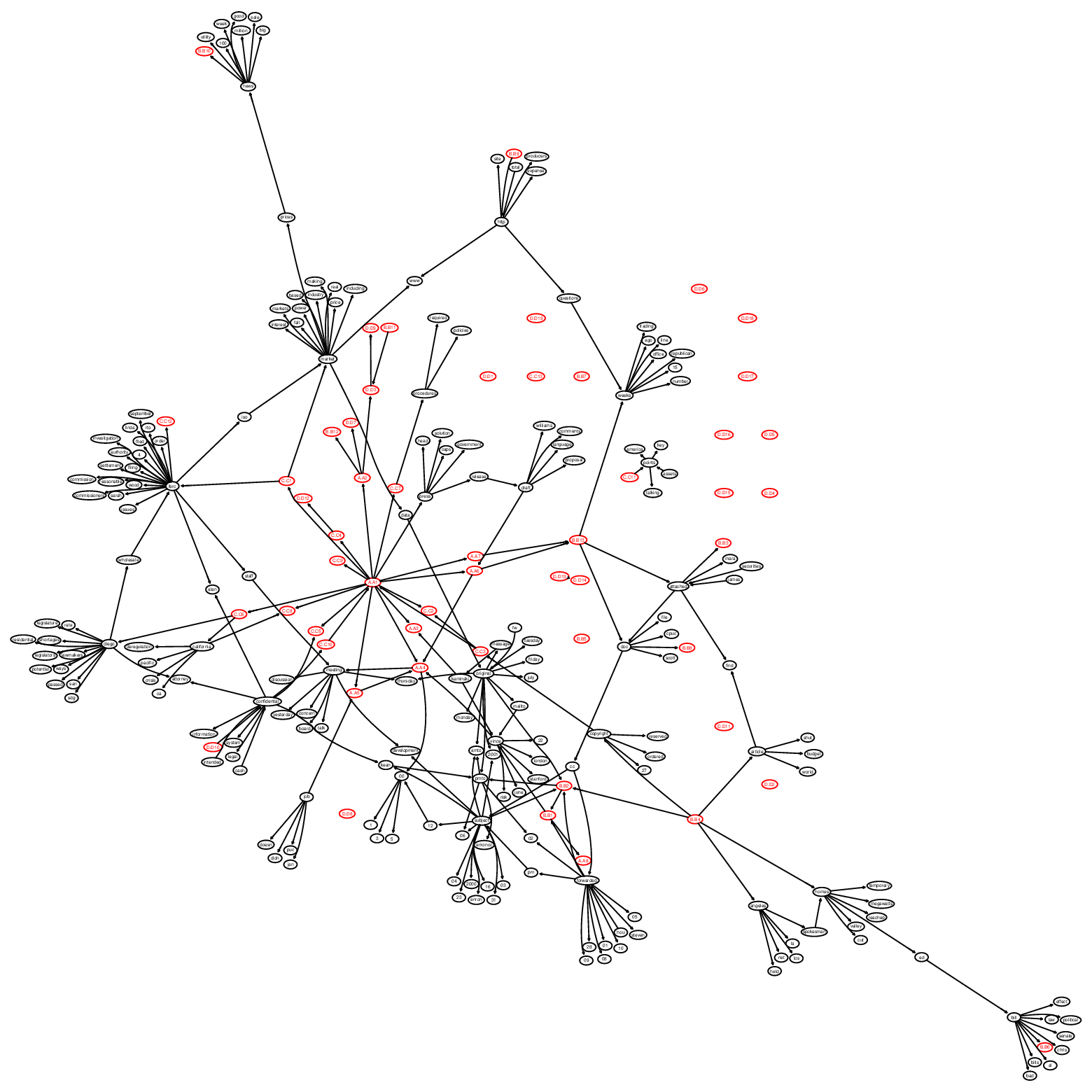}
  }
		  \caption{The local BN structures learned by MMHC (left plot) and H2PC (right plot) on a single cross-validation split, on Medical and Enron.}
  \label{fig:learned-dags-4}
\end{figure*}

\newpage
\bibliographystyle{model5-names}
\bibliography{biblio}

\begin{thebibliography}{75}
\expandafter\ifx\csname natexlab\endcsname\relax\def\natexlab#1{#1}\fi
\providecommand{\url}[1]{\texttt{#1}}
\providecommand{\href}[2]{#2}
\providecommand{\path}[1]{#1}
\providecommand{\DOIprefix}{doi:}
\providecommand{\ArXivprefix}{arXiv:}
\providecommand{\URLprefix}{URL: }
\providecommand{\Pubmedprefix}{pmid:}
\providecommand{\doi}[1]{\href{http://dx.doi.org/#1}{\path{#1}}}
\providecommand{\Pubmed}[1]{\href{pmid:#1}{\path{#1}}}
\providecommand{\bibinfo}[2]{#2}
\ifx\xfnm\relax \def\xfnm[#1]{\unskip,\space#1}\fi
\bibitem[{Agresti(2002)}]{Agresti02}
\bibinfo{author}{Agresti, A.} (\bibinfo{year}{2002}).
\newblock {\it \bibinfo{title}{Categorical Data Analysis}\/}.
\newblock \bibinfo{publisher}{Wiley, 2nd edition}.
\bibitem[{Aliferis et~al.(2010)Aliferis, Statnikov, Tsamardinos, Mani \&
  Koutsoukos}]{Aliferis10a}
\bibinfo{author}{Aliferis, C.~F.}, \bibinfo{author}{Statnikov, A.~R.},
  \bibinfo{author}{Tsamardinos, I.}, \bibinfo{author}{Mani, S.}, \&
  \bibinfo{author}{Koutsoukos, X.~D.} (\bibinfo{year}{2010}).
\newblock \bibinfo{title}{Local causal and markov blanket induction for causal
  discovery and feature selection for classification part i: Algorithms and
  empirical evaluation}.
\newblock {\it \bibinfo{journal}{Journal of Machine Learning Research,
  JMLR}\/},  {\it \bibinfo{volume}{11}\/}, \bibinfo{pages}{171--234}.
\bibitem[{Alvares-Cherman et~al.(2011)Alvares-Cherman, Metz \&
  Monard}]{Alvares-Cherman2011}
\bibinfo{author}{Alvares-Cherman, E.}, \bibinfo{author}{Metz, J.}, \&
  \bibinfo{author}{Monard, M.} (\bibinfo{year}{2011}).
\newblock \bibinfo{title}{{Incorporating label dependency into the binary
  relevance framework for multi-label classification}}.
\newblock {\it \bibinfo{journal}{Expert Systems With Applications, ESWA}\/},
  {\it \bibinfo{volume}{39}\/}, \bibinfo{pages}{1647--1655}.
\bibitem[{Armen \& Tsamardinos(2011)}]{ArmenT11}
\bibinfo{author}{Armen, A.~P.}, \& \bibinfo{author}{Tsamardinos, I.}
  (\bibinfo{year}{2011}).
\newblock \bibinfo{title}{A unified approach to estimation and control of the
  false discovery rate in bayesian network skeleton identification}.
\newblock In {\it \bibinfo{booktitle}{European Symposium on Artificial Neural
  Networks, ESANN}\/}.
\bibitem[{Aussem et~al.(2012)Aussem, {Rodrigues de Morais} \&
  Corbex}]{Aussem12a}
\bibinfo{author}{Aussem, A.}, \bibinfo{author}{{Rodrigues de Morais}, S.}, \&
  \bibinfo{author}{Corbex, M.} (\bibinfo{year}{2012}).
\newblock \bibinfo{title}{Analysis of nasopharyngeal carcinoma risk factors
  with bayesian networks}.
\newblock {\it \bibinfo{journal}{Artificial Intelligence in Medicine}\/},  {\it
  \bibinfo{volume}{54}\/}.
\bibitem[{Aussem et~al.(2010)Aussem, Tchernof, {Rodrigues de Morais} \&
  Rome}]{Aussem10c}
\bibinfo{author}{Aussem, A.}, \bibinfo{author}{Tchernof, A.},
  \bibinfo{author}{{Rodrigues de Morais}, S.}, \& \bibinfo{author}{Rome, S.}
  (\bibinfo{year}{2010}).
\newblock \bibinfo{title}{Analysis of lifestyle and metabolic predictors of
  visceral obesity with bayesian networks}.
\newblock {\it \bibinfo{journal}{BMC Bioinformatics}\/},  {\it
  \bibinfo{volume}{11}\/}, \bibinfo{pages}{487}.
\bibitem[{Badea(2004)}]{Badea2004}
\bibinfo{author}{Badea, A.} (\bibinfo{year}{2004}).
\newblock \bibinfo{title}{Determining the direction of causal influence in
  large probabilistic networks: A constraint-based approach.}
\newblock In {\it \bibinfo{booktitle}{Proceedings of the Sixteenth European
  Conference on Artificial Intelligence}\/} (pp. \bibinfo{pages}{263--267}).
\bibitem[{Bernard \& Hartemink(2005)}]{Bernard2005}
\bibinfo{author}{Bernard, A.}, \& \bibinfo{author}{Hartemink, A.}
  (\bibinfo{year}{2005}).
\newblock \bibinfo{title}{Informative structure priors: Joint learning of
  dynamic regulatory networks from multiple types of data.}
\newblock In {\it \bibinfo{booktitle}{Proceedings of the Pacific Symposium on
  Biocomputing}\/} (pp. \bibinfo{pages}{459--470}).
\bibitem[{Blockeel et~al.(1998)Blockeel, Raedt \& Ramon}]{Blockeel1998}
\bibinfo{author}{Blockeel, H.}, \bibinfo{author}{Raedt, L.~D.}, \&
  \bibinfo{author}{Ramon, J.} (\bibinfo{year}{1998}).
\newblock \bibinfo{title}{Top-down induction of clustering trees}.
\newblock In \bibinfo{editor}{J.~W. Shavlik} (Ed.), {\it
  \bibinfo{booktitle}{{International Conference on Machine Learning, ICML}}\/}
  (pp. \bibinfo{pages}{55--63}).
\newblock \bibinfo{publisher}{Morgan Kaufmann}.
\bibitem[{Borchani et~al.(2013)Borchani, Bielza, Toro \&
  Larra{\~n}aga}]{Borchani13}
\bibinfo{author}{Borchani, H.}, \bibinfo{author}{Bielza, C.},
  \bibinfo{author}{Toro, C.}, \& \bibinfo{author}{Larra{\~n}aga, P.}
  (\bibinfo{year}{2013}).
\newblock \bibinfo{title}{Predicting human immunodeficiency virus inhibitors
  using multi-dimensional bayesian network classifiers}.
\newblock {\it \bibinfo{journal}{Artificial Intelligence in Medicine}\/},  {\it
  \bibinfo{volume}{57}\/}, \bibinfo{pages}{219--229}.
\bibitem[{Breiman(2001)}]{Breiman01}
\bibinfo{author}{Breiman, L.} (\bibinfo{year}{2001}).
\newblock \bibinfo{title}{Random forests}.
\newblock {\it \bibinfo{journal}{Machine Learning}\/},  {\it
  \bibinfo{volume}{45}\/}, \bibinfo{pages}{5--32}.
\bibitem[{Brown \& Tsamardinos(2008)}]{Brown08}
\bibinfo{author}{Brown, L.~E.}, \& \bibinfo{author}{Tsamardinos, I.}
  (\bibinfo{year}{2008}).
\newblock \bibinfo{title}{A strategy for making predictions under
  manipulation}.
\newblock {\it \bibinfo{journal}{Journal of Machine Learning Research,
  JMLR}\/},  {\it \bibinfo{volume}{3}\/}, \bibinfo{pages}{35--52}.
\bibitem[{Buntine(1991)}]{Bun91}
\bibinfo{author}{Buntine, W.} (\bibinfo{year}{1991}).
\newblock \bibinfo{title}{Theory refinement on {B}ayesian networks}.
\newblock In \bibinfo{editor}{B.~{D'Ambrosio}}, \bibinfo{editor}{P.~Smets}, \&
  \bibinfo{editor}{P.~Bonissone} (Eds.), {\it \bibinfo{booktitle}{Uncertainty
  in Artificial Intelligence, UAI}\/} (pp. \bibinfo{pages}{52--60}).
\newblock \bibinfo{address}{San Mateo, CA, USA}: \bibinfo{publisher}{Morgan
  Kaufmann Publishers}.
\bibitem[{Cawley(2008)}]{Cawley08}
\bibinfo{author}{Cawley, G.} (\bibinfo{year}{2008}).
\newblock \bibinfo{title}{Causal and non-causal feature selection for ridge
  regression}.
\newblock {\it \bibinfo{journal}{JMLR: Workshop and Conference Proceedings}\/},
   {\it \bibinfo{volume}{3}\/}, \bibinfo{pages}{107--128}.
\bibitem[{Cheng et~al.(2002)Cheng, Greiner, Kelly, Bell \& Liu}]{Cheng02}
\bibinfo{author}{Cheng, J.}, \bibinfo{author}{Greiner, R.},
  \bibinfo{author}{Kelly, J.}, \bibinfo{author}{Bell, D.~A.}, \&
  \bibinfo{author}{Liu, W.} (\bibinfo{year}{2002}).
\newblock \bibinfo{title}{Learning {B}ayesian networks from data: An
  information-theory based approach}.
\newblock {\it \bibinfo{journal}{Artificial Intelligence}\/},  {\it
  \bibinfo{volume}{137}\/}, \bibinfo{pages}{43--90}.
\bibitem[{Chickering et~al.(2004)Chickering, Heckerman \&
  Meek}]{ChickeringHM04}
\bibinfo{author}{Chickering, D.}, \bibinfo{author}{Heckerman, D.}, \&
  \bibinfo{author}{Meek, C.} (\bibinfo{year}{2004}).
\newblock \bibinfo{title}{Large-sample learning of {B}ayesian networks is
  {NP}-hard.}
\newblock {\it \bibinfo{journal}{Journal of Machine Learning Research,
  JMLR}\/},  {\it \bibinfo{volume}{5}\/}, \bibinfo{pages}{1287--1330}.
\bibitem[{Chickering(2002)}]{Chickering02}
\bibinfo{author}{Chickering, D.~M.} (\bibinfo{year}{2002}).
\newblock \bibinfo{title}{Optimal structure identification with greedy search}.
\newblock {\it \bibinfo{journal}{Journal of Machine Learning Research,
  JMLR}\/},  {\it \bibinfo{volume}{3}\/}, \bibinfo{pages}{507--554}.
\bibitem[{Cussens \& Bartlett(2013)}]{Cussens2013}
\bibinfo{author}{Cussens, J.}, \& \bibinfo{author}{Bartlett, M.}
  (\bibinfo{year}{2013}).
\newblock \bibinfo{title}{{Advances in Bayesian Network Learning using Integer
  Programming}}.
\newblock In {\it \bibinfo{booktitle}{Uncertainty in Artificial
  Intelligence}\/} (pp. \bibinfo{pages}{182--191}).
\bibitem[{Dembczyński et~al.(2012)Dembczyński, Waegeman, Cheng \&
  H\"{u}llermeier}]{Dembczynski2012}
\bibinfo{author}{Dembczyński, K.}, \bibinfo{author}{Waegeman, W.},
  \bibinfo{author}{Cheng, W.}, \& \bibinfo{author}{H\"{u}llermeier, E.}
  (\bibinfo{year}{2012}).
\newblock \bibinfo{title}{{On label dependence and loss minimization in
  multi-label classification}}.
\newblock {\it \bibinfo{journal}{Machine Learning}\/},  {\it
  \bibinfo{volume}{88}\/}, \bibinfo{pages}{5--45}.
\bibitem[{Ellis \& Wong(2008)}]{Byron08}
\bibinfo{author}{Ellis, B.}, \& \bibinfo{author}{Wong, W.~H.}
  (\bibinfo{year}{2008}).
\newblock \bibinfo{title}{Learning causal bayesian network structures from
  experimental data}.
\newblock {\it \bibinfo{journal}{Journal of the American Statistical
  Association}\/},  {\it \bibinfo{volume}{103}\/}, \bibinfo{pages}{778--789}.
\bibitem[{Friedman et~al.(1999{\natexlab{a}})Friedman, Nachman,  \&
  Pe’er}]{Friedman1999}
\bibinfo{author}{Friedman, N.}, \bibinfo{author}{Nachman, I.}, , \&
  \bibinfo{author}{Pe’er, D.} (\bibinfo{year}{1999}{\natexlab{a}}).
\newblock \bibinfo{title}{Learning bayesian network structure from massive
  datasets: The “sparse candidate” algorithm.}
\newblock In {\it \bibinfo{booktitle}{Proceedings of the Fifteenth Conference
  on Uncertainty in Artificial Intelligence}\/} (pp.
  \bibinfo{pages}{206--215}).
\bibitem[{Friedman et~al.(1999{\natexlab{b}})Friedman, Nachman \&
  Pe'er}]{Friedman99}
\bibinfo{author}{Friedman, N.}, \bibinfo{author}{Nachman, I.}, \&
  \bibinfo{author}{Pe'er, D.} (\bibinfo{year}{1999}{\natexlab{b}}).
\newblock \bibinfo{title}{Learning bayesian network structure from massive
  datasets: the "sparse candidate" algorithm}.
\newblock In \bibinfo{editor}{K.~B. Laskey}, \& \bibinfo{editor}{H.~Prade}
  (Eds.), {\it \bibinfo{booktitle}{Uncertainty in Artificial Intelligence,
  UAI}\/} (pp. \bibinfo{pages}{21--30}).
\newblock \bibinfo{publisher}{Morgan Kaufmann Publishers}.
\bibitem[{Gasse et~al.(2012)Gasse, Aussem \& Elghazel}]{Gasse12}
\bibinfo{author}{Gasse, M.}, \bibinfo{author}{Aussem, A.}, \&
  \bibinfo{author}{Elghazel, H.} (\bibinfo{year}{2012}).
\newblock \bibinfo{title}{Comparison of hybrid algorithms for bayesian network
  structure learning}.
\newblock In {\it \bibinfo{booktitle}{European Conference on Machine Learning
  and Principles and Practice of Knowledge Discovery in Databases,
  ECML-PKDD}\/} (pp. \bibinfo{pages}{58--73}).
\newblock \bibinfo{publisher}{Springer} volume \bibinfo{volume}{7523} of {\it
  \bibinfo{series}{Lecture Notes in Computer Science}\/}.
\bibitem[{Gu et~al.(2011)Gu, Li \& Han}]{Gu2011}
\bibinfo{author}{Gu, Q.}, \bibinfo{author}{Li, Z.}, \& \bibinfo{author}{Han,
  J.} (\bibinfo{year}{2011}).
\newblock \bibinfo{title}{Correlated multi-label feature selection}.
\newblock In \bibinfo{editor}{C.~Macdonald}, \bibinfo{editor}{I.~Ounis}, \&
  \bibinfo{editor}{I.~Ruthven} (Eds.), {\it \bibinfo{booktitle}{CIKM}\/} (pp.
  \bibinfo{pages}{1087--1096}).
\newblock \bibinfo{publisher}{ACM}.
\bibitem[{Guo \& Gu(2011)}]{Guo11}
\bibinfo{author}{Guo, Y.}, \& \bibinfo{author}{Gu, S.} (\bibinfo{year}{2011}).
\newblock \bibinfo{title}{{Multi-label classification using conditional
  dependency networks}}.
\newblock In {\it \bibinfo{booktitle}{{International Joint Conference on
  Artificial Intelligence, IJCAI}}\/} (pp. \bibinfo{pages}{1300--1305}).
\newblock \bibinfo{publisher}{AAAI Press}.
\bibitem[{Heckerman et~al.({1995})Heckerman, Geiger \&
  Chickering}]{Heckerman95}
\bibinfo{author}{Heckerman, D.}, \bibinfo{author}{Geiger, D.}, \&
  \bibinfo{author}{Chickering, D.} (\bibinfo{year}{{1995}}).
\newblock \bibinfo{title}{Learning bayesian networks: The combination of
  knowledge and statistical data}.
\newblock {\it \bibinfo{journal}{Machine Learning}\/},  {\it
  \bibinfo{volume}{{20}}\/}, \bibinfo{pages}{{197--243}}.
\bibitem[{Kocev et~al.(2007)Kocev, Vens, Struyf \& D\v{z}eroski}]{Kocev2007}
\bibinfo{author}{Kocev, D.}, \bibinfo{author}{Vens, C.},
  \bibinfo{author}{Struyf, J.}, \& \bibinfo{author}{D\v{z}eroski, S.}
  (\bibinfo{year}{2007}).
\newblock \bibinfo{title}{{Ensembles of multi-objective decision trees}}.
\newblock In \bibinfo{editor}{J.~N. Kok} (Ed.), {\it
  \bibinfo{booktitle}{European Conference on Machine Learning, ECML}\/} (pp.
  \bibinfo{pages}{624--631}).
\newblock \bibinfo{publisher}{Springer} volume \bibinfo{volume}{4701} of {\it
  \bibinfo{series}{Lecture Notes in Artificial Intelligence}\/}.
\bibitem[{Koivisto \& Sood(2004)}]{Koivisto04}
\bibinfo{author}{Koivisto, M.}, \& \bibinfo{author}{Sood, K.}
  (\bibinfo{year}{2004}).
\newblock \bibinfo{title}{Exact bayesian structure discovery in bayesian
  networks}.
\newblock {\it \bibinfo{journal}{Journal of Machine Learning Research,
  JMLR}\/},  {\it \bibinfo{volume}{5}\/}, \bibinfo{pages}{549--573}.
\bibitem[{Kojima et~al.(2010)Kojima, Perrier, Imoto \& Miyano}]{Kojima10}
\bibinfo{author}{Kojima, K.}, \bibinfo{author}{Perrier, E.},
  \bibinfo{author}{Imoto, S.}, \& \bibinfo{author}{Miyano, S.}
  (\bibinfo{year}{2010}).
\newblock \bibinfo{title}{Optimal search on clustered structural constraint for
  learning bayesian network structure}.
\newblock {\it \bibinfo{journal}{Journal of Machine Learning Research,
  JMLR}\/},  {\it \bibinfo{volume}{11}\/}, \bibinfo{pages}{285--310}.
\bibitem[{Koller \& Friedman(2009)}]{Koller09}
\bibinfo{author}{Koller, D.}, \& \bibinfo{author}{Friedman, N.}
  (\bibinfo{year}{2009}).
\newblock {\it \bibinfo{title}{Probabilistic Graphical Models: Principles and
  Techniques}\/}.
\newblock \bibinfo{publisher}{MIT Press}.
\bibitem[{Koller \& Sahami(1996)}]{Koller96}
\bibinfo{author}{Koller, D.}, \& \bibinfo{author}{Sahami, M.}
  (\bibinfo{year}{1996}).
\newblock \bibinfo{title}{Toward optimal feature selection}.
\newblock In {\it \bibinfo{booktitle}{International Conference on Machine
  Learning, ICML}\/} (pp. \bibinfo{pages}{284--292}).
\bibitem[{Liaw \& Wiener(2002)}]{randomForest}
\bibinfo{author}{Liaw, A.}, \& \bibinfo{author}{Wiener, M.}
  (\bibinfo{year}{2002}).
\newblock \bibinfo{title}{Classification and regression by randomforest}.
\newblock {\it \bibinfo{journal}{R News}\/},  {\it \bibinfo{volume}{2}\/},
  \bibinfo{pages}{18--22}.
\bibitem[{Luaces et~al.(2012)Luaces, D\'{\i}ez, Barranquero, del Coz \&
  Bahamonde}]{Luaces2012}
\bibinfo{author}{Luaces, O.}, \bibinfo{author}{D\'{\i}ez, J.},
  \bibinfo{author}{Barranquero, J.}, \bibinfo{author}{del Coz, J.~J.}, \&
  \bibinfo{author}{Bahamonde, A.} (\bibinfo{year}{2012}).
\newblock \bibinfo{title}{Binary relevance efficacy for multilabel
  classification}.
\newblock {\it \bibinfo{journal}{Progress in AI}\/},  {\it
  \bibinfo{volume}{1}\/}, \bibinfo{pages}{303--313}.
\bibitem[{Madjarov et~al.(2012)Madjarov, Kocev, Gjorgjevikj \&
  D\v{z}eroski}]{Madjarov2012}
\bibinfo{author}{Madjarov, G.}, \bibinfo{author}{Kocev, D.},
  \bibinfo{author}{Gjorgjevikj, D.}, \& \bibinfo{author}{D\v{z}eroski, S.}
  (\bibinfo{year}{2012}).
\newblock \bibinfo{title}{{An extensive experimental comparison of methods for
  multi-label learning}}.
\newblock {\it \bibinfo{journal}{Pattern Recognition}\/},  {\it
  \bibinfo{volume}{45}\/}, \bibinfo{pages}{3084--3104}.
\bibitem[{Maron \& Ratan(1998)}]{Maron1998}
\bibinfo{author}{Maron, O.}, \& \bibinfo{author}{Ratan, A.~L.}
  (\bibinfo{year}{1998}).
\newblock \bibinfo{title}{{Multiple-Instance Learning for Natural Scene
  Classification}}.
\newblock In {\it \bibinfo{booktitle}{International Conference on Machine
  Learning, ICML}\/} (pp. \bibinfo{pages}{341--349}).
\newblock \bibinfo{publisher}{Citeseer} volume~\bibinfo{volume}{7}.
\bibitem[{McCallum(1999)}]{McCallum99}
\bibinfo{author}{McCallum, A.} (\bibinfo{year}{1999}).
\newblock \bibinfo{title}{Multi-label text classification with a mixture model
  trained by em}.
\newblock In {\it \bibinfo{booktitle}{AAAI Workshop on Text Learning}\/}.
\bibitem[{Moore \& Wong(2003)}]{Moore03}
\bibinfo{author}{Moore, A.}, \& \bibinfo{author}{Wong, W.}
  (\bibinfo{year}{2003}).
\newblock \bibinfo{title}{Optimal reinsertion: A new search operator for
  accelerated and more accurate {B}ayesian network structure learning}.
\newblock In \bibinfo{editor}{T.~Fawcett}, \& \bibinfo{editor}{N.~Mishra}
  (Eds.), {\it \bibinfo{booktitle}{International Conference on Machine
  Learning, ICML}\/}.
\bibitem[{Nagarajan et~al.(2013)Nagarajan, Scutari \& Lèbre}]{Nagarajan13}
\bibinfo{author}{Nagarajan, R.}, \bibinfo{author}{Scutari, M.}, \&
  \bibinfo{author}{Lèbre, S.} (\bibinfo{year}{2013}).
\newblock {\it \bibinfo{title}{Bayesian Networks in R: with Applications in
  Systems Biology (Use R!)}\/}.
\newblock \bibinfo{publisher}{Springer}.
\bibitem[{Neapolitan(2004)}]{Neapolitan04}
\bibinfo{author}{Neapolitan, R.~E.} (\bibinfo{year}{2004}).
\newblock {\it \bibinfo{title}{Learning {B}ayesian Networks}\/}.
\newblock \bibinfo{address}{Upper Saddle River, NJ}:
  \bibinfo{publisher}{Pearson Prentice Hall}.
\bibitem[{Ott et~al.(2004)Ott, Imoto \& Miyano}]{Ott2004}
\bibinfo{author}{Ott, S.}, \bibinfo{author}{Imoto, S.}, \&
  \bibinfo{author}{Miyano, S.} (\bibinfo{year}{2004}).
\newblock \bibinfo{title}{Finding optimal models for small gene networks}.
\newblock In {\it \bibinfo{booktitle}{Proceedings of the Pacific Symposium on
  Biocomputing}\/} (pp. \bibinfo{pages}{557--567}).
\bibitem[{{Pe{\~n}a} et~al.(2007){Pe{\~n}a}, Nilsson, Bj{\"o}rkegren \&
  Tegn{\'e}r}]{Pen07}
\bibinfo{author}{{Pe{\~n}a}, J.}, \bibinfo{author}{Nilsson, R.},
  \bibinfo{author}{Bj{\"o}rkegren, J.}, \& \bibinfo{author}{Tegn{\'e}r, J.}
  (\bibinfo{year}{2007}).
\newblock \bibinfo{title}{Towards scalable and data efficient learning of
  {M}arkov boundaries}.
\newblock {\it \bibinfo{journal}{International Journal of Approximate
  Reasoning}\/},  {\it \bibinfo{volume}{45}\/}, \bibinfo{pages}{211--232}.
\bibitem[{{Pe{\~n}a} et~al.(2005){Pe{\~n}a}, Bj{\"o}rkegren \&
  Tegn{\'e}r}]{Pen05}
\bibinfo{author}{{Pe{\~n}a}, J.~M.}, \bibinfo{author}{Bj{\"o}rkegren, J.}, \&
  \bibinfo{author}{Tegn{\'e}r, J.} (\bibinfo{year}{2005}).
\newblock \bibinfo{title}{Growing {B}ayesian network models of gene networks
  from seed genes}.
\newblock {\it \bibinfo{journal}{Bioinformatics}\/},  {\it
  \bibinfo{volume}{40}\/}, \bibinfo{pages}{224--229}.
\bibitem[{Pearl(1988)}]{Pea88}
\bibinfo{author}{Pearl, J.} (\bibinfo{year}{1988}).
\newblock {\it \bibinfo{title}{Probabilistic Reasoning in Intelligent Systems:
  Networks of Plausible Inference.}\/}.
\newblock \bibinfo{address}{San Francisco, CA, USA}: \bibinfo{publisher}{Morgan
  Kaufmann}.
\bibitem[{Pe{\~n}a(2008)}]{Pena08}
\bibinfo{author}{Pe{\~n}a, J.~M.} (\bibinfo{year}{2008}).
\newblock \bibinfo{title}{Learning gaussian graphical models of gene networks
  with false discovery rate control}.
\newblock In {\it \bibinfo{booktitle}{European Conference on Evolutionary
  Computation, Machine Learning and Data Mining in Bioinformatics}\/}
  \bibinfo{number}{6} (pp. \bibinfo{pages}{165--176}).
\bibitem[{Pe{\~n}a(2012)}]{Pen12}
\bibinfo{author}{Pe{\~n}a, J.~M.} (\bibinfo{year}{2012}).
\newblock \bibinfo{title}{Finding consensus bayesian network structures}.
\newblock {\it \bibinfo{journal}{Journal of Artificial Intelligence
  Research}\/},  {\it \bibinfo{volume}{42}\/}, \bibinfo{pages}{661--687}.
\bibitem[{Perrier et~al.(2008)Perrier, Imoto \& Miyano}]{Perrier08}
\bibinfo{author}{Perrier, E.}, \bibinfo{author}{Imoto, S.}, \&
  \bibinfo{author}{Miyano, S.} (\bibinfo{year}{2008}).
\newblock \bibinfo{title}{Finding optimal bayesian network given a
  super-structure}.
\newblock {\it \bibinfo{journal}{Journal of Machine Learning Research,
  JMLR}\/},  {\it \bibinfo{volume}{9}\/}, \bibinfo{pages}{2251--2286}.
\bibitem[{Pe’er et~al.(2001)Pe’er, Regev, Elidan \& Friedman}]{Peer2001}
\bibinfo{author}{Pe’er, D.}, \bibinfo{author}{Regev, A.},
  \bibinfo{author}{Elidan, G.}, \& \bibinfo{author}{Friedman, N.}
  (\bibinfo{year}{2001}).
\newblock \bibinfo{title}{Inferring subnetworks from perturbed expression
  profiles.}
\newblock {\it \bibinfo{journal}{Bioinformatics}\/},  {\it
  \bibinfo{volume}{17}\/}, \bibinfo{pages}{215--224}.
\bibitem[{Prestat et~al.(2013)Prestat, {Rodrigues de Morais}, Vendrell,
  Thollet, Gautier, Cohen \& Aussem}]{Prestat13}
\bibinfo{author}{Prestat, E.}, \bibinfo{author}{{Rodrigues de Morais}, S.},
  \bibinfo{author}{Vendrell, J.}, \bibinfo{author}{Thollet, A.},
  \bibinfo{author}{Gautier, C.}, \bibinfo{author}{Cohen, P.}, \&
  \bibinfo{author}{Aussem, A.} (\bibinfo{year}{2013}).
\newblock \bibinfo{title}{{Learning the local Bayesian network structure around
  the ZNF217 oncogene in breast tumours}}.
\newblock {\it \bibinfo{journal}{Computers in Biology and Medicine}\/},  {\it
  \bibinfo{volume}{4}\/}, \bibinfo{pages}{334--341}.
\bibitem[{{R Core Team}(2013)}]{R}
\bibinfo{author}{{R Core Team}} (\bibinfo{year}{2013}).
\newblock {\it \bibinfo{title}{R: A Language and Environment for Statistical
  Computing}\/}.
\newblock \bibinfo{organization}{R Foundation for Statistical Computing}
  \bibinfo{address}{Vienna, Austria}.
\newblock \URLprefix \url{http://www.R-project.org}.
\bibitem[{Read et~al.(2009)Read, Pfahringer, Holmes \& Frank}]{Read2009}
\bibinfo{author}{Read, J.}, \bibinfo{author}{Pfahringer, B.},
  \bibinfo{author}{Holmes, G.}, \& \bibinfo{author}{Frank, E.}
  (\bibinfo{year}{2009}).
\newblock \bibinfo{title}{{Classifier chains for multi-label classification}}.
\newblock In {\it \bibinfo{booktitle}{European Conference on Machine Learning
  and Principles and Practice of Knowledge Discovery in Databases,
  ECML-PKDD}\/} (pp. \bibinfo{pages}{254--269}).
\newblock \bibinfo{publisher}{Springer} volume \bibinfo{volume}{5782} of {\it
  \bibinfo{series}{Lecture Notes in Computer Science}\/}.
\bibitem[{{Rodrigues de Morais} \& Aussem(2010{\natexlab{a}})}]{Morais10b}
\bibinfo{author}{{Rodrigues de Morais}, S.}, \& \bibinfo{author}{Aussem, A.}
  (\bibinfo{year}{2010}{\natexlab{a}}).
\newblock \bibinfo{title}{An efficient learning algorithm for local bayesian
  network structure discovery}.
\newblock In {\it \bibinfo{booktitle}{European Conference on Machine Learning
  and Principles and Practice of Knowledge Discovery in Databases,
  ECML-PKDD}\/} (pp. \bibinfo{pages}{164--169}).
\bibitem[{{Rodrigues de Morais} \& Aussem(2010{\natexlab{b}})}]{Morais10a}
\bibinfo{author}{{Rodrigues de Morais}, S.}, \& \bibinfo{author}{Aussem, A.}
  (\bibinfo{year}{2010}{\natexlab{b}}).
\newblock \bibinfo{title}{A novel {M}arkov boundary based feature subset
  selection algorithm}.
\newblock {\it \bibinfo{journal}{Neurocomputing}\/},  {\it
  \bibinfo{volume}{73}\/}, \bibinfo{pages}{578--584}.
\bibitem[{Roth \& Fischer(2007)}]{Roth07}
\bibinfo{author}{Roth, V.}, \& \bibinfo{author}{Fischer, B.}
  (\bibinfo{year}{2007}).
\newblock \bibinfo{title}{{Improved functional prediction of proteins by
  learning kernel combinations in multilabel settings}}.
\newblock {\it \bibinfo{journal}{BMC Bioinformatics}\/},  {\it
  \bibinfo{volume}{8}\/}, \bibinfo{pages}{S12+}.
\bibitem[{Schwarz(1978)}]{Schwarz78}
\bibinfo{author}{Schwarz, G.~E.} (\bibinfo{year}{1978}).
\newblock \bibinfo{title}{Estimating the dimension of a model}.
\newblock {\it \bibinfo{journal}{Journal of Biomedical Informatics}\/},  {\it
  \bibinfo{volume}{6}\/}, \bibinfo{pages}{461--464}.
\bibitem[{Scutari(2010)}]{Scutari10}
\bibinfo{author}{Scutari, M.} (\bibinfo{year}{2010}).
\newblock \bibinfo{title}{Learning bayesian networks with the bnlearn {R}
  package}.
\newblock {\it \bibinfo{journal}{Journal of Statistical Software}\/},  {\it
  \bibinfo{volume}{35}\/}, \bibinfo{pages}{1--22}.
\bibitem[{Scutari(2011)}]{Scutari11}
\bibinfo{author}{Scutari, M.} (\bibinfo{year}{2011}).
\newblock {\it \bibinfo{title}{Measures of Variability for Graphical
  Models}\/}.
\newblock Ph.D. thesis School in Statistical Sciences, University of Padova.
\bibitem[{Scutari \& Brogini(2012)}]{Scutari12}
\bibinfo{author}{Scutari, M.}, \& \bibinfo{author}{Brogini, A.}
  (\bibinfo{year}{2012}).
\newblock \bibinfo{title}{Bayesian network structure learning with permutation
  tests}.
\newblock {\it \bibinfo{journal}{Communications in Statistics – Theory and
  Methods}\/},  {\it \bibinfo{volume}{41}\/}, \bibinfo{pages}{3233--3243}.
\bibitem[{Scutari \& Nagarajan(2013)}]{Scutari13}
\bibinfo{author}{Scutari, M.}, \& \bibinfo{author}{Nagarajan, R.}
  (\bibinfo{year}{2013}).
\newblock \bibinfo{title}{Identifying significant edges in graphical models of
  molecular networks}.
\newblock {\it \bibinfo{journal}{Artificial Intelligence in Medicine}\/},  {\it
  \bibinfo{volume}{57}\/}, \bibinfo{pages}{207--217}.
\bibitem[{Silander \& Myllym{\"a}ki(2006)}]{Silander06}
\bibinfo{author}{Silander, T.}, \& \bibinfo{author}{Myllym{\"a}ki, P.}
  (\bibinfo{year}{2006}).
\newblock \bibinfo{title}{{A Simple Approach for Finding the Globally Optimal
  Bayesian Network Structure}}.
\newblock In {\it \bibinfo{booktitle}{Uncertainty in Artificial Intelligence,
  UAI}\/} (pp. \bibinfo{pages}{445--452}).
\bibitem[{Snoek et~al.(2006)Snoek, Worring, Gemert, Geusebroek \&
  Smeulders}]{Snoek06}
\bibinfo{author}{Snoek, C.}, \bibinfo{author}{Worring, M.},
  \bibinfo{author}{Gemert, J.~V.}, \bibinfo{author}{Geusebroek, J.}, \&
  \bibinfo{author}{Smeulders, A.} (\bibinfo{year}{2006}).
\newblock \bibinfo{title}{The challenge problem for automated detection of 101
  semantic concepts in multimedia}.
\newblock In {\it \bibinfo{booktitle}{ACM International Conference on
  Multimedia}\/} (pp. \bibinfo{pages}{421--430}).
\newblock \bibinfo{publisher}{ACM Press}.
\bibitem[{Spirtes et~al.(2000)Spirtes, Glymour \& Scheines}]{Spi00}
\bibinfo{author}{Spirtes, P.}, \bibinfo{author}{Glymour, C.}, \&
  \bibinfo{author}{Scheines, R.} (\bibinfo{year}{2000}).
\newblock {\it \bibinfo{title}{Causation, Prediction, and Search}\/}.
\newblock (\bibinfo{edition}{2nd} ed.).
\newblock \bibinfo{publisher}{The {MIT} Press}.
\bibitem[{Spola{\^o}r et~al.(2013)Spola{\^o}r, Cherman, Monard \&
  Lee}]{Spolaor2013}
\bibinfo{author}{Spola{\^o}r, N.}, \bibinfo{author}{Cherman, E.~A.},
  \bibinfo{author}{Monard, M.~C.}, \& \bibinfo{author}{Lee, H.~D.}
  (\bibinfo{year}{2013}).
\newblock \bibinfo{title}{A comparison of multi-label feature selection methods
  using the problem transformation approach}.
\newblock {\it \bibinfo{journal}{Electronic Notes in Theoretical Computer
  Science}\/},  {\it \bibinfo{volume}{292}\/}, \bibinfo{pages}{135--151}.
\bibitem[{Studen\'{y} \& Haws(2014)}]{Studeny2014}
\bibinfo{author}{Studen\'{y}, M.}, \& \bibinfo{author}{Haws, D.}
  (\bibinfo{year}{2014}).
\newblock \bibinfo{title}{{Learning Bayesian network structure: Towards the
  essential graph by integer linear programming tools}}.
\newblock {\it \bibinfo{journal}{International Journal of Approximate
  Reasoning}\/},  {\it \bibinfo{volume}{55}\/}, \bibinfo{pages}{1043--1071}.
\bibitem[{Trohidis et~al.(2008)Trohidis, Tsoumakas, Kalliris \&
  Vlahavas}]{Trohidis08}
\bibinfo{author}{Trohidis, K.}, \bibinfo{author}{Tsoumakas, G.},
  \bibinfo{author}{Kalliris, G.}, \& \bibinfo{author}{Vlahavas, I.}
  (\bibinfo{year}{2008}).
\newblock \bibinfo{title}{Multi-label classification of music into emotions}.
\newblock In {\it \bibinfo{booktitle}{ISMIR}\/} (pp.
  \bibinfo{pages}{325--330}).
\bibitem[{Tsamardinos et~al.(2003)Tsamardinos, Aliferis \&
  Statnikov}]{Tsamardinos03}
\bibinfo{author}{Tsamardinos, I.}, \bibinfo{author}{Aliferis, C.}, \&
  \bibinfo{author}{Statnikov, A.} (\bibinfo{year}{2003}).
\newblock \bibinfo{title}{Algorithms for large scale {M}arkov blanket
  discovery.}
\newblock In {\it \bibinfo{booktitle}{Florida Artificial Intelligence Research
  Society Conference FLAIRS'03}\/} (pp. \bibinfo{pages}{376--381}).
\bibitem[{Tsamardinos \& Borboudakis(2010)}]{Tsamardinos10}
\bibinfo{author}{Tsamardinos, I.}, \& \bibinfo{author}{Borboudakis, G.}
  (\bibinfo{year}{2010}).
\newblock \bibinfo{title}{Permutation testing improves bayesian network
  learning}.
\newblock In {\it \bibinfo{booktitle}{European Conference on Machine Learning
  and Knowledge Discovery in Databases, ECML-PKDD}\/} (pp.
  \bibinfo{pages}{322--337}).
\bibitem[{Tsamardinos et~al.(2006)Tsamardinos, Brown \&
  Aliferis}]{Tsamardinos06}
\bibinfo{author}{Tsamardinos, I.}, \bibinfo{author}{Brown, L.}, \&
  \bibinfo{author}{Aliferis, C.} (\bibinfo{year}{2006}).
\newblock \bibinfo{title}{{The Max-Min Hill-Climbing {B}ayesian Network
  Structure Learning Algorithm}}.
\newblock {\it \bibinfo{journal}{Machine Learning}\/},  {\it
  \bibinfo{volume}{65}\/}, \bibinfo{pages}{31--78}.
\bibitem[{Tsamardinos \& Brown(2008)}]{Tsamardinos08}
\bibinfo{author}{Tsamardinos, I.}, \& \bibinfo{author}{Brown, L.~E.}
  (\bibinfo{year}{2008}).
\newblock \bibinfo{title}{Bounding the false discovery rate in local {B}ayesian
  network learning}.
\newblock In {\it \bibinfo{booktitle}{AAAI Conference on Artificial
  Intelligence}\/} (pp. \bibinfo{pages}{1100--1105}).
\bibitem[{Tsoumakas et~al.(2010{\natexlab{a}})Tsoumakas, Katakis \&
  Vlahavas}]{Tsoumakas2010a}
\bibinfo{author}{Tsoumakas, G.}, \bibinfo{author}{Katakis, I.}, \&
  \bibinfo{author}{Vlahavas, I.} (\bibinfo{year}{2010}{\natexlab{a}}).
\newblock \bibinfo{title}{{Mining Multi-label Data}}.
\newblock {\it \bibinfo{journal}{Transformation}\/},  {\it
  \bibinfo{volume}{135}\/}, \bibinfo{pages}{1--20}.
\bibitem[{Tsoumakas et~al.(2010{\natexlab{b}})Tsoumakas, Katakis \&
  Vlahavas}]{Tsoumakas2010}
\bibinfo{author}{Tsoumakas, G.}, \bibinfo{author}{Katakis, I.}, \&
  \bibinfo{author}{Vlahavas, I.} (\bibinfo{year}{2010}{\natexlab{b}}).
\newblock \bibinfo{title}{{Random k-labelsets for Multi-Label Classification}}.
\newblock {\it \bibinfo{journal}{{IEEE Transactions on Knowledge and Data
  Engineering, TKDE}}\/},  {\it \bibinfo{volume}{23}\/},
  \bibinfo{pages}{1--12}.
\bibitem[{Tsoumakas \& Vlahavas(2007)}]{Tsoumakas2007}
\bibinfo{author}{Tsoumakas, G.}, \& \bibinfo{author}{Vlahavas, I.}
  (\bibinfo{year}{2007}).
\newblock \bibinfo{title}{{Random k-labelsets: An ensemble method for
  multilabel classification}}.
\newblock {\it \bibinfo{journal}{Proceedings of the 18th European Conference on
  Machine Learning}\/},  {\it \bibinfo{volume}{4701}\/},
  \bibinfo{pages}{406--417}.
\bibitem[{Villanueva \& Maciel(2012)}]{Villanueva12}
\bibinfo{author}{Villanueva, E.}, \& \bibinfo{author}{Maciel, C.}
  (\bibinfo{year}{2012}).
\newblock \bibinfo{title}{Optimized algorithm for learning bayesian network
  superstructures}.
\newblock In {\it \bibinfo{booktitle}{International Conference on Pattern
  Recognition Applications and Methods, ICPRAM}\/}.
\bibitem[{Villanueva \& Maciel(2014)}]{VillanuevaM14}
\bibinfo{author}{Villanueva, E.}, \& \bibinfo{author}{Maciel, C.~D.}
  (\bibinfo{year}{2014}).
\newblock \bibinfo{title}{Efficient methods for learning bayesian network
  super-structures.}
\newblock {\it \bibinfo{journal}{Neurocomputing}\/},  (pp.
  \bibinfo{pages}{3--12}).
\bibitem[{Zhang \& Zhang(2010)}]{Zhang10}
\bibinfo{author}{Zhang, M.~L.}, \& \bibinfo{author}{Zhang, K.}
  (\bibinfo{year}{2010}).
\newblock \bibinfo{title}{{Multi-label learning by exploiting label
  dependency}}.
\newblock In {\it \bibinfo{booktitle}{European Conference on Machine Learning
  and Principles and Practice of Knowledge Discovery in Databases, ECML
  PKDD}\/} (p. \bibinfo{pages}{999}).
\newblock \bibinfo{publisher}{ACM Press} volume~\bibinfo{volume}{16} of {\it
  \bibinfo{series}{Knowledge Discovery in Databases, KDD}\/}.
\bibitem[{Zhang \& Zhou.(2006)}]{Zhang2006}
\bibinfo{author}{Zhang, M.-L.}, \& \bibinfo{author}{Zhou., Z.-H.}
  (\bibinfo{year}{2006}).
\newblock \bibinfo{title}{Multilabel neural networks with applications to
  functional genomics and text categorization}.
\newblock {\it \bibinfo{journal}{IEEE Transactions on Knowledge and Data
  Engineering}\/},  {\it \bibinfo{volume}{18}\/}, \bibinfo{pages}{1338 --
  1351}.

\end{thebibliography}

\end{document}